\documentclass{article}

\usepackage[final]{neurips_glfrontiers_2023}

\usepackage{natbib}
\setcitestyle{round}

\usepackage[utf8]{inputenc} 
\usepackage[T1]{fontenc}    

\usepackage{hyperref}       
\usepackage{url}            
\usepackage{booktabs}       
\usepackage{amsfonts}       
\usepackage{nicefrac}       
\usepackage{microtype}      
\usepackage{xcolor,colortbl}         

\usepackage{rotating}

\usepackage{graphicx}
\usepackage{caption}

\usepackage{tikz}

\usepackage[lofdepth,lotdepth]{subfig}
\usepackage{amsmath}
\usepackage[inline]{enumitem}

\usepackage{color}
\definecolor{orange}{rgb}{1,0.5,0}

\newcommand{\m}[1]{$#1$}
\newcommand{\inm}{\ensuremath{\!\in\!}}

\newcommand{\rotateheader}[1]{\begin{turn}{80}#1\end{turn}}

\title{Knowledge Graphs are not Created Equal: Exploring the Properties and Structure of Real KGs}

\author{%
   Nedelina Teneva  \thanks{Corresponding author} \\
   Megagon Labs\\
   Mountain View, CA \\
  \texttt{nedteneva@gmail.com} \\
   \And
   Estevam Hruschka \\
   Megagon Labs \\
   Mountain View, CA  \\
   \texttt{estevam@megagon.ai} \\
}

\begin{document}
\maketitle
\setcounter{footnote}{0} 

\begin{abstract}
Despite the recent popularity of knowledge graph (KG) related tasks and benchmarks such as KG embeddings, link prediction, entity alignment and evaluation of the reasoning abilities of pretrained language models as KGs, 
the structure and properties of real KGs are not well studied. 
In this paper, we perform a large scale comparative study of \m{29} real KG datasets from diverse domains 
such as the natural sciences, medicine, and NLP to analyze 
their properties and structural patterns. Based on our findings, we make several recommendations regarding KG-based model development and evaluation. We believe that the rich structural information contained in KGs can benefit the development of better KG models across fields and we hope this study will contribute to breaking the existing data silos between different areas of research (e.g., ML, NLP, AI for sciences).
\end{abstract}

\vspace{-6pt}
\section{Introduction}\vspace{-6pt}
Recent years have been marked by an increased use of multimodal and structured datasets in the form of knowledge graphs (KGs) to enhance applications in diverse scientific and technical disciplines such as natural language processing (NLP), natural sciences and medicine, manufacturing and process automation to name a few \citep{KGphysics,kg-opportunities}. The wide applicability of KGs is not surprising: they are scalable data objects that store factual (i.e., with high degree of certainty) information in the form of triples and allow for encoding both topological 
and semantic information. 

The growing interest in using KG across various domains has led to a surge in new KG dataset releases: 
for example,  \m{43\%} the $37$ datasets available in the PyKEEN v1.10 \footnote{Permalink: https://ezproxy.library.und.edu/login?url=https://github.com/pykeen/pykeen} KG embedding library  \citep{ali2021pykeen} have been published since 2020.
Many other libraries -- e.g., OGB \citep{ogb,org-largescale}, LibKGE \citep{ruffinelli2020you, libkge} and PyTorch Geometric \citep{pyg} --  consolidate multiple KG models in a central repository or provide tools for task-specific benchmarking.
Lastly, several recent studies focus on scalable benchmarking of KG related tasks:
for example \cite{bringingiglight} compare the performance of \m{21} KG link prediction models, \cite{sun2020benchmarking} evaluate \m{12} embedding-based entity alignment (EA) approaches on dedicated benchmark datasets, while \cite{alkhamissi2022review} propose a KG-based framework for assessing
the performance of pretrained language models (PLMs) with the goal of achieving parity between PLMs and KGs.

Even with the abundance of KG datasets, benchmarking tools, and extensive large-scale model comparisons across various KG tasks, 
to the best of our knowledge, there are no studies that address a much more fundamental question, namely:  \textit{what properties and structure do real KGs have and how do they compare to each other in terms of these properties?}
We argue that a systematic approach of analyzing KG properties (the goal of this paper) 
has the potential to inform algorithmic and dataset development across disciplines and empower the next generation of KG-based applications in NLP, biomedicine and other disciplines.

\textbf{Our Contribution}. In order to begin addressing the above question, 
we analyze the structure of KGs in terms of their network statistics, topology and relation types.
Our large scale comparative study is based on \m{29} real KG datasets from diverse domains such as biology, medicine, and NLP. Towards our goal, we: (i) measure various KG properties (e.g., KG density, degree distribution); (2) analyze the KG structure in terms of the relational types and the KG topology; and (3) describe common/distinct structural patterns we observe in KG datasets derived from fundamentally different underlying domains. 
We also highlight potential issues that may arise 
when using off-the-shelf KG techniques (e.g. embeddings, link prediction models) 
without accounting for the domain-specific  KG structural properties.
Based on our findings, we make several recommendations for future model development and evaluation.

\textbf{Scope}. Our primary goal is to analyze KG datasets and their properties along different dimensions, rather than benchmark task-specific models on said datasets. While evaluating how KG properties affect downstream algorithms such as link prediction or EA is a natural extension of this work, we intentionally leave this analysis for a follow up study in order to focus the analysis on the KG properties themselves.

\begin{table}[t]
  \caption{Datasets that were analyzed in this study. \#E, \#R, and \#T denote the number of entities, relations and triples, respectively. \m{deg} denotes the average degree of all the KG entities. \m{d} denotes the KG density, shown in \m{\log} scale: a lower column value implies a denser KG.} 
  \label{tbl: datasetstable}
  \centering
  \begin{tabular}{llrrrlrr}
    \toprule
    \cmidrule(r){1-2}
    \# & \textbf{dataset}   & \textbf{\# E}  & \textbf{\# R} & \textbf{\# T} & \textbf{category}  & \textbf{\m{deg}} & \textbf{\m{-\log(d)}} \\
    \midrule
1	&	AristoV4	& \m{	42,016	} & \m{	1,593	} & \m{	279,425	} &	biomed	& \m{	7	} & \m{	3.80	}	\\
2	&	BioKG	& \m{	105,524	} & \m{	17	} & \m{	2,067,997	} &	biomed	& \m{	20	} & \m{	3.73	}	\\
3	&	CoDExLarge	& \m{	77,951	} & \m{	69	} & \m{	612,437	} &	semantic	& \m{	8	} & \m{	4.00	}	\\
4	&	CoDExMedium	& \m{	17,050	} & \m{	51	} & \m{	206,205	} &	semantic	& \m{	12	} & \m{	3.15	}	\\
5	&	CoDExSmall	& \m{	2,034	} & \m{	42	} & \m{	36,543	} &	semantic	& \m{	18	} & \m{	2.05	}	\\
6	&	ConceptNet	& \m{	28,370,083	} & \m{	50	} & \m{	34,074,917	} &	semantic	& \m{	1	} & \m{	7.37	}	\\
7	&	Countries	& \m{	271	} & \m{	2	} & \m{	1,158	} &	society	& \m{	4	} & \m{	1.80	}	\\
8	&	CSKG	& \m{	2,087,833	} & \m{	58	} & \m{	4,598,728	} &	semantic	& \m{	3	} & \m{	5.98	}	\\
9	&	DB100K	& \m{	99,604	} & \m{	470	} & \m{	697,479	} &	semantic	& \m{	7	} & \m{	4.15	}	\\
10	&	DBpedia50	& \m{	24,624	} & \m{	351	} & \m{	34,421	} &	semantic	& \m{	1	} & \m{	4.25	}	\\
11	&	DRKG	& \m{	97,238	} & \m{	107	} & \m{	5,874,257	} &	biomed	& \m{	60	} & \m{	3.21	}	\\
12	&	FB15k	& \m{	14,951	} & \m{	1,345	} & \m{	592,213	} &	semantic	& \m{	40	} & \m{	2.58	}	\\
13	&	FB15k-237	& \m{	14,505	} & \m{	237	} & \m{	310,079	} &	semantic	& \m{	21	} & \m{	2.83	}	\\
14	&	Globi	& \m{	404,207	} & \m{	39	} & \m{	1,966,385	} &	biomed	& \m{	5	} & \m{	4.92	}	\\
15	&	Hetionet	& \m{	45,158	} & \m{	24	} & \m{	2,250,197	} &	biomed	& \m{	50	} & \m{	2.96	}	\\
16	&	Kinships	& \m{	104	} & \m{	25	} & \m{	10,686	} &	society	& \m{	103	} & \m{	0.01	}	\\
17	&	Nations	& \m{	14	} & \m{	55	} & \m{	1,992	} &	society	& \m{	143	} & \m{	-1.01	}	\\
18	&	OGBWikiKG2	& \m{	2,500,604	} & \m{	535	} & \m{	17,137,181	} &	semantic	& \m{	7	} & \m{	5.56	}	\\
19	&	OpenBioLink	& \m{	180,992	} & \m{	28	} & \m{	4,563,407	} &	biomed	& \m{	25	} & \m{	3.86	}	\\
20	&	OpenEA	& \m{	15,000	} & \m{	248	} & \m{	38,265	} &	semantic	& \m{	3	} & \m{	3.77	}	\\
21	&	PharmKG	& \m{	188,296	} & \m{	39	} & \m{	1,093,236	} &	biomed	& \m{	6	} & \m{	4.51	}	\\
22	&	PharmKG8k	& \m{	7,247	} & \m{	28	} & \m{	485,787	} &	biomed	& \m{	67	} & \m{	2.03	}	\\
23	&	PrimeKG	& \m{	129,375	} & \m{	30	} & \m{	8,100,498	} &	biomed	& \m{	63	} & \m{	3.32	}	\\
24	&	UMLS	& \m{	135	} & \m{	46	} & \m{	6,529	} &	biomed	& \m{	48	} & \m{	0.45	}	\\
25	&	WD50K	& \m{	40,107	} & \m{	473	} & \m{	232,344	} &	semantic	& \m{	6	} & \m{	3.84	}	\\
26	&	Wikidata5M	& \m{	4,594,149	} & \m{	822	} & \m{	20,624,239	} &	semantic	& \m{	4	} & \m{	6.01	}	\\
27	&	WN18	& \m{	40,943	} & \m{	18	} & \m{	151,442	} &	semantic	& \m{	4	} & \m{	4.04	}	\\
28	&	WN18RR	& \m{	40,559	} & \m{	11	} & \m{	92,583	} &	semantic	& \m{	2	} & \m{	4.25	}	\\
29	&	YAGO3-10	& \m{	123,143	} & \m{	37	} & \m{	1,089,000	} &	semantic	& \m{	9	} & \m{	4.14	}	\\
    \bottomrule
  \end{tabular}
\end{table}
\vspace{-6pt}
\section{Application of Knowledge Graphs in Different Domains}\label{sec: kgapplication}
\vspace{-6pt}
\textbf{Notation}. For a given set of entities \m{E} and a set of relations \m{R}, a knowledge graph 
\m{\mathcal{K}\!\!\subseteq \!\! K \!=\! E \!\times\! R \!\times\! E} is a directed multi-relational graph that contains triples of the form \m{(h,r,t)\inm \mathcal{K}} in which \m{h,t\inm E} represent the head and tail entities and \m{r\inm R} is the relation between them. 
KG embedding models (e.g., TransE \citep{transe}, DistMult \citep{distmult}) learn latent vector representations of the entities \m{e\inm E} and relations \m{r\inm R} that best preserve the KG's structural properties. 

\textbf{Natural Language Processing}.
In NLP PLMs have gained immense popularity in recent years due to their impressive ability to process and generate human-like text. PLMs, such as GPT-4 \citep{gpt4}, Llama \citep{llama} or Alpaca \citep{alpaca}, are able to generate answers to complex user queries on a variety of technical topics. 
However, these models are known to suffer from a lack of grounding of their outputs (in factual, common sense and domain specific knowledge) and from having difficulties in properly dealing with the meaning of inter-related concepts \citep{carta2023grounding}.
Based on these facts, many approaches have been proposed for strengthening PLMs with KGs. 
\cite {yang2023chatgpt} categorize these approaches in three main groups: 
(i) before-training enhancement, 
(ii) during-training enhancement, and 
(iii) post-training enhancement. Other recent surveys on knowledge-enhanced PLMs 
\citep{wei2021knowledge,zamini2022review,hu2023survey} also propose approaches which  
incorporate factuality, common sense, physical and domain specific knowledge to mitigate the weaknesses of PLMs, 
but none of those approaches consider the properties of the KGs at hand. 
Spurred by the LAMA benchmark \citep{lama}, which analyzed the amount of relational knowledge already present in a wide range of PLMs (without fine-tuning), many works also use KGs for fine-grained evaluation of different aspects of PLMs such as their ability to recover factual knowledge  or their  consistency \citep{heinzerling-inui-2021-language}. 
For a survey of recent methods for interpreting and evaluating "LMs as KBs", see \citep{alkhamissi2022review}. 

\begin{figure} [t]
     \centering
     \begin{minipage}{0.32\textwidth}
         \includegraphics[width=\textwidth]{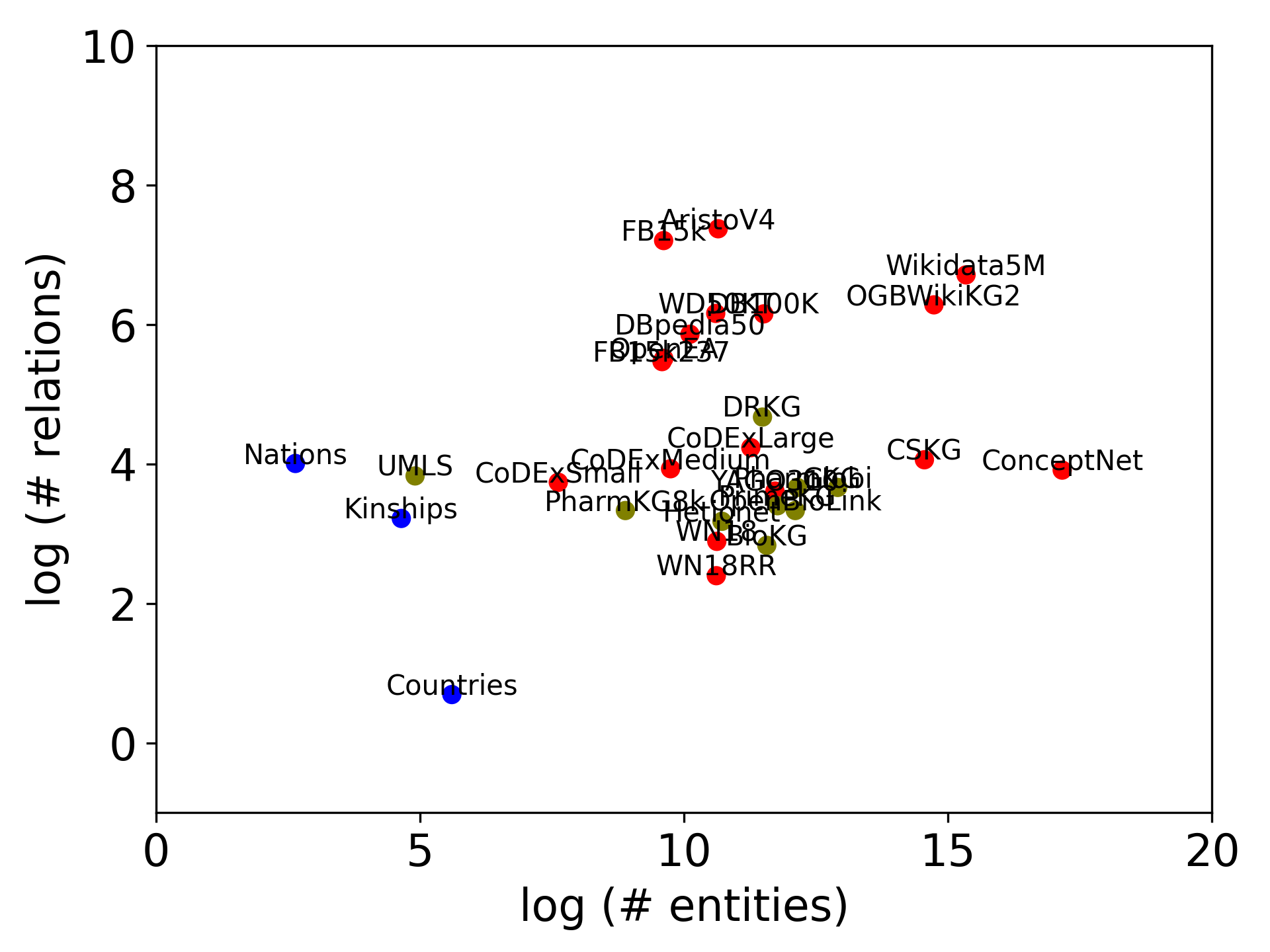}
     \end{minipage}
     \begin{minipage}{0.32\textwidth}
         \includegraphics[width=\textwidth]{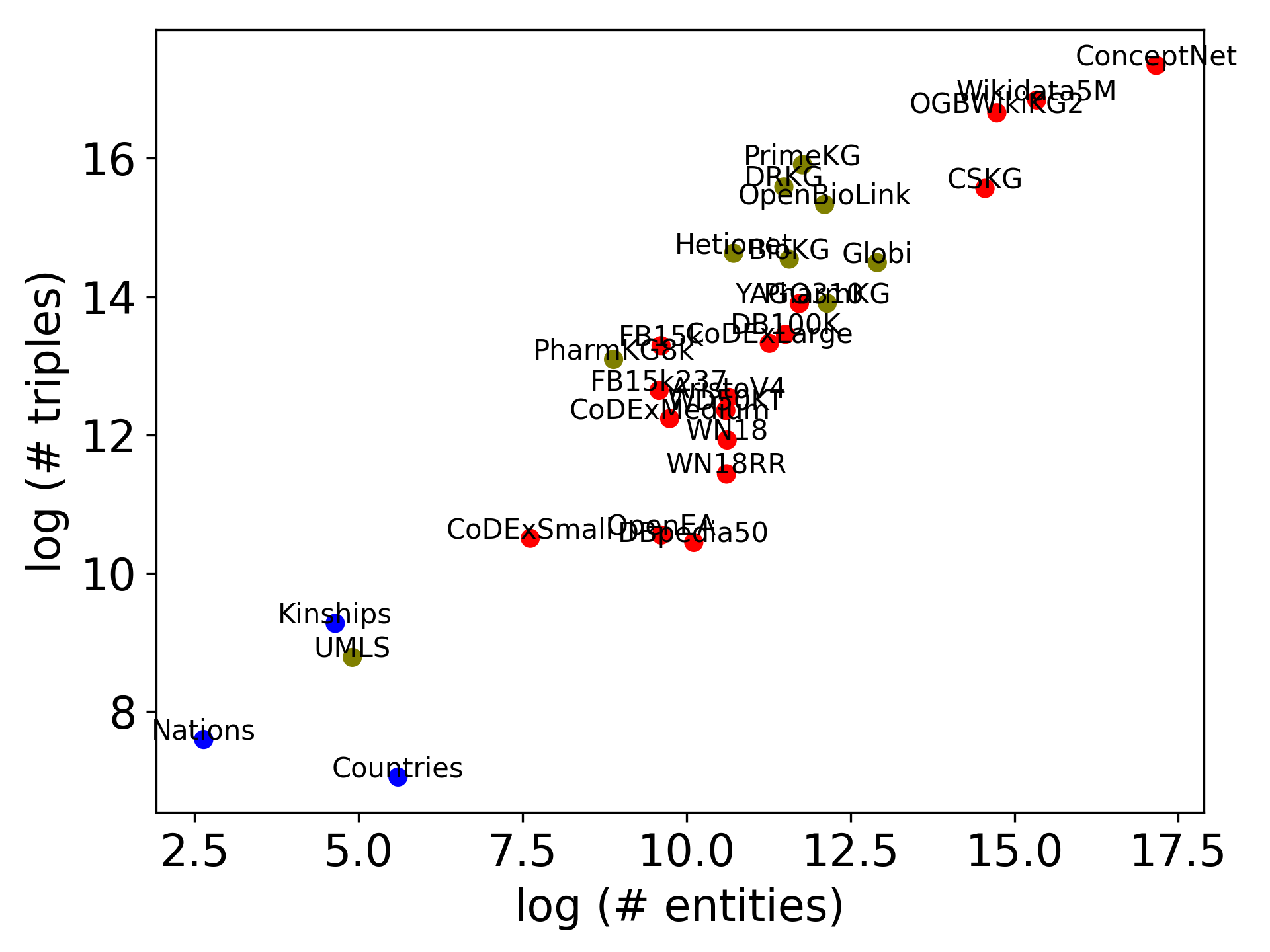}
    \end{minipage}
    \begin{minipage}{0.32\textwidth}
         \includegraphics[width=\textwidth]{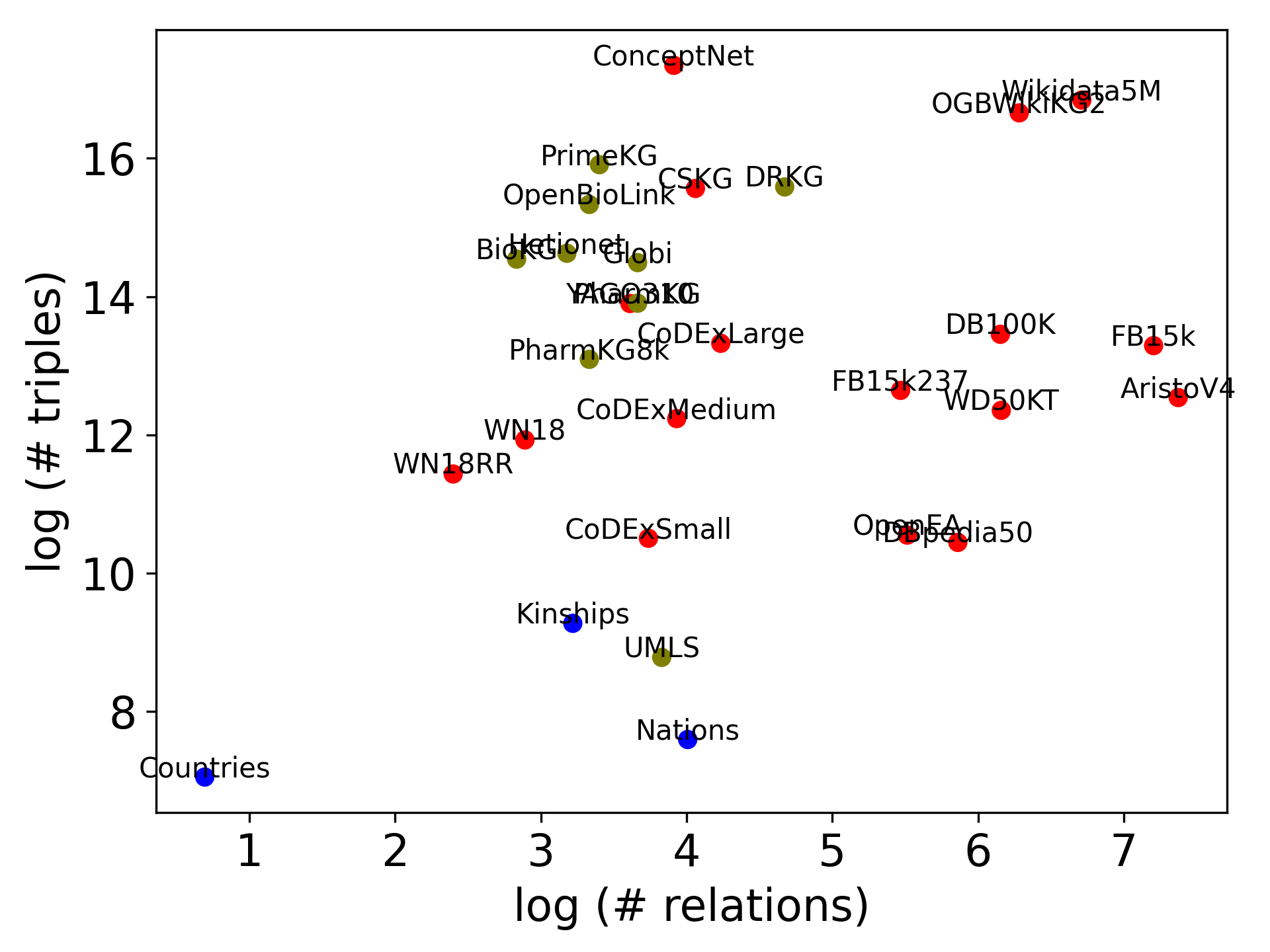}
    \end{minipage}
        \caption{Relations between (left) number of entities vs. number of relations; (center) number of entities vs. number of triples; (right) number of relations vs. number of triples in different KGs. Biomedical, semantic web and societal datasets are colored in resp. olive, red, and blue. }
        \label{fig: relationships}
\end{figure}
\begin{figure}[t]
     \centering
     \begin{minipage}{0.32\textwidth}
         \includegraphics[width=\textwidth]{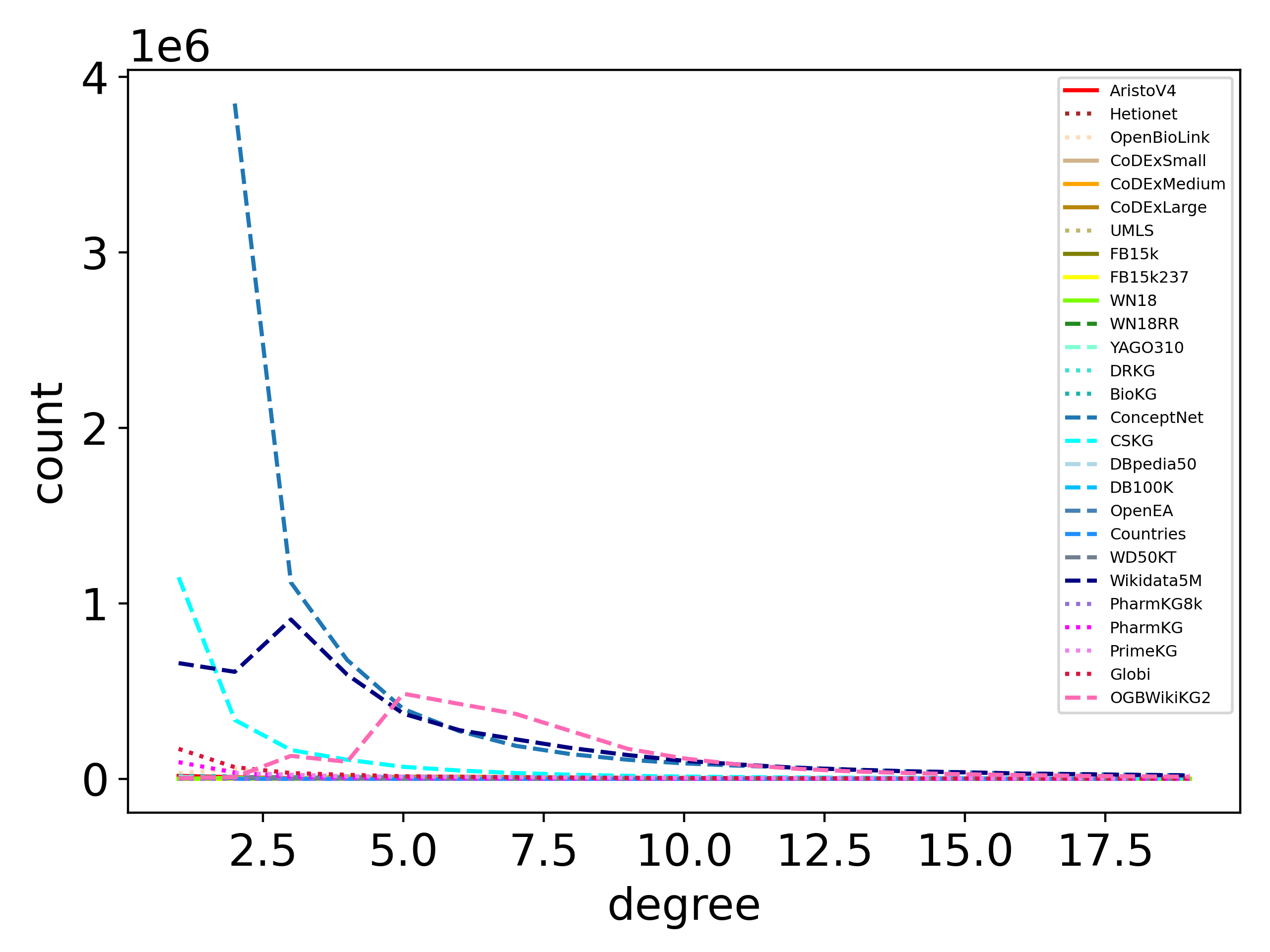}
     \end{minipage}
      \begin{minipage}{0.32\textwidth}
          \includegraphics[width=\textwidth]{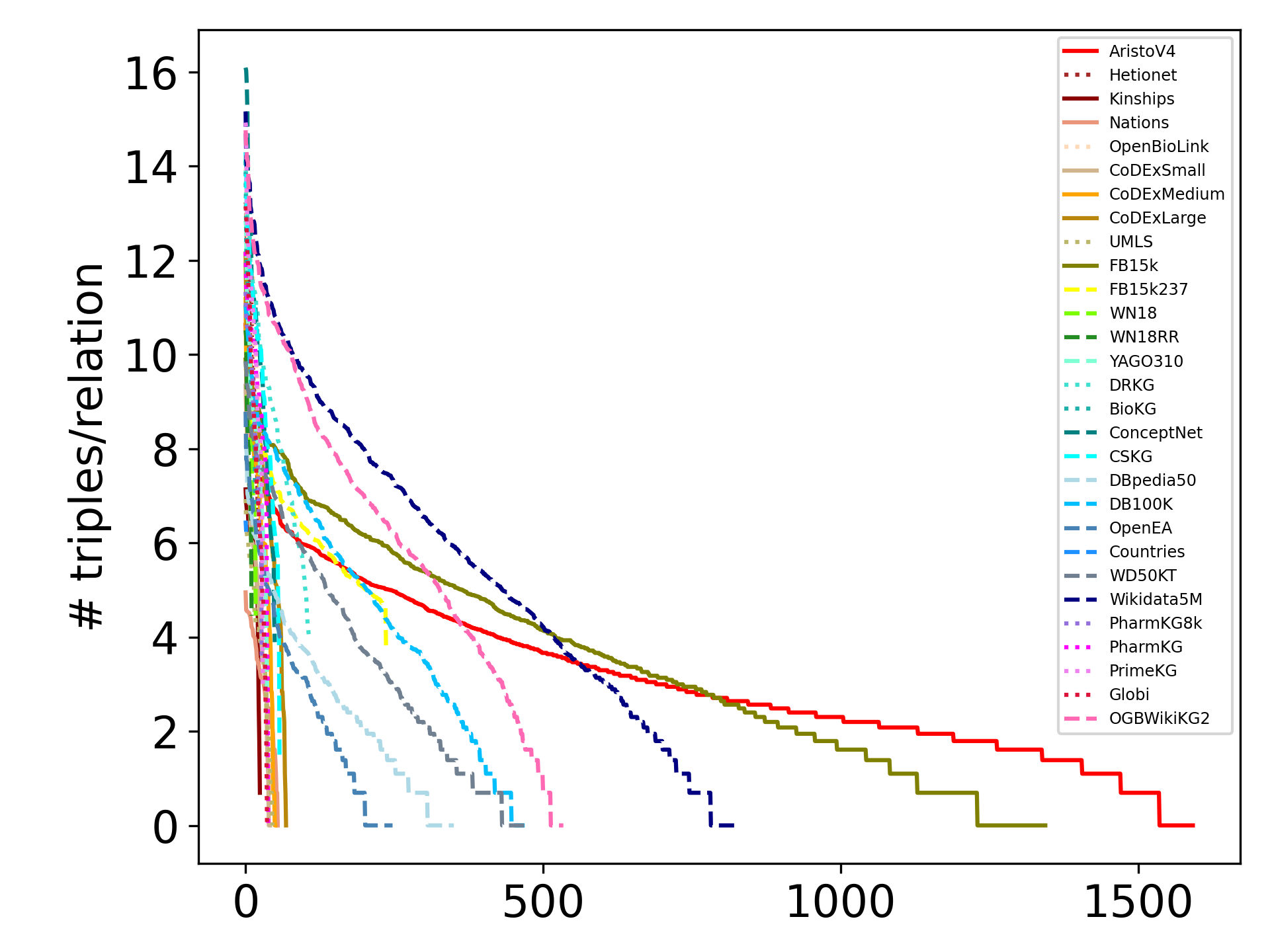}
     \end{minipage}
     \begin{minipage}{0.32\textwidth}
          \includegraphics[width=\textwidth]{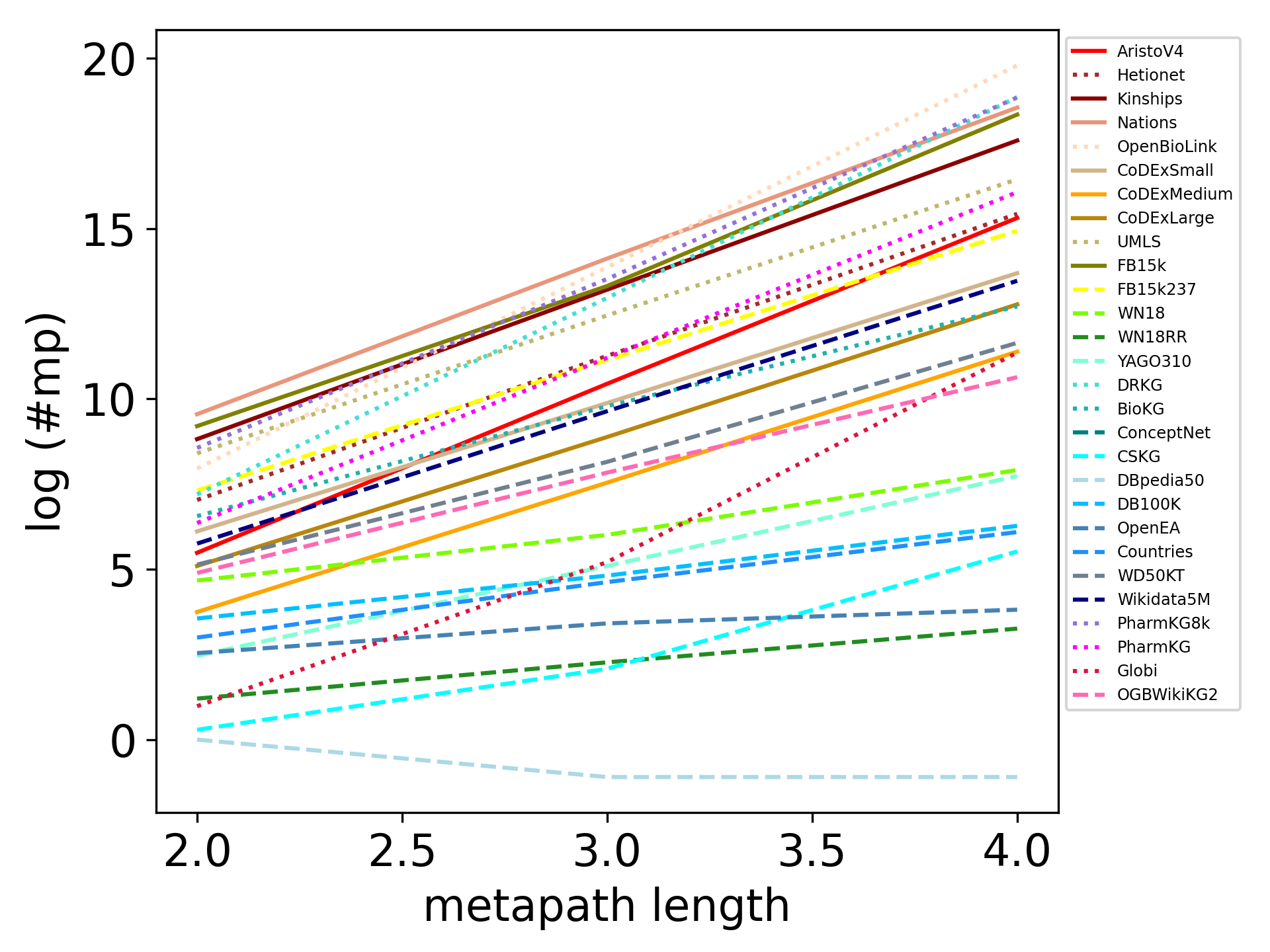}
     \end{minipage}
        \caption{(left) Degree distribution; (center) number of triples per KG relation, \m{log} scale; (right) metapath (\texttt{mp}) length distribution on the \m{y}-axis, \m{log} scale.}
        \label{fig: multiples}
\end{figure}

\textbf{Natural Sciences and Medicine}. KGs are used in various biomedical applications \citep{nicholson2020constructing,zitnikmultimodal} and  have recently found use in precision medicine \citep{KGprecisionmedicine}. In biology and medicine KGs typically describe the relationships between biomedical entities such as diseases, drugs, phenotypes, or regulatory pathways. They are a convenient tool for aggregating knowledge fragmented across publications, repositories, ontologies and databses \citep{KGprecisionmedicine}. KG embeddings and link prediction find application in pharmaceutical applications (e.g. discovery and drug repurposing), clinical applications (e.g., disease diagnosis and treatment) and genomics (e.g., the study of phenotyping) \citep{morselli2021network,KGprecisionmedicine,wang2023scientific}.
Other natural science disciplines such as physics \citep{KGphysics} and geology \citep{KGgeology} make use of multimodal data such as scientific literature and other natural language datasets to construct domain-specific KGs.

\textbf{Other domains}. Many other domains such as cybersecuirty, finance, education, factory monitoring and process automation, geopolitical events and combating human trafficking benefit from KGs and apply KG tasks such as EA and link prediction -- \cite {li2020domain} and \cite{KGphysics} provide a review of domain-specific KGs and their downstream use in these areas.

\textbf {Knowledge Graph Datasets Analyzed}. 
Among all the datasets in the various domains described above, we used a set of \m{29} KGs listed in Tab. \ref{tbl: datasetstable} together with their summary statistics. We categorize the KGs into three distinct groups: 
\vspace{-6pt}
\begin{enumerate} [label=(\alph*), left= 1pt]
    \item \m{9} \textbf{biomedical KGs} which store facts  related to biology and medicine such as relationships between genes, proteins or cellular pathways. Datasets in this group are typically derived from high quality public databases such as DrugBank \citep{drugbank} and PubChem \citep{kim2016pubchem} -- for construction details see \citep{pharmkg}. 
    \item \m{17} \textbf{semantic web KGs} which incorporate knowledge extracted using the tools of the semantic web or analogous mechanism such as RDF \citep{fensel2005spinning}. While the underlying KG ontologies in these datasets may be manually created by experts, the underlying \m{(h,r,t)} triples are typically extracted by parsing outsourced textual content such as Wikipedia or its structured counterpart Wikidata 
    \footnote {https://www.wikidata.org/wiki/Wikidata:Database\_download}. 
    Many datasets in this category are derived from each other or share common sources -- for example ConceptNet \citep{conceptnet} is based, in part, on DBpedia \citep{dbpedia}, while CSKG \citep{cskg} makes use of ConceptNet and  Wikidata.
    \item \m{3} \textbf{societal KGs} are a set of manually curated datasets that contain factual information about different domains such as geography and international relations -- e.g., the Nations dataset encodes relationships between countries such as "military alliance" and "independence [sic. from]". We note that these KGs are conceptually similar to the ones in the biomedical domain in the sense that they are based on relationships between objects, rathern than semantics.
\end{enumerate}
\vspace{-6pt}
We use the datasets through the PyKEEN v\m{1.10} software package
(see details in the Appx.) 

\vspace{-6pt}
\section{The Properties and Structure of Knowledge Graphs}
\vspace{-6pt}
The structural characteristics of KGs play an important role in the applicability and the performance of various tasks such as KG embeddings, link prediction and reasoning. For example, KG properties such as relation type (e.g., inverse, symmetric), cardinality and statistics  affect the KG connectivity patterns which get encoded in the KG node and relation embeddings and used in downstream models. 
The effect releation types have was demonstrated by \cite {toutanova} who first described the "inverse relation problem" in KG link prediction: essentially, an information leakage between the train and test splits due to the presence of \textit{inverse} relations in the training dataset splits. They identified the issue in the FB15k dataset and released its updated version FB15k-237, both commonly used in NLP benchmarks.
Moreover, the inference abilities of KG embedding algorithms vary by relation type and carinality 
For example, TransE cannot model symmetric and one-to-many relations well due to its scoring function \m{f (r,h,t) = -|| h\! +\!r\!-\!t||}. Similarly, the distance functions for DistMult and ComplEx cannot model composite relations. \cite{cao2022} provide a detailed review on this topic. 

In the machine learning (ML) and NLP KG-related literature semantic web datasets are prevalent. 
At the same time the performance of KG embeddings, EA models and link prediction  (both embedding-based and Graph Neural Network (GNN)-based \citep{cucala2021explainable} approaches)  is often demonstrated only on a 
small set of semantic web datasets such as FB15k, WN18 and Wikidata5M.
On the other hand, in the biomedical domain it has been widely accepted that semantic web datasets 
do not reflect the domain specific properties of biomedical KGs due to a variety of factors  \citep{pharmkg,openbiolink}.
One such factor are the interaction effects in KGs: biomedical KGs have  been found to be sparse, incomplete and containing richly structured ontological hierarchies with large interaction networks instead of capturing knowledge networks (e.g., FB15k) or hierarchical taxonomies (e.g., WN18)\citep {pharmkg,openbiolink}. 
Another distinguishing characteristic of biomedical KGs is the nature of the entities stored in them -- for example automatically created datasets such as OpenBiolink \citep{openbiolink} contain large number of meta data entities and trivial biomedical entities which can interfere with KG embeddings and link prediction models as reported by \cite {pharmkg}.

\textbf{KG vs. Graph Structure}. 
KGs remain relatively unexplored as topological objects, partially because
KG datasets 
were not readily available in libraries or benchmarks, such as PyKEEN or PyTorch Geometric until recent years. In contrast, as network analysis gained momentum in the 1990s, thanks to the growing availability of web, social and other real networks, the topology and numerical characteristics of directed and undirected graphs (the homogeneous counterparts of KGs) have been extensively studied. 
Multiple libraries and benchmarks have been established for the analysis of graph/network properties -- examples include the general purpose network and graph mining library SNAP \citep{snap} and the SuiteSparse matrix collection \citep{suitesparse-collection,suitesparse-web} which systematizes graphs together with their numerical characteristics.

\textbf{Graph Structure in Existing Benchmarks}.
Many KG benchmarks \citep{bringingiglight,kgxboard} focus on standardizing model training, hyper parameter tuning or task-specific model evaluation. However, only a few benchmarks provide support for evaluation of KG tasks with respect to the underlying KG structural properties. 
One example is the KGxBoard framework for KG link prediction evaluation \citep {kgxboard}.
KG link prediction performance is often measured using metrics (e.g., precision) averaged over a held-out set, however, as noted by \cite{kgxboard}, single-score summary metrics cannot reveal exactly what the model has learned or failed to learn. 
To remedy this, the KGxBoard framework implements a fine-grained performance reporting per relation type. 
While in principle the KGxBoard framework offers support for multiple relation types, 
the authors did not perform
a systematic evaluation of link prediction performance per relation type/property.
In another benchmark, \cite {bringingiglight} evaluate \m{21} KG link prediction models with respect to four relational patterns on \m{4} datasets (mix of semantic web and societal),  
 however, they do not   analyze the link prediction performance in the context of the KG relational distributions and properties.
Lastly, \citep{sun2020benchmarking,eareeval} benchmark KG heterogenity and its effect on EA performance.  
%

\begin{table}[t]
  \caption{Summary of findings from Sec. \ref{sec: results}: size, density (triples per relation), relation cardinaility, relation pattern and number of metapaths. Colors blue, red, gray denote low, medium, mixed levels of each characteristic for the specified KG domain (each row). 
  } 
  \label{tbl: comparisontable}
  \centering
  \begin{tabular}{lrrrrrrrrrrrrrr}
    \toprule
    KG domain &  \rotateheader{size} & \rotateheader{dens.} & \rotateheader{M-M} &  \rotateheader{M-1} & \rotateheader{1-M} & \rotateheader{1-1}  & \rotateheader{antiS} & \rotateheader{S}  & \rotateheader{Inv}  & \rotateheader{Comp} & \rotateheader{\#paths} \\
    \midrule
    semantic   & \cellcolor{red}  & \cellcolor{blue}    & \cellcolor{blue} & \cellcolor{gray} &\cellcolor{gray} & \cellcolor{red}  & \cellcolor{red}  &  \cellcolor{gray}  & \cellcolor{gray} & \cellcolor{gray}  & \cellcolor{blue} \\
    biomedical  & \cellcolor{gray} &  \cellcolor{red}	& \cellcolor{red} & \cellcolor{red} & \cellcolor{blue}& \cellcolor{blue} & \cellcolor{red} &\cellcolor{gray} & \cellcolor{blue} & \cellcolor{blue} & \cellcolor{red}	\\
    societal   & \cellcolor{blue}  &  \cellcolor{blue}  & \cellcolor{red} & \cellcolor{blue} & \cellcolor{blue} & \cellcolor{blue} & \cellcolor{red}  & \cellcolor{red}  & \cellcolor{blue}  & \cellcolor{red}  &  \cellcolor{red} \\
    \bottomrule
  \end{tabular}
\end{table}
\vspace{-6pt}
\section{Methodology and Results}\label{sec: results}
\vspace{-6pt}
\textbf{Methodology}. 
We perform a series of data analysis steps to measure various KG properties and structural dimensions across all datasets.
Below we provide details on each KG dimension we considered, describe the experiments we conducted along each dimension and summarize empirical observations and findings. Unless otherwise noted, we used the datasets in their entirety (incl. train, test, and validation splits).

\textbf{Entities, relations and triples}. In Fig.  \ref{fig: relationships} we show the relations between the number of entities, relations and triples in each dataset.
When plotting the number of entities vs. number of relations in Fig.  \ref{fig: relationships} (left), we notice that on average semantic web KGs have more diversity in terms of the entity/relation count, while most biomedical KGs cluster at similar entity/relation count. This implies that regardless of the fact that many of the semantic web datasets are derivatives of each other, as mentioned in Sec. \ref{sec: kgapplication}, they exhibit some diversity along these two dimensions.
Notably, Fig.  \ref{fig: relationships} (center) -- plotting the number of entities vs. number of triples -- shows  the biomedical KGs cluster together  with the exception of the PharmKG8K and UMLS datasetas. In the same panel (top right corner)  we also observe that the largest KG datasets in our study are predominantly semantic web ones. 
Finally, Fig.  \ref{fig: relationships} (right) shows that biomedical datasets exhibit less diversity in terms of the relation vs. triples count, in comparison the semantic web datasets. 

\textbf{Average degree and degree distribution}
 KG entities are connected to each other by directed edges (corresponding to the relations), hence an entity can have outgoing edges (or an out-degree) when it is the head entity in a triple and incoming edges (or an in-degree) when it is the tail entity in a triple. 
 We sum the in-degrees and out-degrees to obtain the entity degrees analogously to the way node degrees are computed in undirected graphs. 
 A smaller average degree indicates that the KG is sparser. 
 Fig. \ref{fig: multiples} (left) shows the degree distribution (over all entitie). 
 The average degree for each KG is also shown in Tab. \ref{tbl: datasetstable}. From this table we notice that the semantic datasets have some of the lowest average degrees across all datasets (e.g., ConceptNet with an average degree of \m{2}. 
 Among the semantic datasets, FB15k has the singularly highest average degree, 
 while its derivative FB15k-237 has a value that aligns with the rest of the semantic datasets.
On the other hand, the average degrees of biomedical  datasets are split into two groups: \m{5} of the biomedical datasets have average degrees in the low \m{100}'s, while PharKG, Globi, OpenBioLink, and BioKG 
 have average degrees that are closer to the values of the semantic datasets. 
The two societal datasets Kinship and Nations demonstrate a significantly higher average degree than the rest of the KGs 
(see the Appx.) 
 In Fig.  \ref{fig: multiples} (left) several datasets show a distinct degree distribution in the lower degrees. 
 ConceptNet, CSKG have the highest number of low-degree nodes, while the degree distributions of Wikidata5M and OGBWikiKG2 are not as smooth, with oscillations at degrees \m{3} and \m{5}, respectively.

\textbf{KG Density}. 
The KG density is computed as the ratio \m{|E|/|R|^2} and like in homegeneous graphs higher density implies more sparsity 
(see Tab. \ref{tbl: datasetstable}, last column). The degree trends we described in the previous paragraph can also be traced along this dimension. 
Related to the KG density, we also quantify the KG connectivity by plotting the triples per KG relation in Fig. \ref{fig: multiples} (center). 
The thick tails in the plot show that AristoV4, followed by FB15k, have a high number of relations with a low number triples per relation.
Several other semantic datasets follow the same pattern, while \textit{none} of the biomedical datasets do, with the exception of DRKG.

\begin{figure}[t]
     \centering
     \begin{minipage}{0.24\textwidth}
         \includegraphics[width=\textwidth]{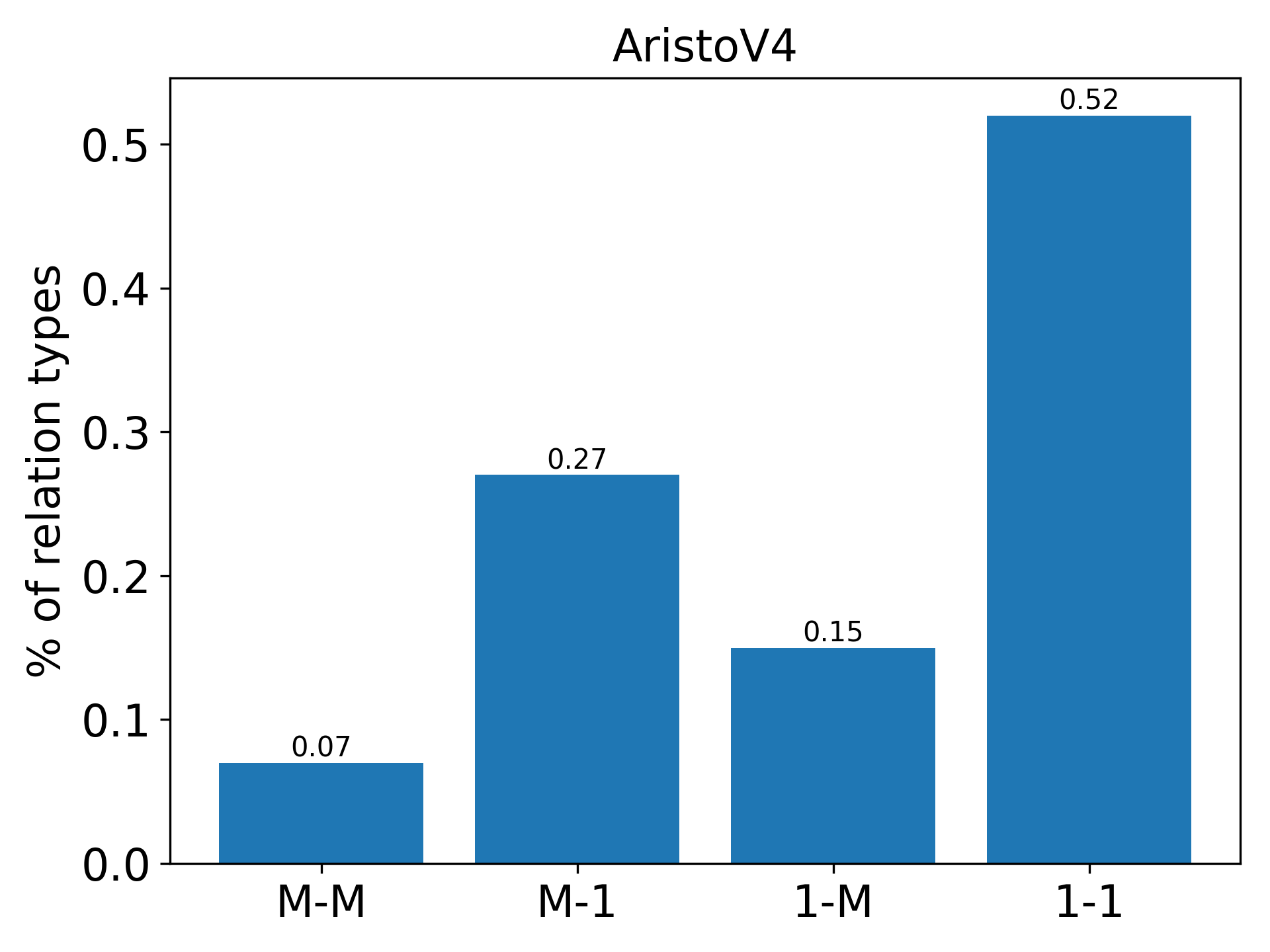}
     \end{minipage}
     \begin{minipage}{0.24\textwidth}
         \includegraphics[width=\textwidth]{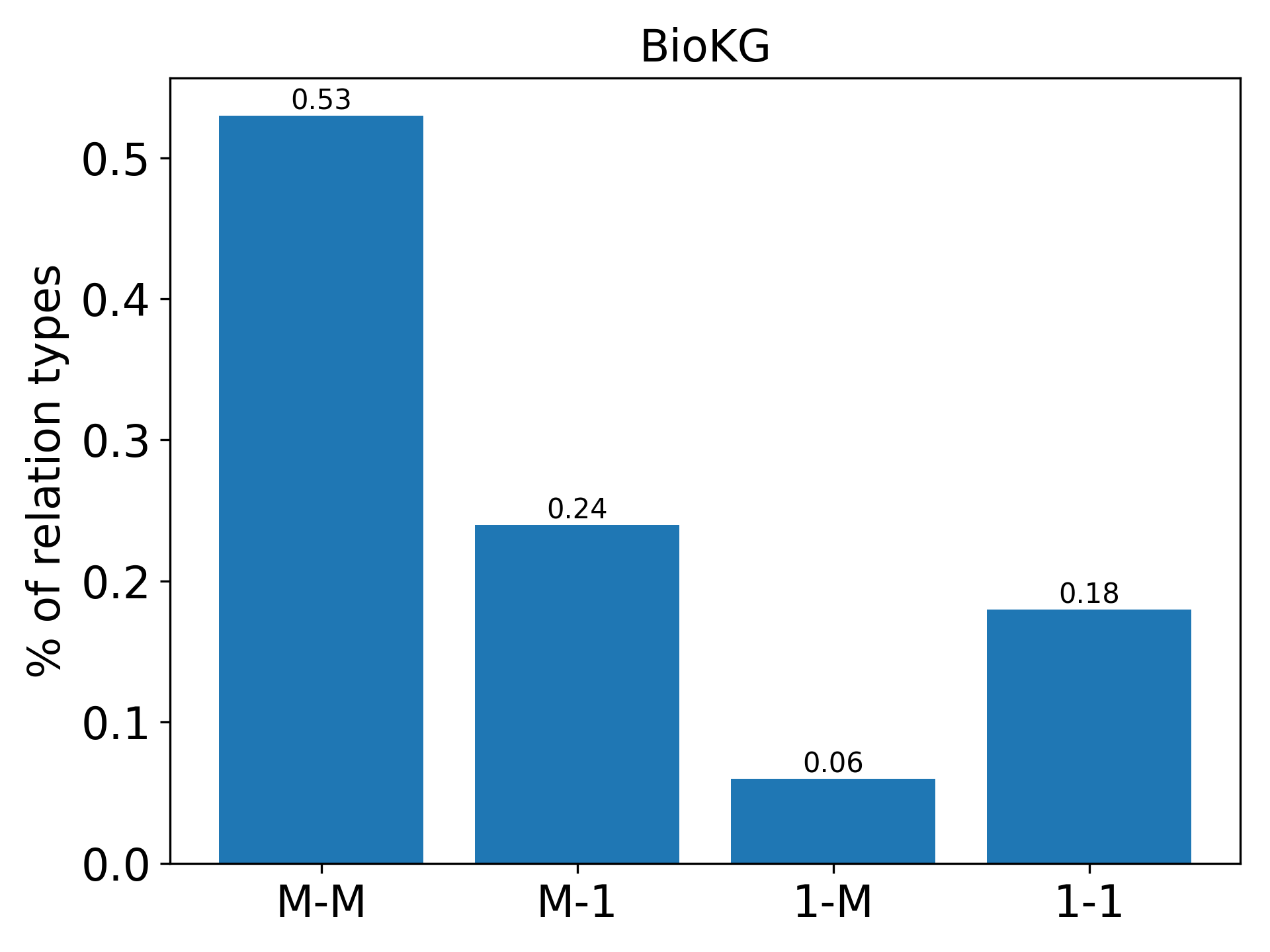}
    \end{minipage}
    \begin{minipage}{0.24\textwidth}
         \includegraphics[width=\textwidth]{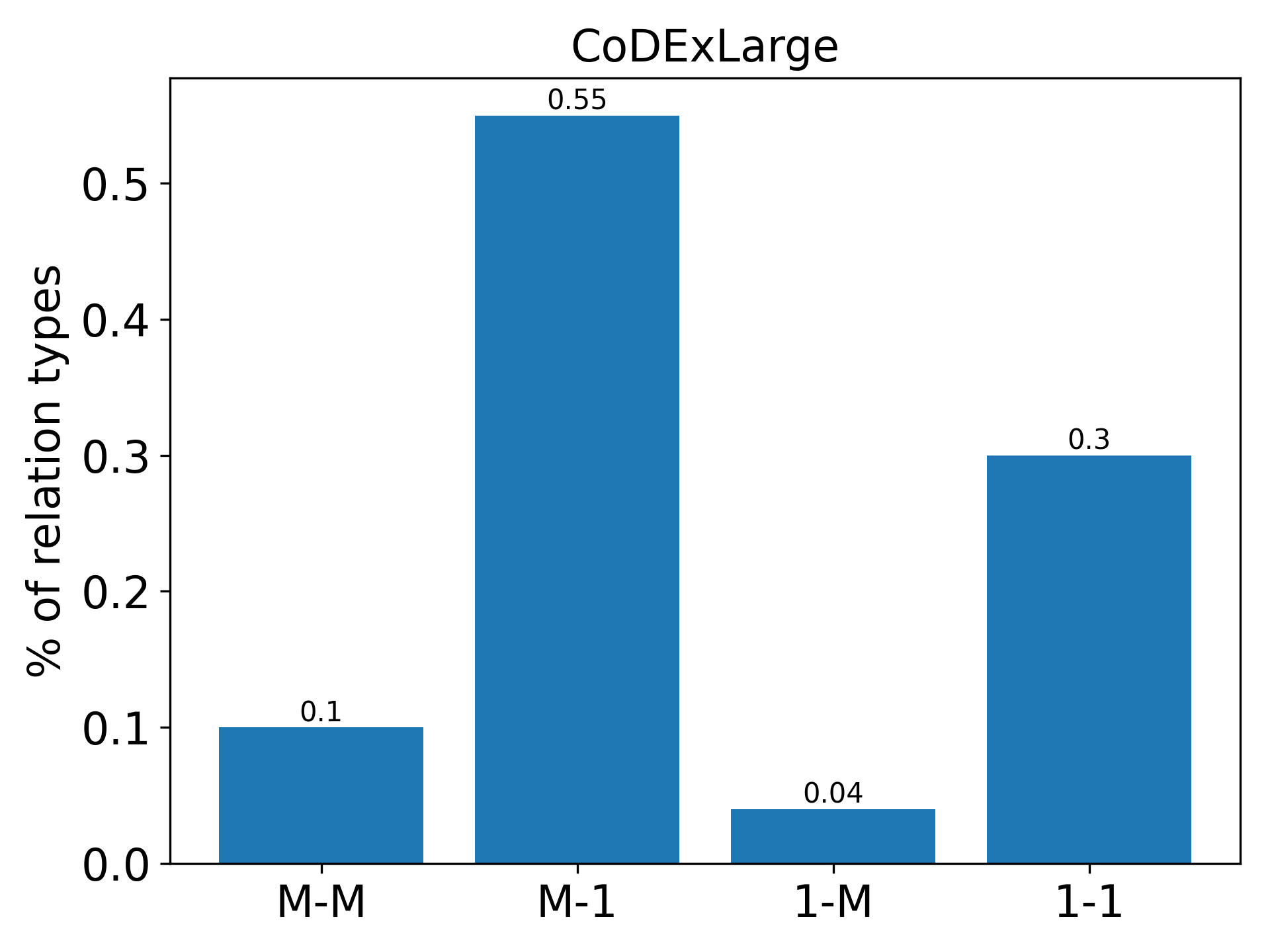}
    \end{minipage}
     \begin{minipage}{0.24\textwidth}
         \includegraphics[width=\textwidth]{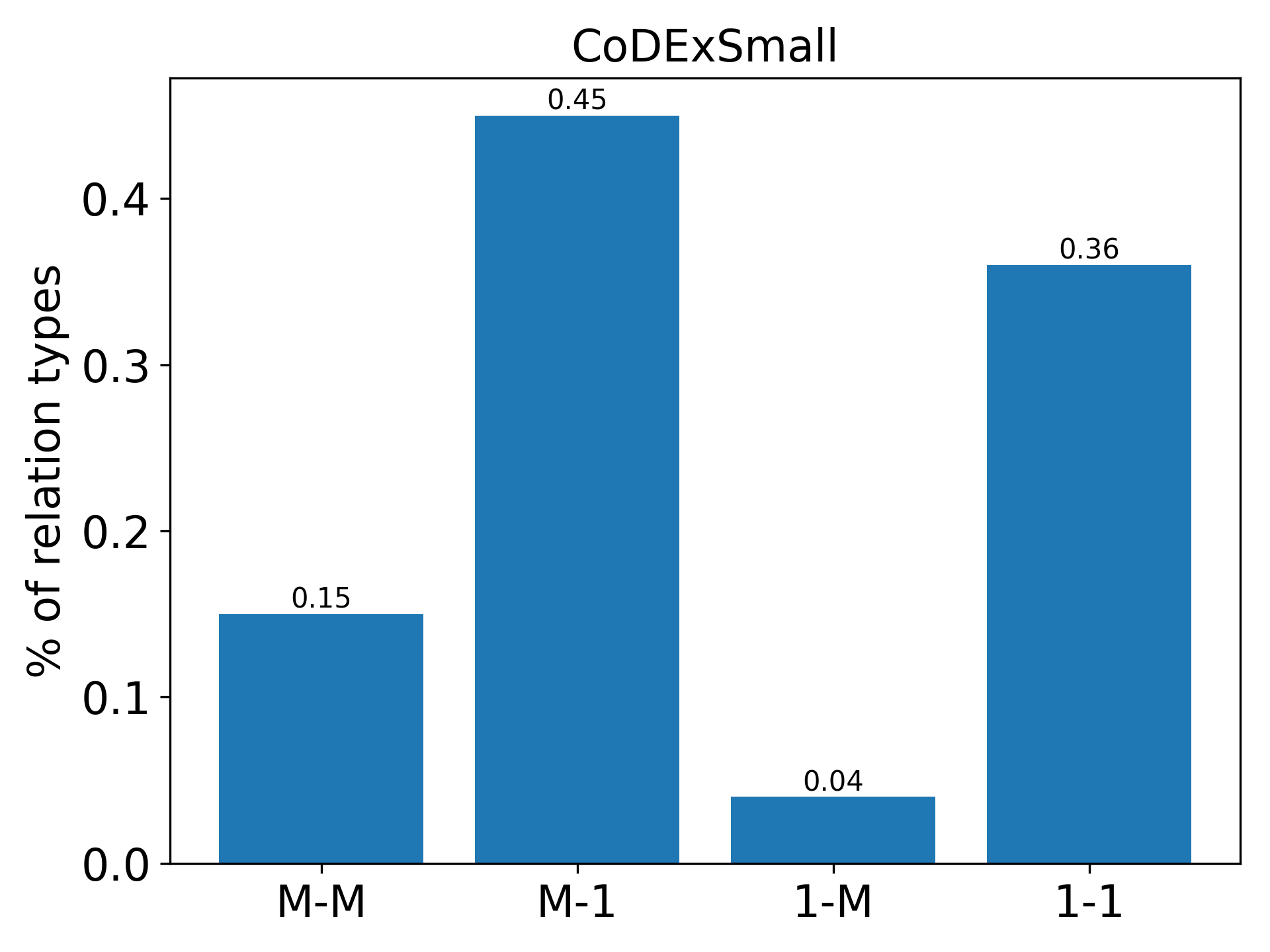 }
    \end{minipage}
    
    \begin{minipage}{0.24\textwidth}
         \includegraphics[width=\textwidth]{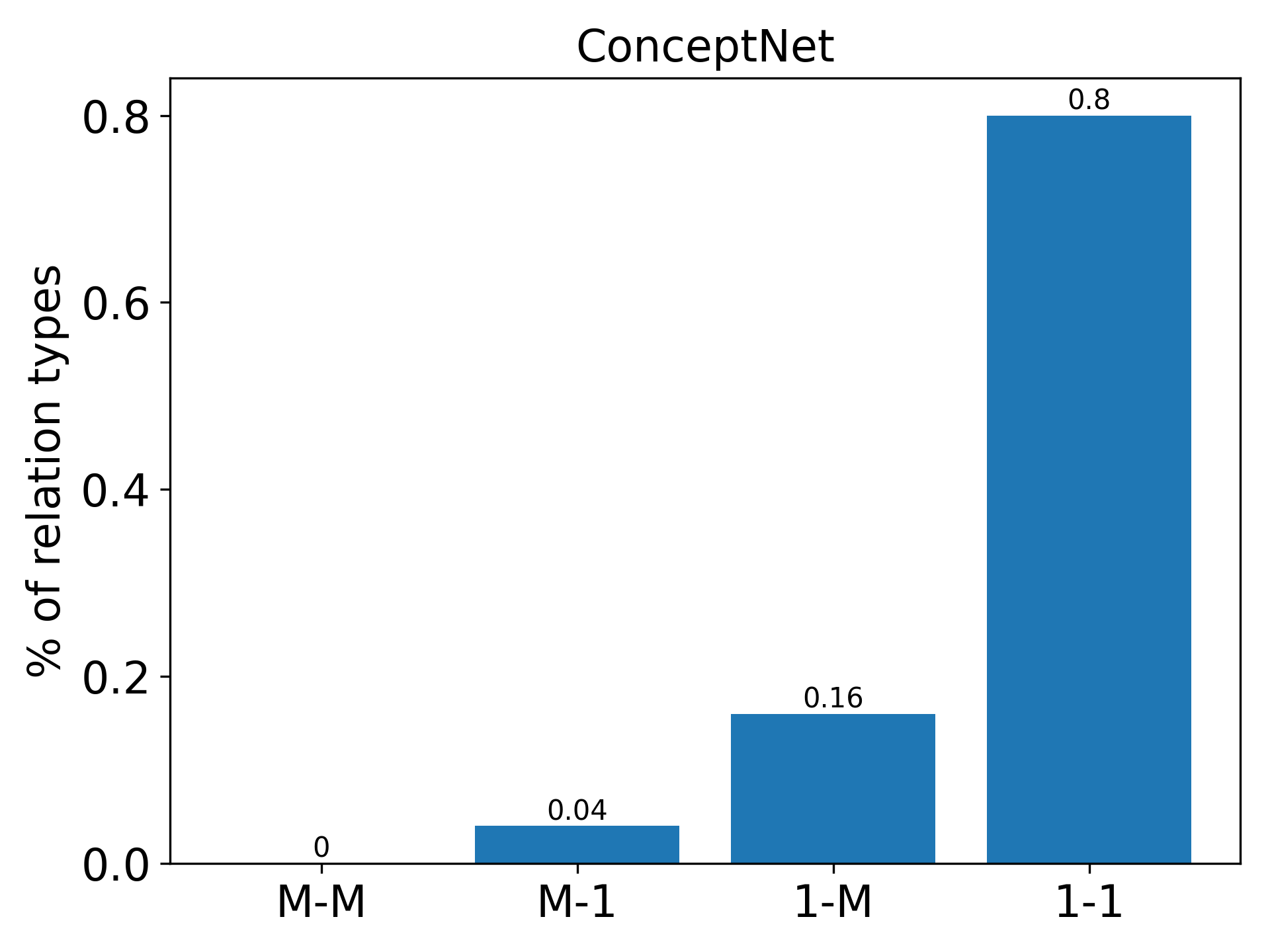 }
    \end{minipage}
     \begin{minipage}{0.24\textwidth}
         \includegraphics[width=\textwidth]{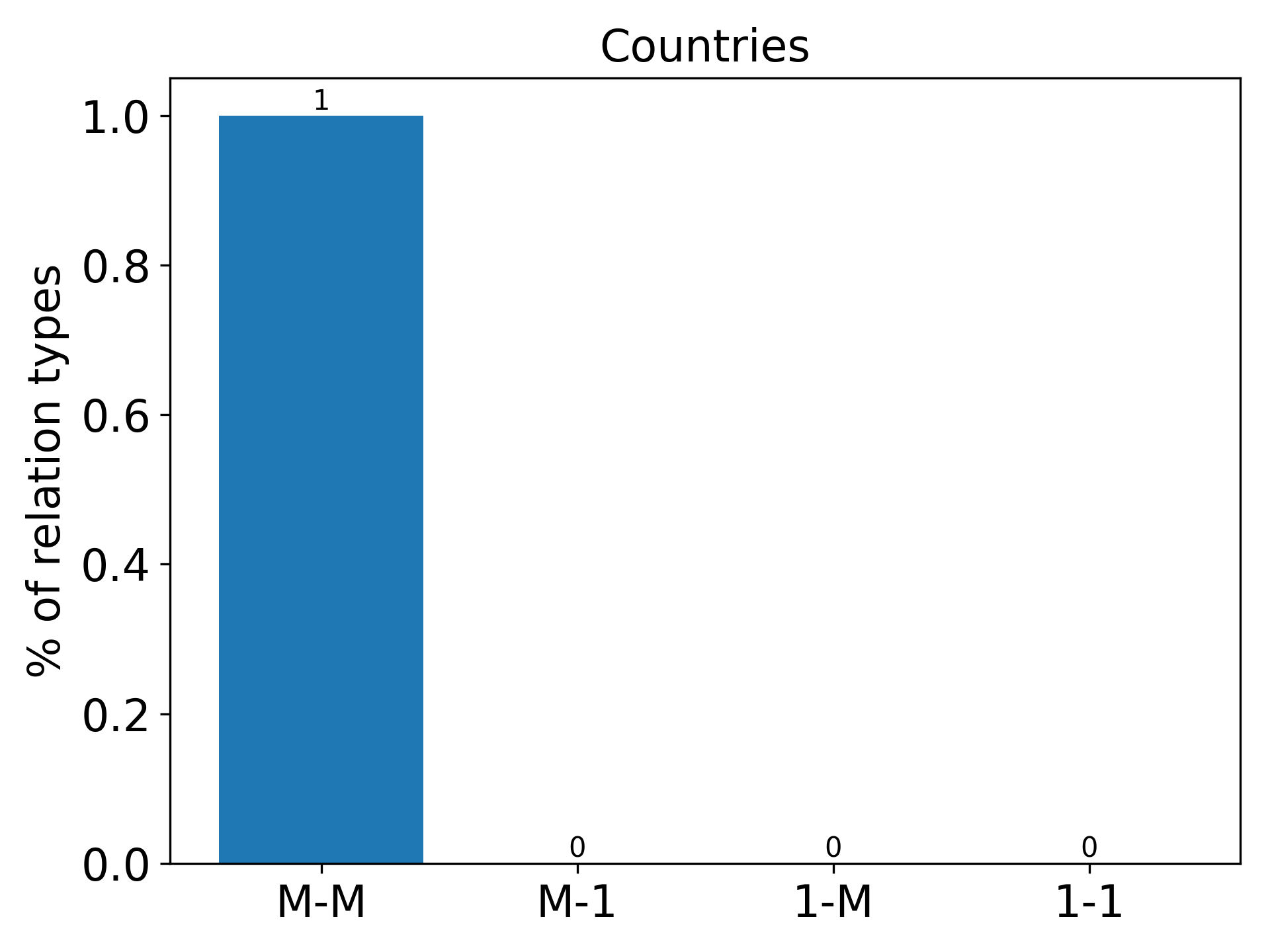 }
     \end{minipage}
     \begin{minipage}{0.24\textwidth}
         \includegraphics[width=\textwidth]{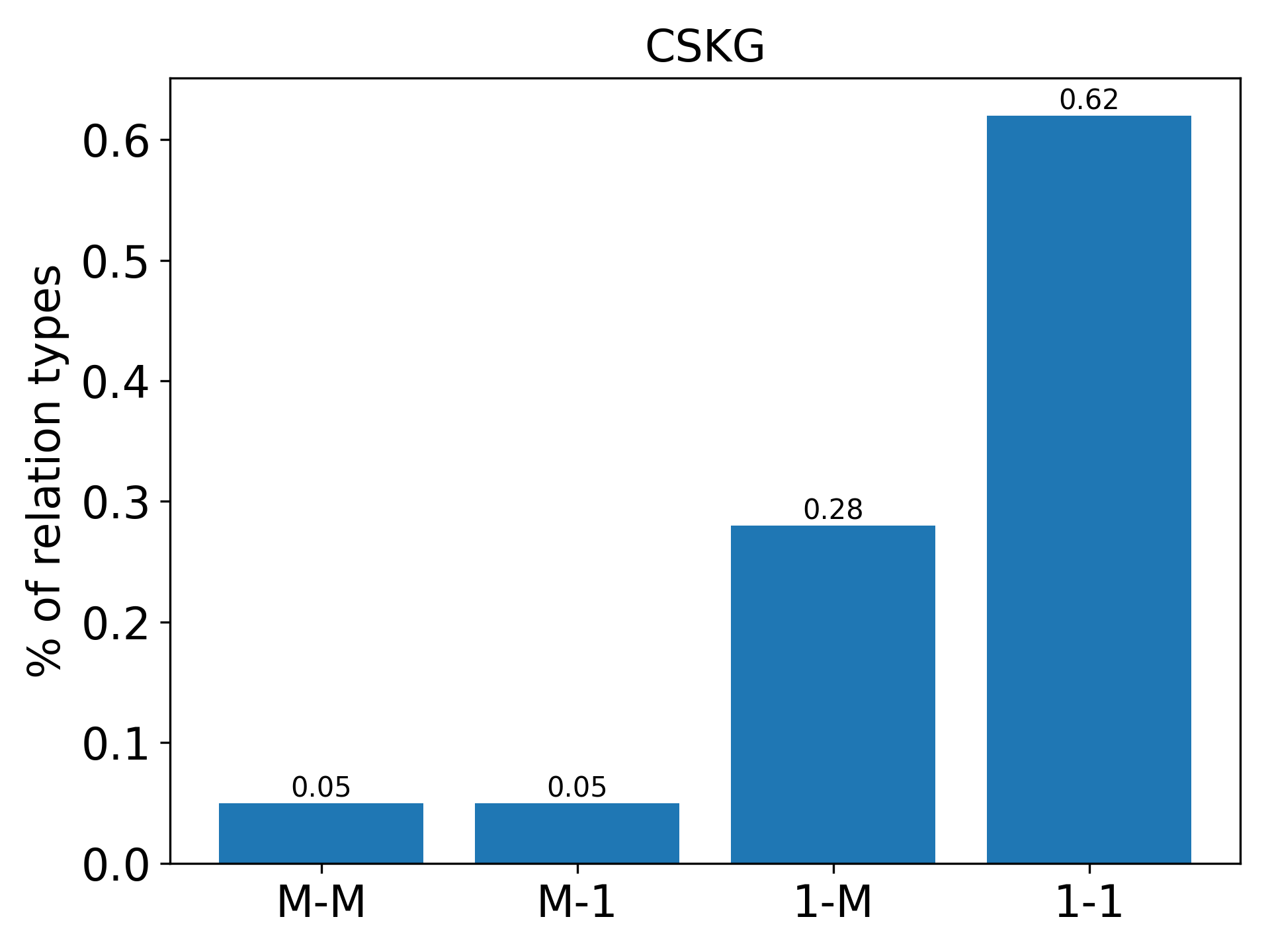 }
    \end{minipage}
    \begin{minipage}{0.24\textwidth}
         \includegraphics[width=\textwidth]{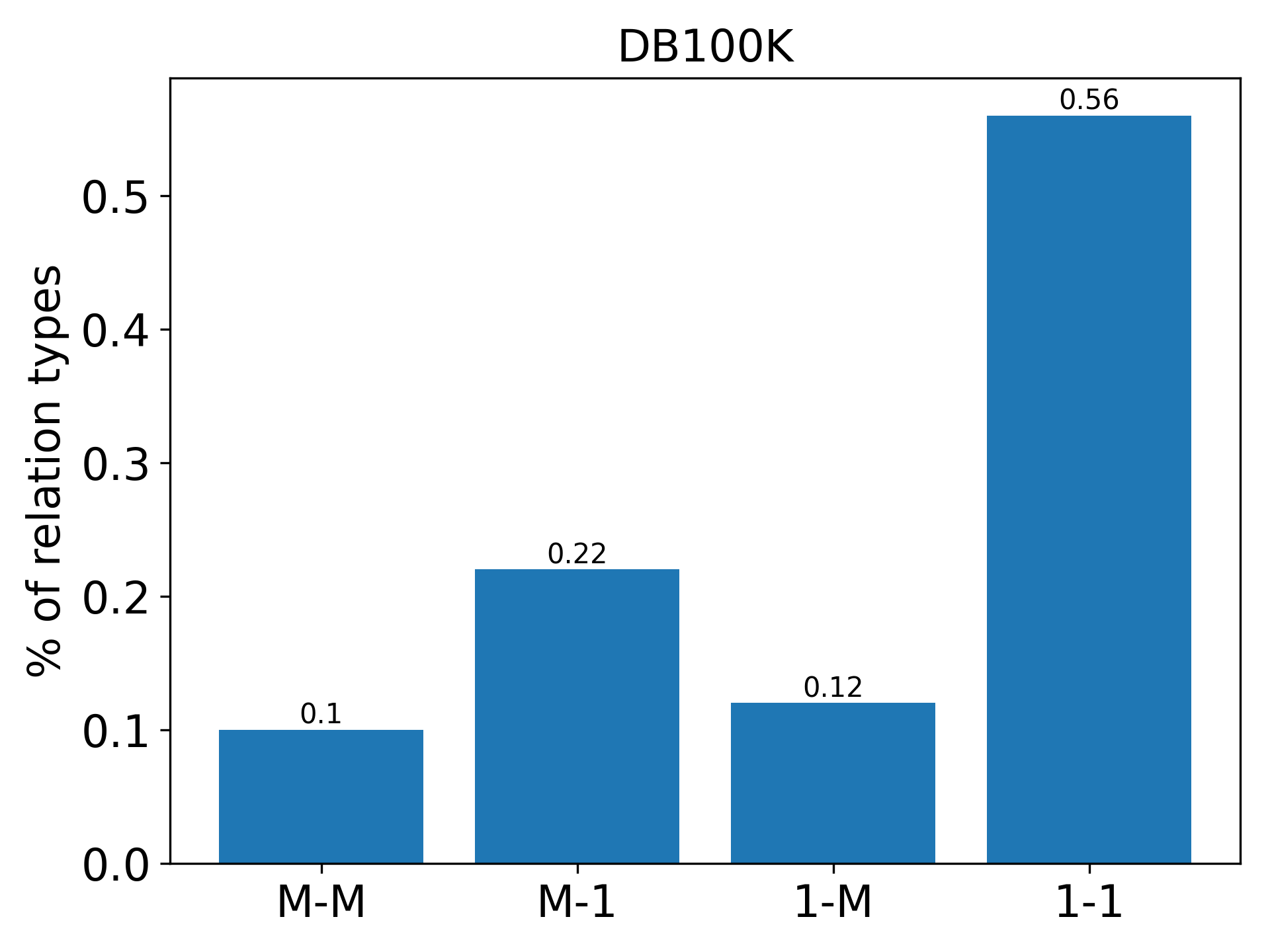 }
    \end{minipage}
    
     \begin{minipage}{0.24\textwidth}
         \includegraphics[width=\textwidth]{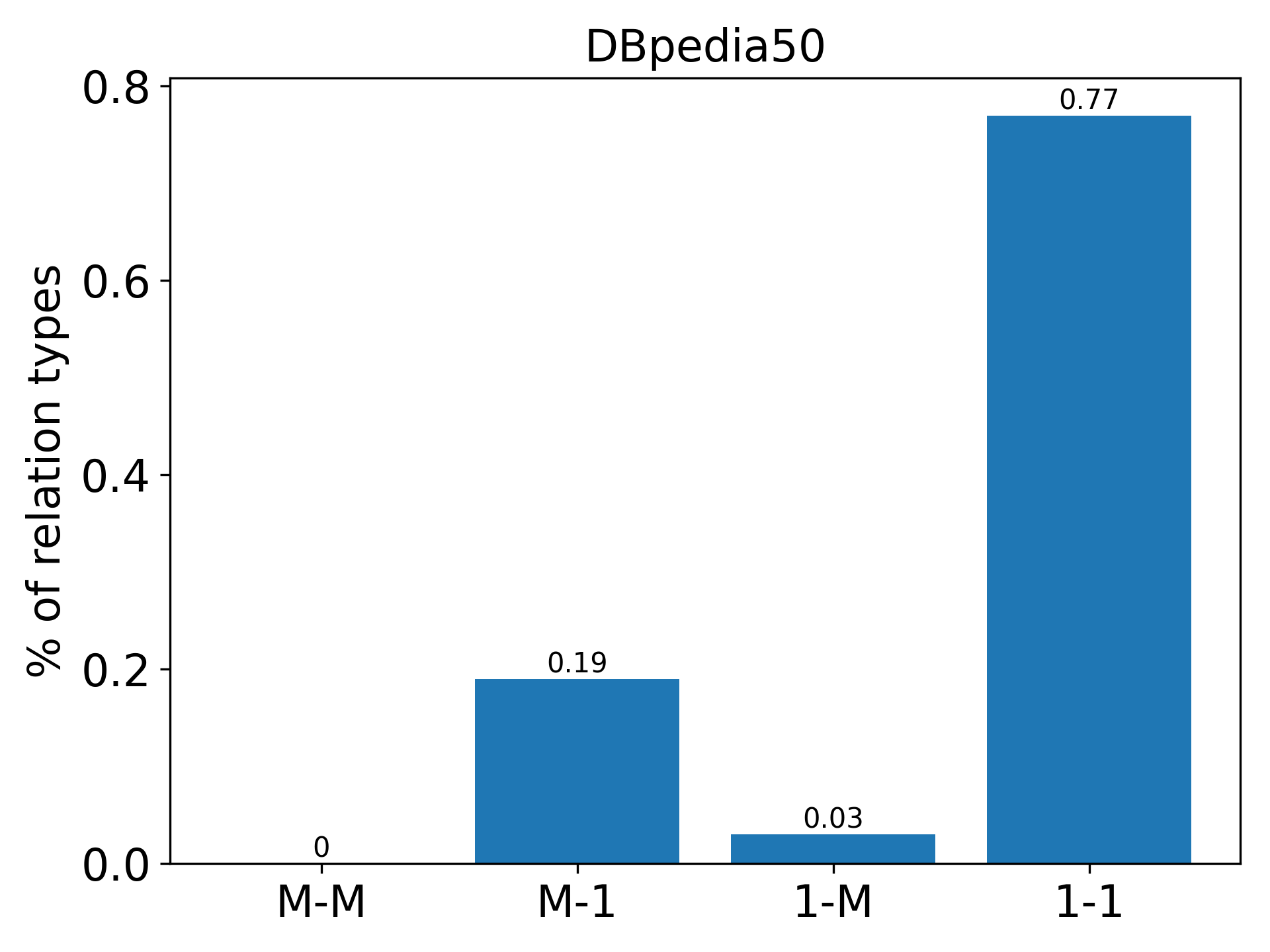 }
     \end{minipage}
     \begin{minipage}{0.24\textwidth}
         \includegraphics[width=\textwidth]{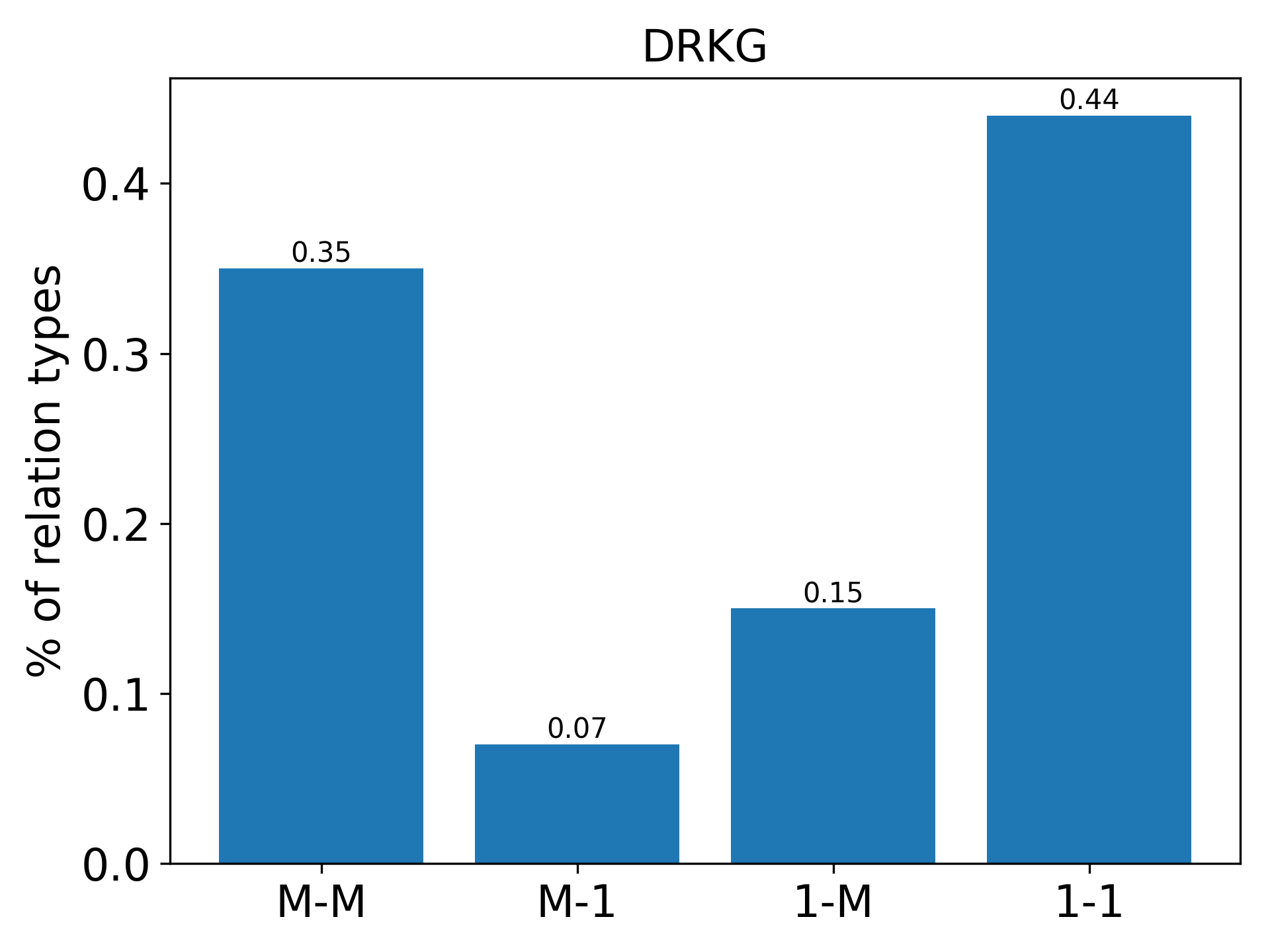 }
    \end{minipage}
    \begin{minipage}{0.24\textwidth}
         \includegraphics[width=\textwidth]{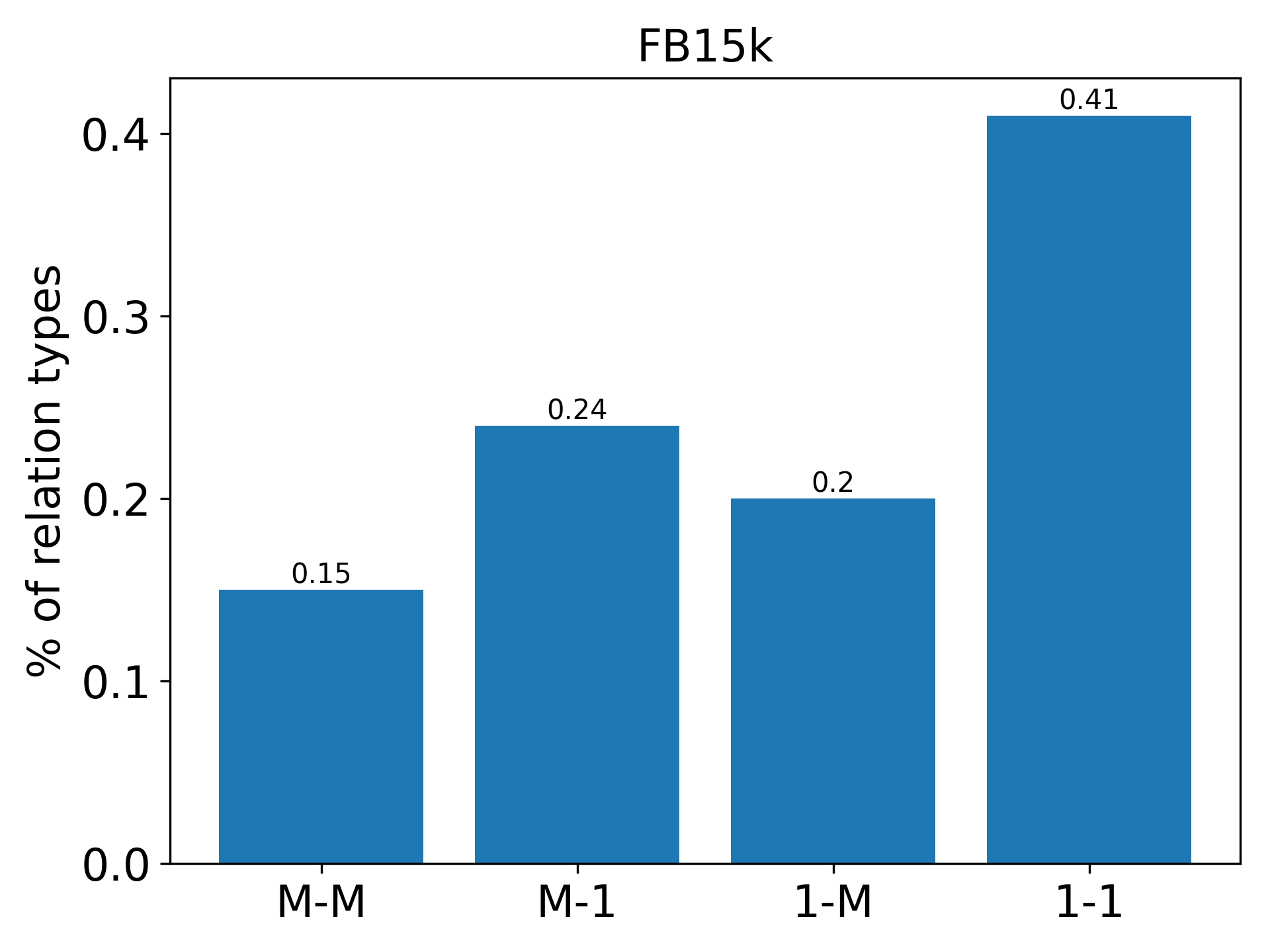 }
    \end{minipage}
     \begin{minipage}{0.24\textwidth}
         \includegraphics[width=\textwidth]{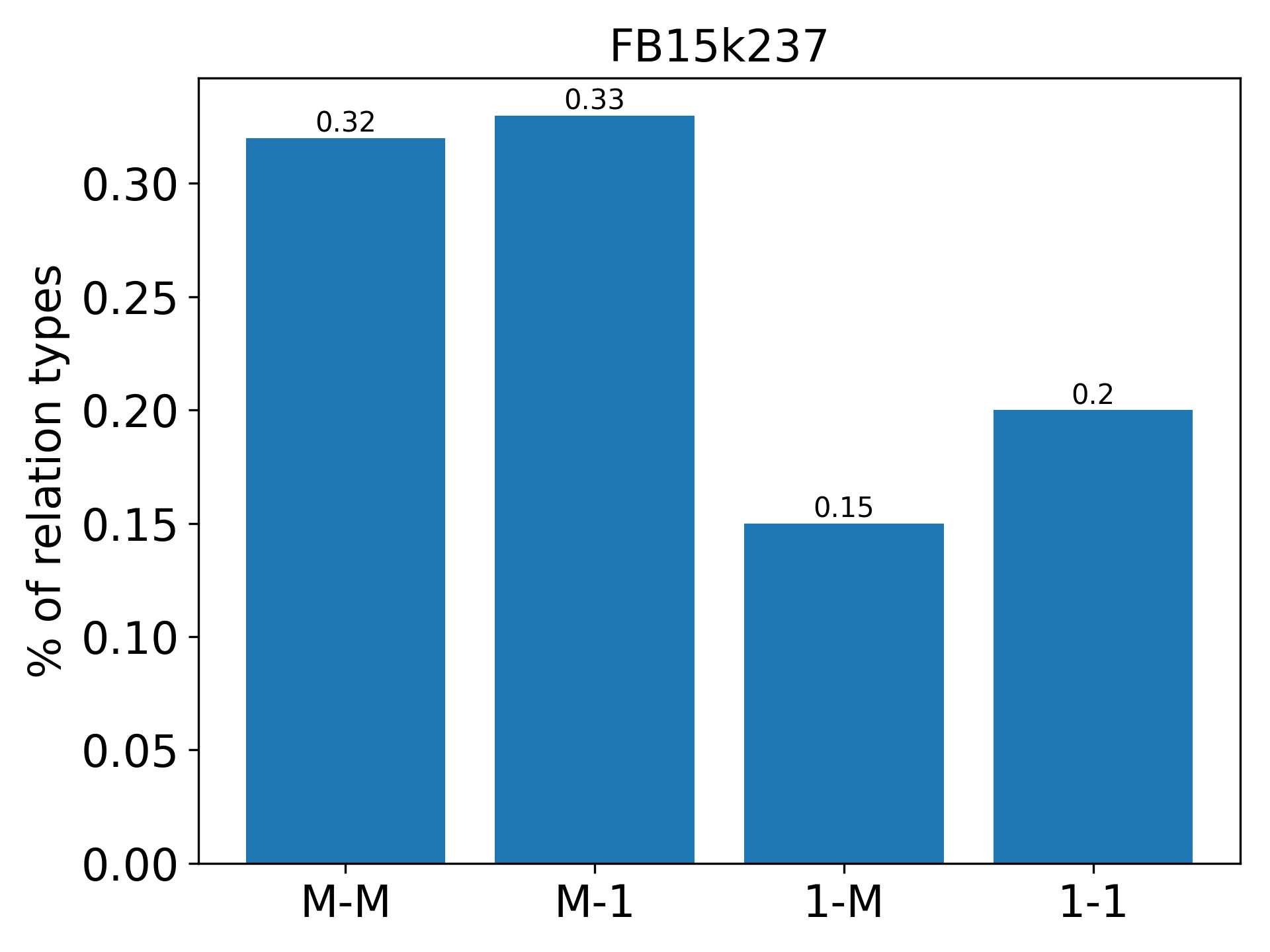 }
     \end{minipage}
     
     \begin{minipage}{0.24\textwidth}
         \includegraphics[width=\textwidth]{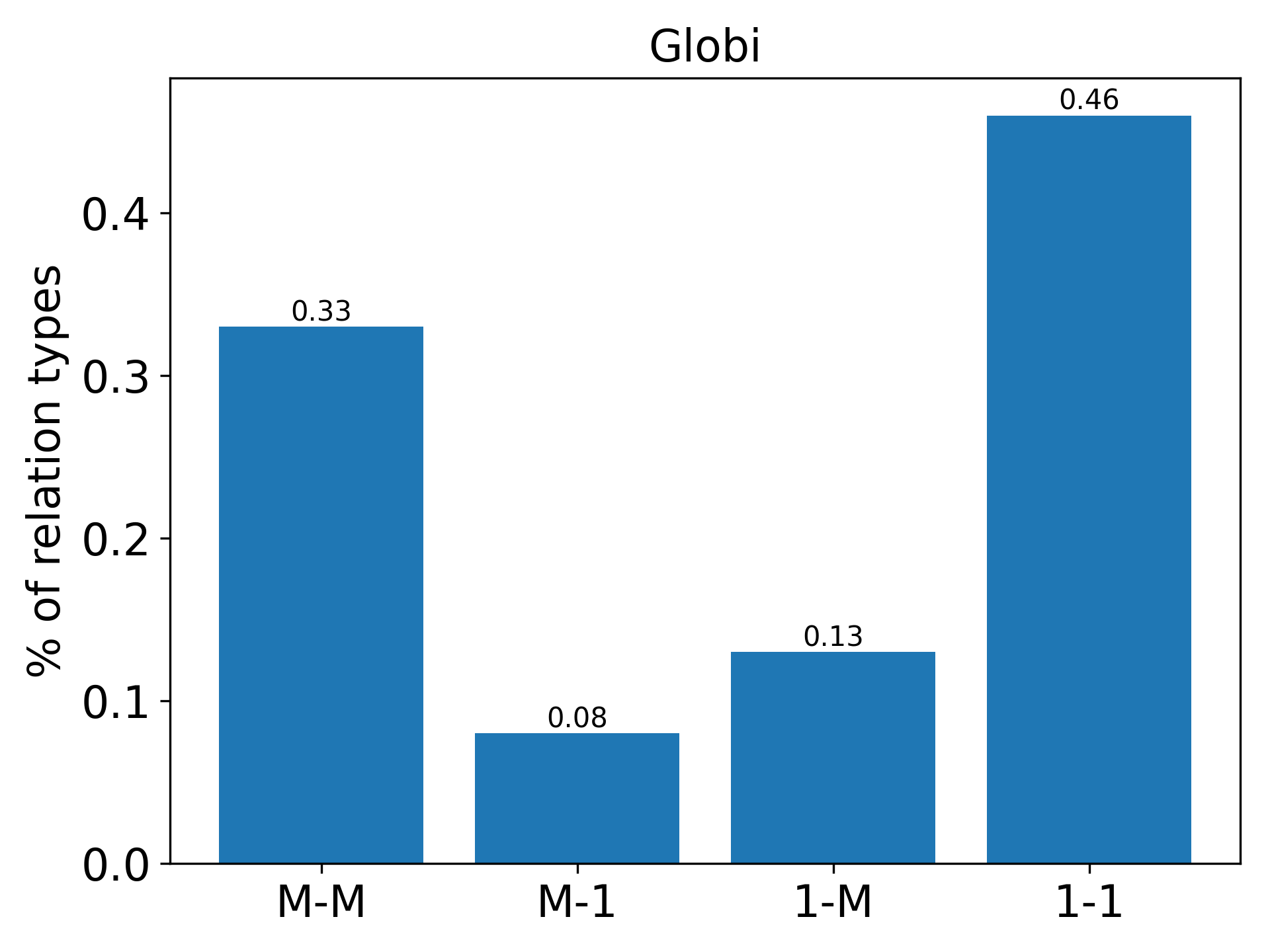 }
    \end{minipage}
    \begin{minipage}{0.24\textwidth}
         \includegraphics[width=\textwidth]{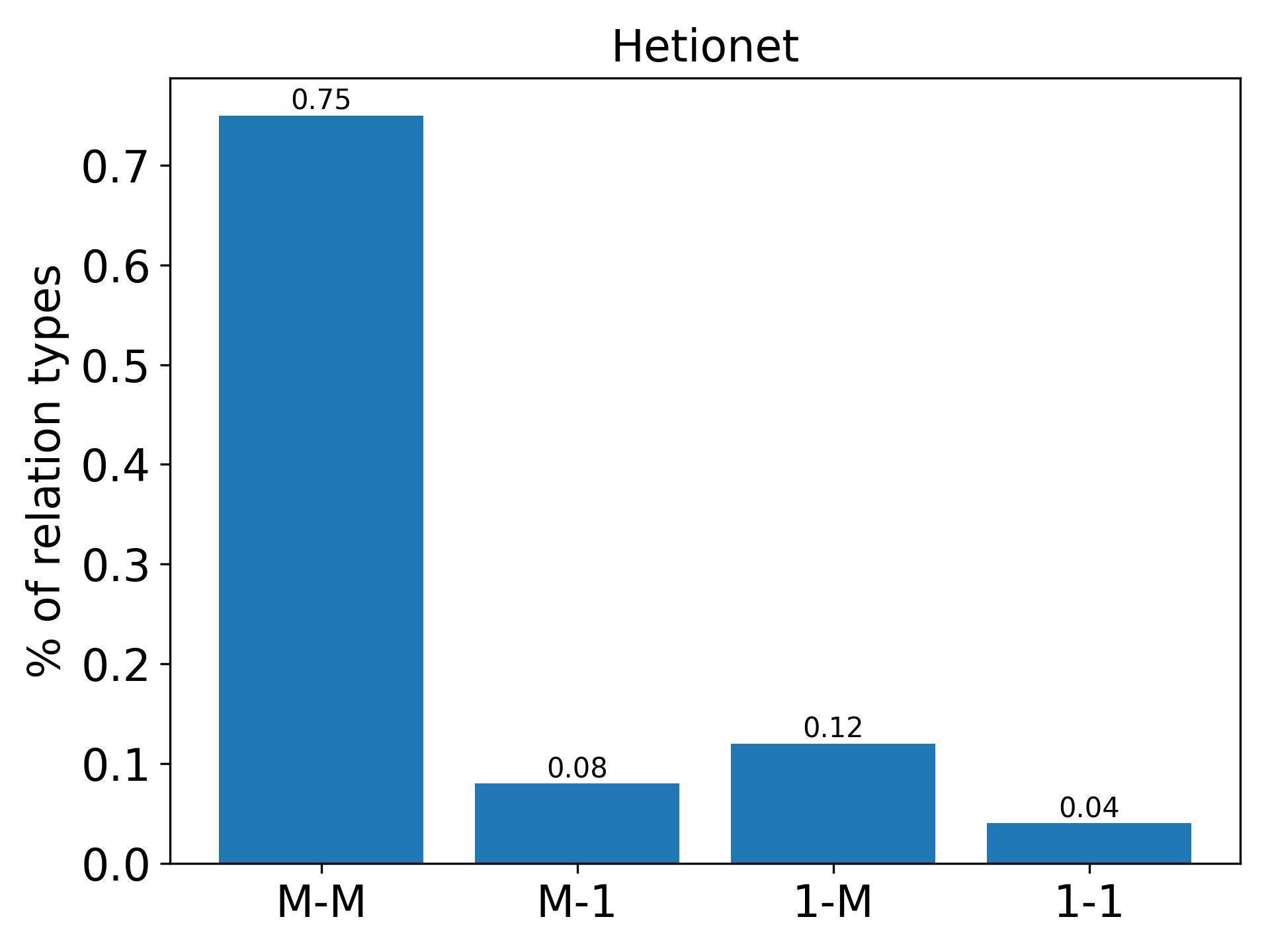 }
    \end{minipage}
     \begin{minipage}{0.24\textwidth}
         \includegraphics[width=\textwidth]{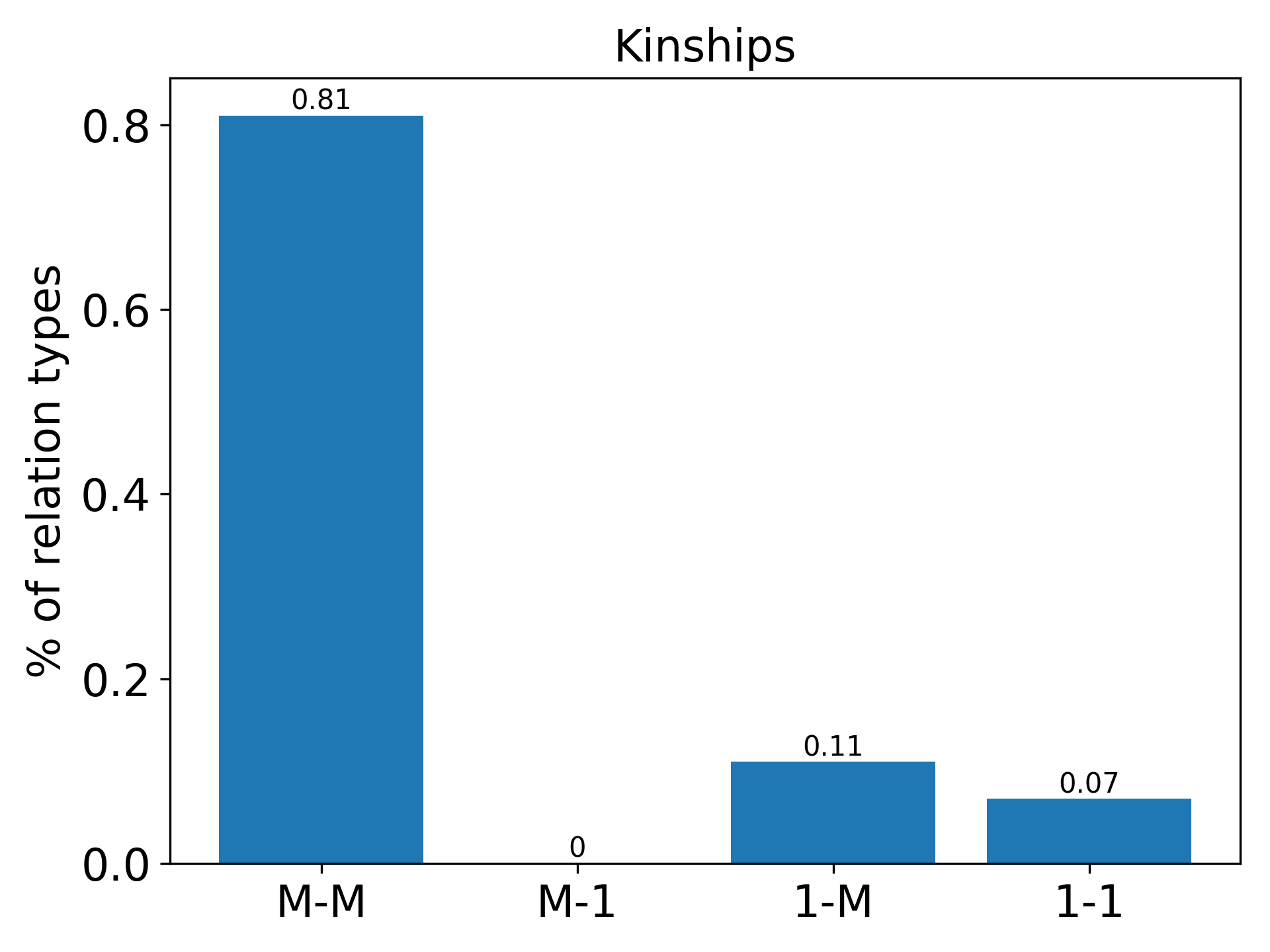 }
     \end{minipage}
     \begin{minipage}{0.24\textwidth}
         \includegraphics[width=\textwidth]{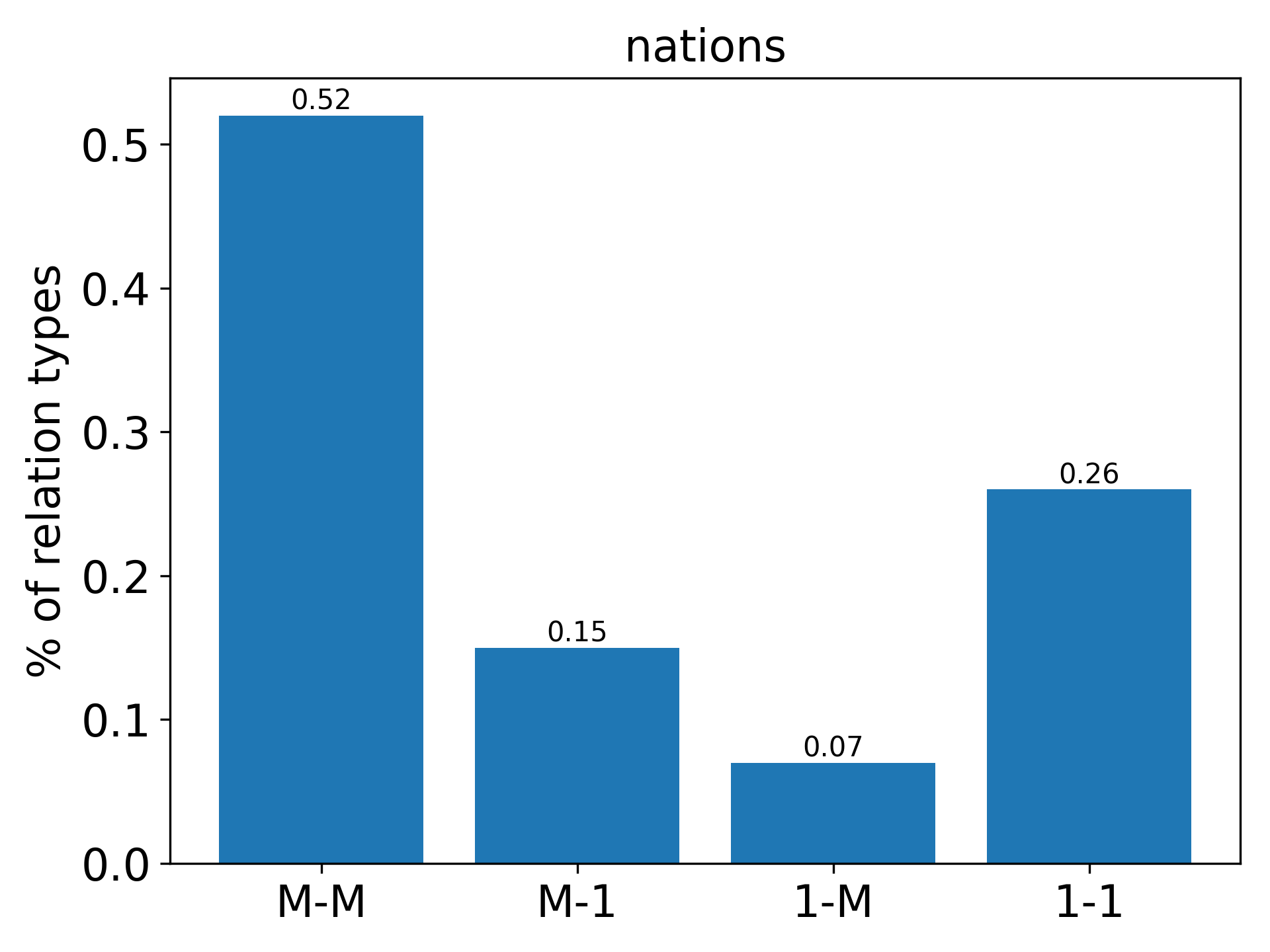}
    \end{minipage}

    \begin{minipage}{0.24\textwidth}
         \includegraphics[width=\textwidth]{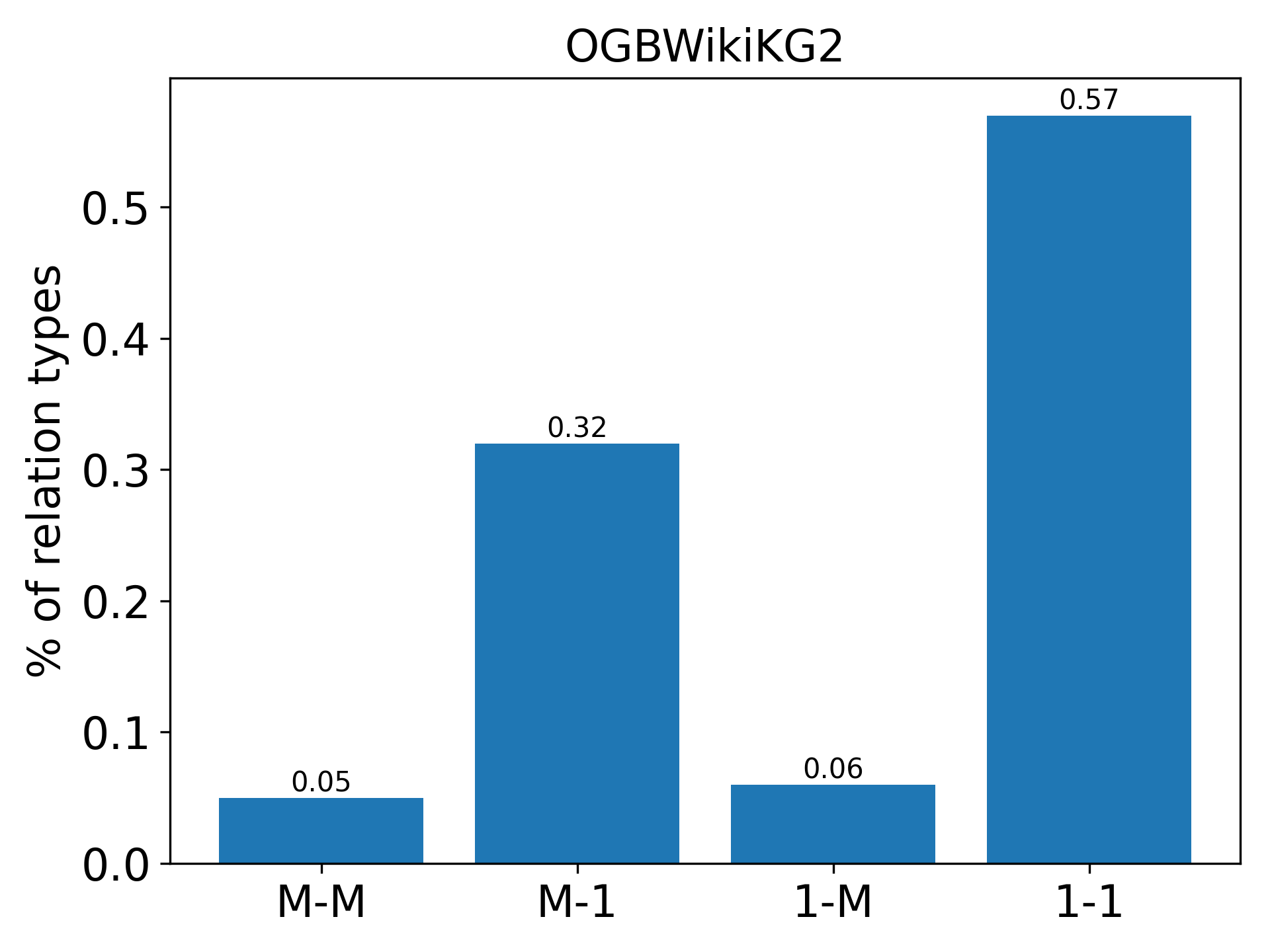 }
    \end{minipage}    
    \begin{minipage}{0.24\textwidth}
         \includegraphics[width=\textwidth]{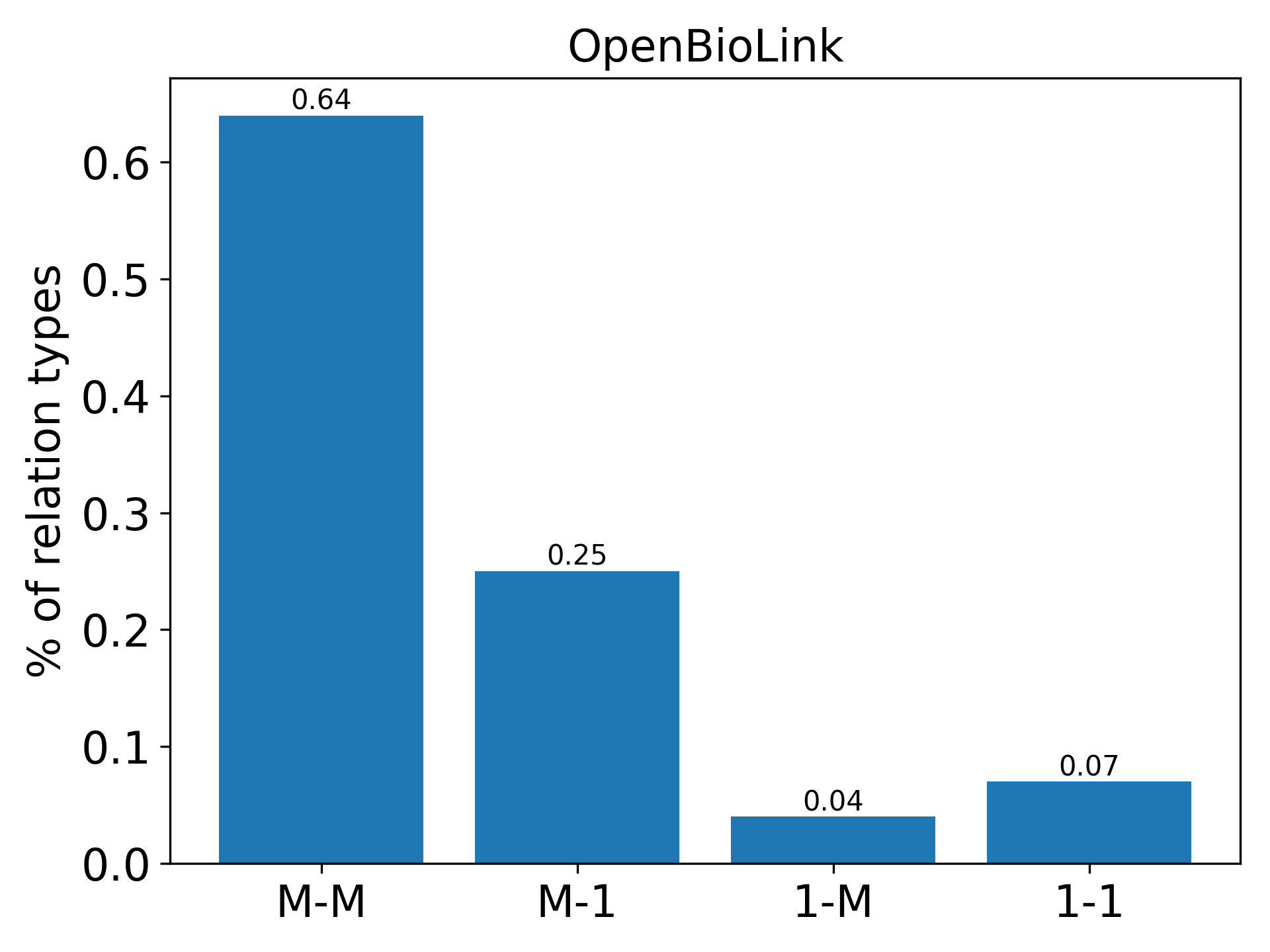 }
    \end{minipage}    
     \begin{minipage}{0.24\textwidth}
         \includegraphics[width=\textwidth]{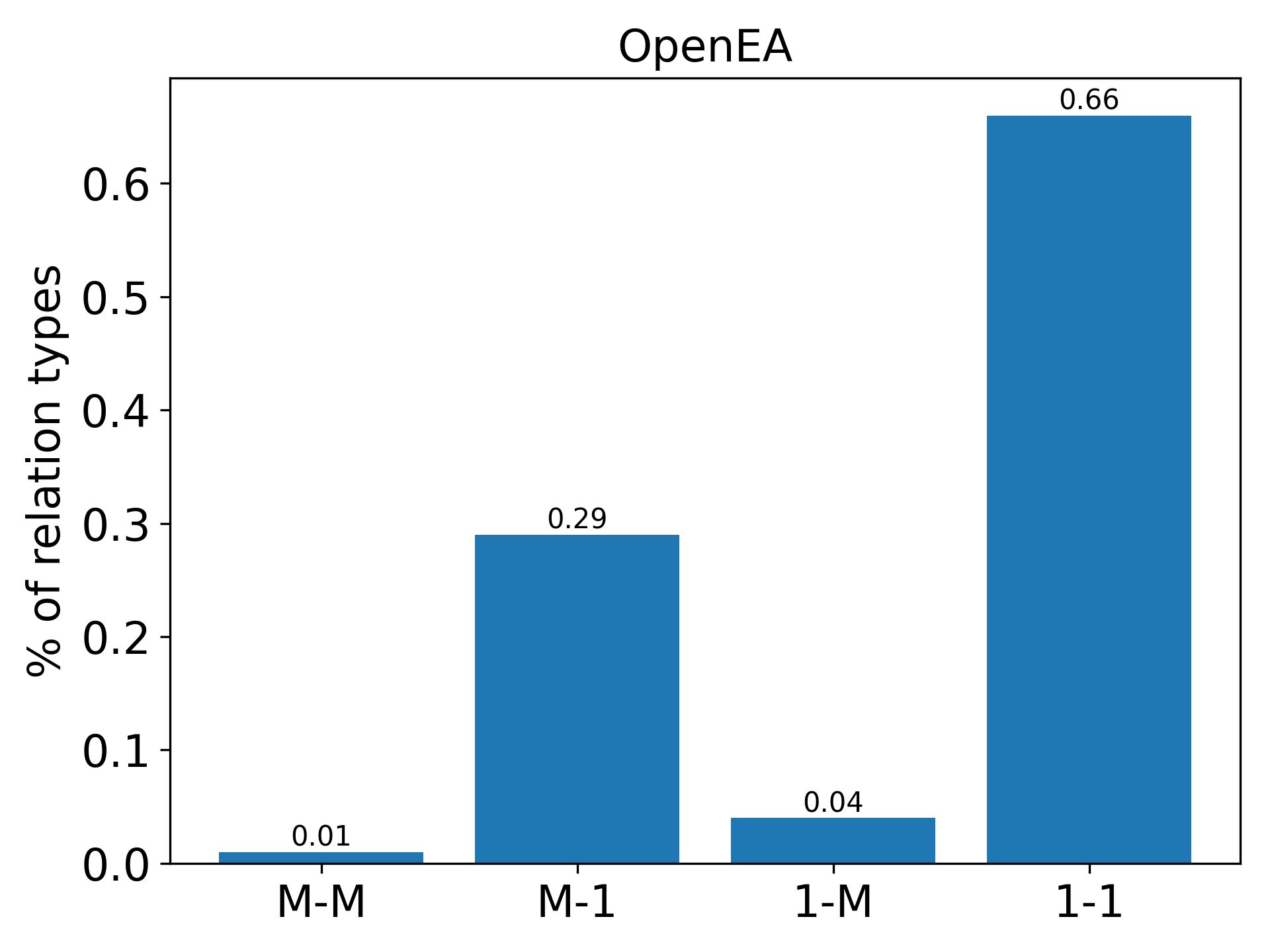 }
     \end{minipage}
     \begin{minipage}{0.24\textwidth}
         \includegraphics[width=\textwidth]{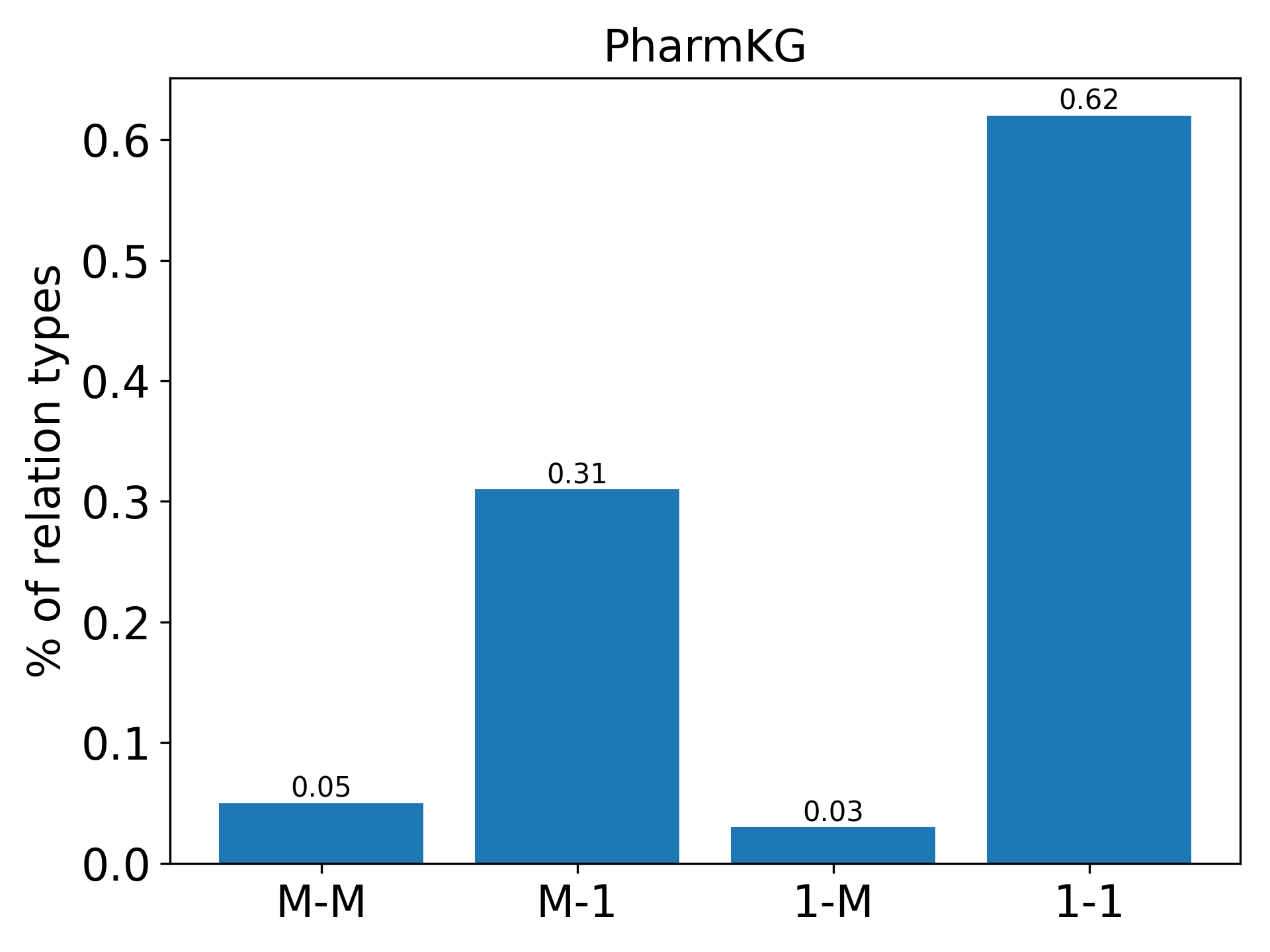 }
    \end{minipage}
    
    \begin{minipage}{0.24\textwidth}
         \includegraphics[width=\textwidth]{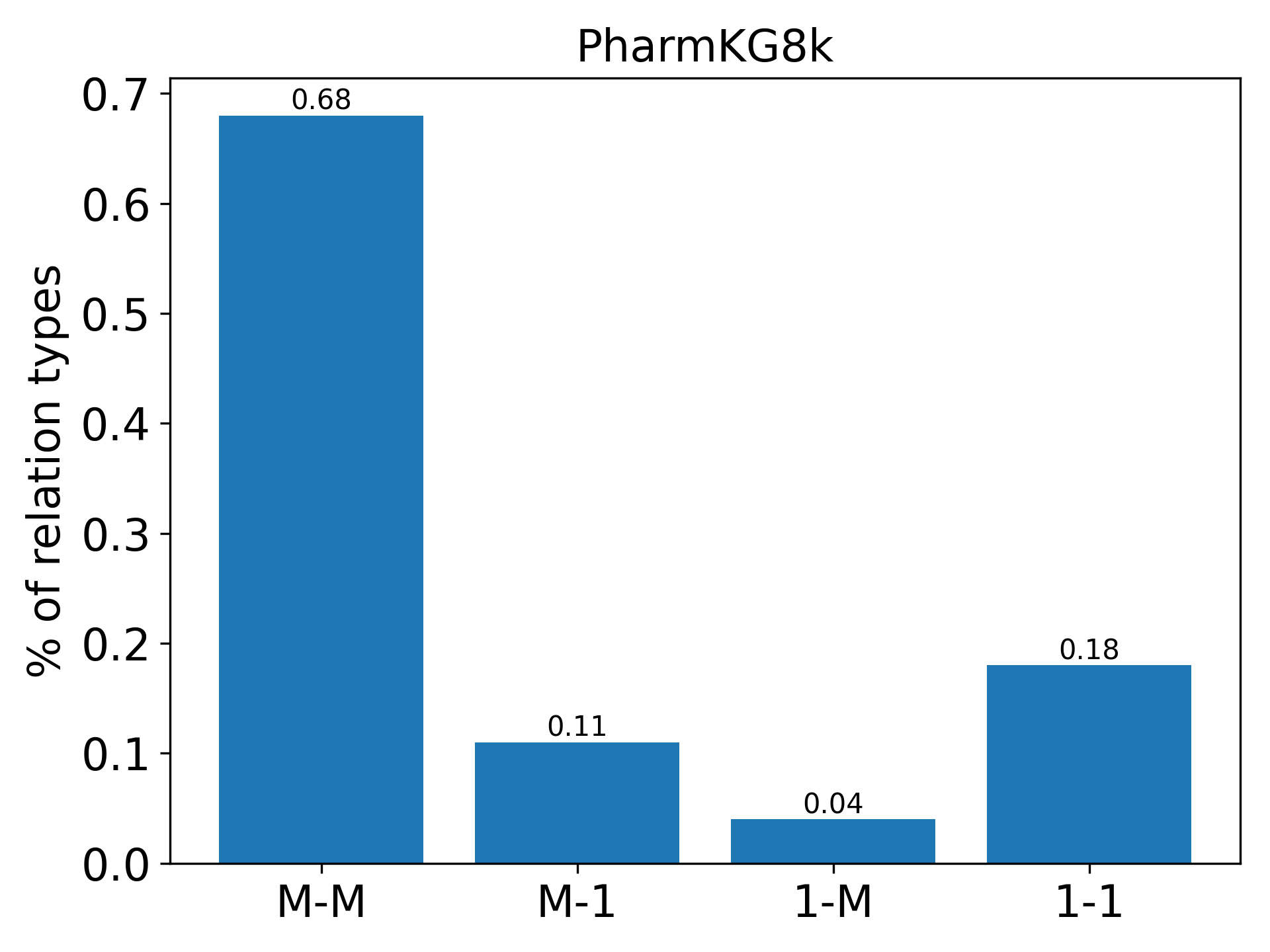 }
    \end{minipage}    
     \begin{minipage}{0.24\textwidth}
         \includegraphics[width=\textwidth]{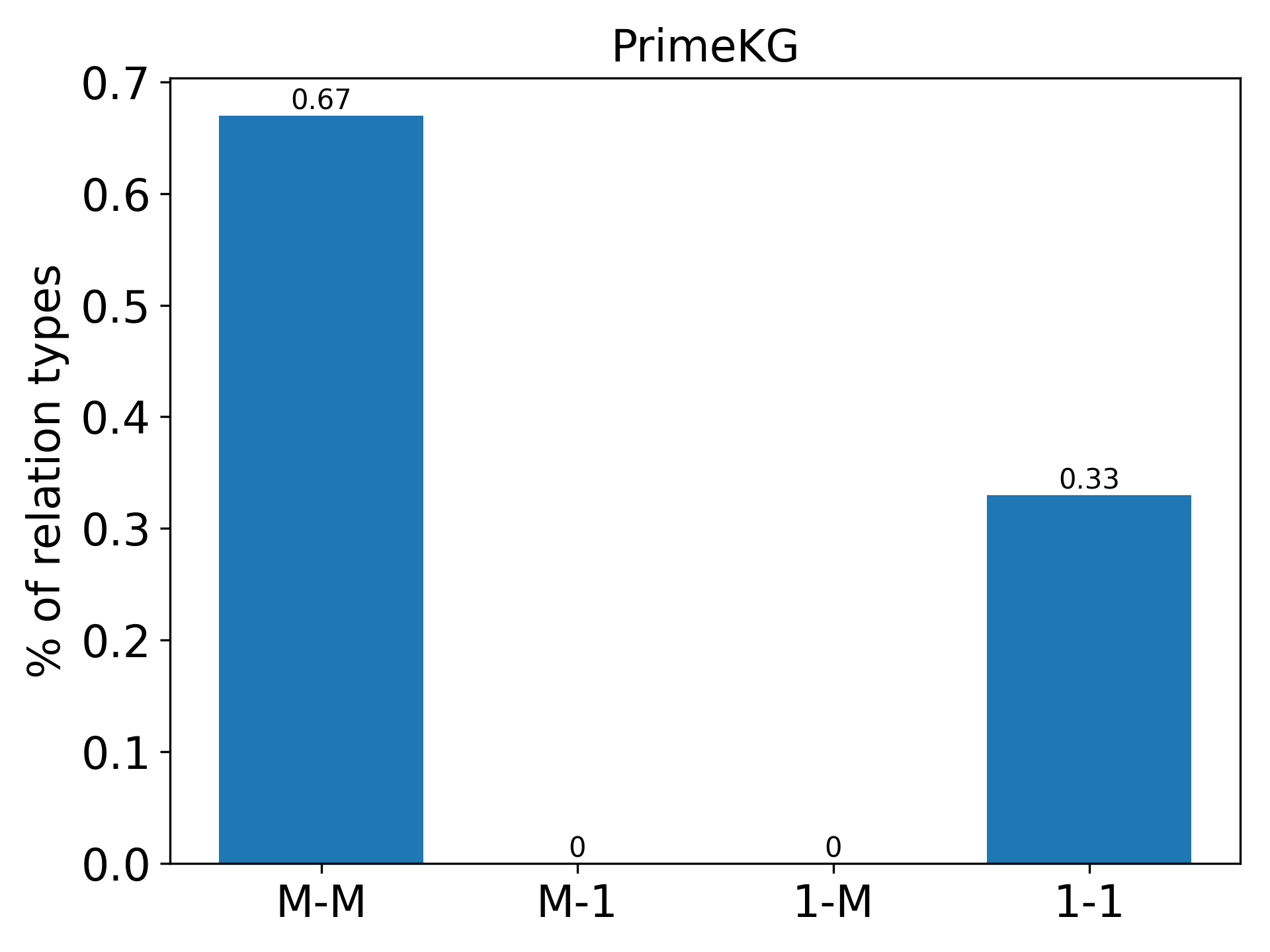 }
     \end{minipage}
     \begin{minipage}{0.24\textwidth}
         \includegraphics[width=\textwidth]{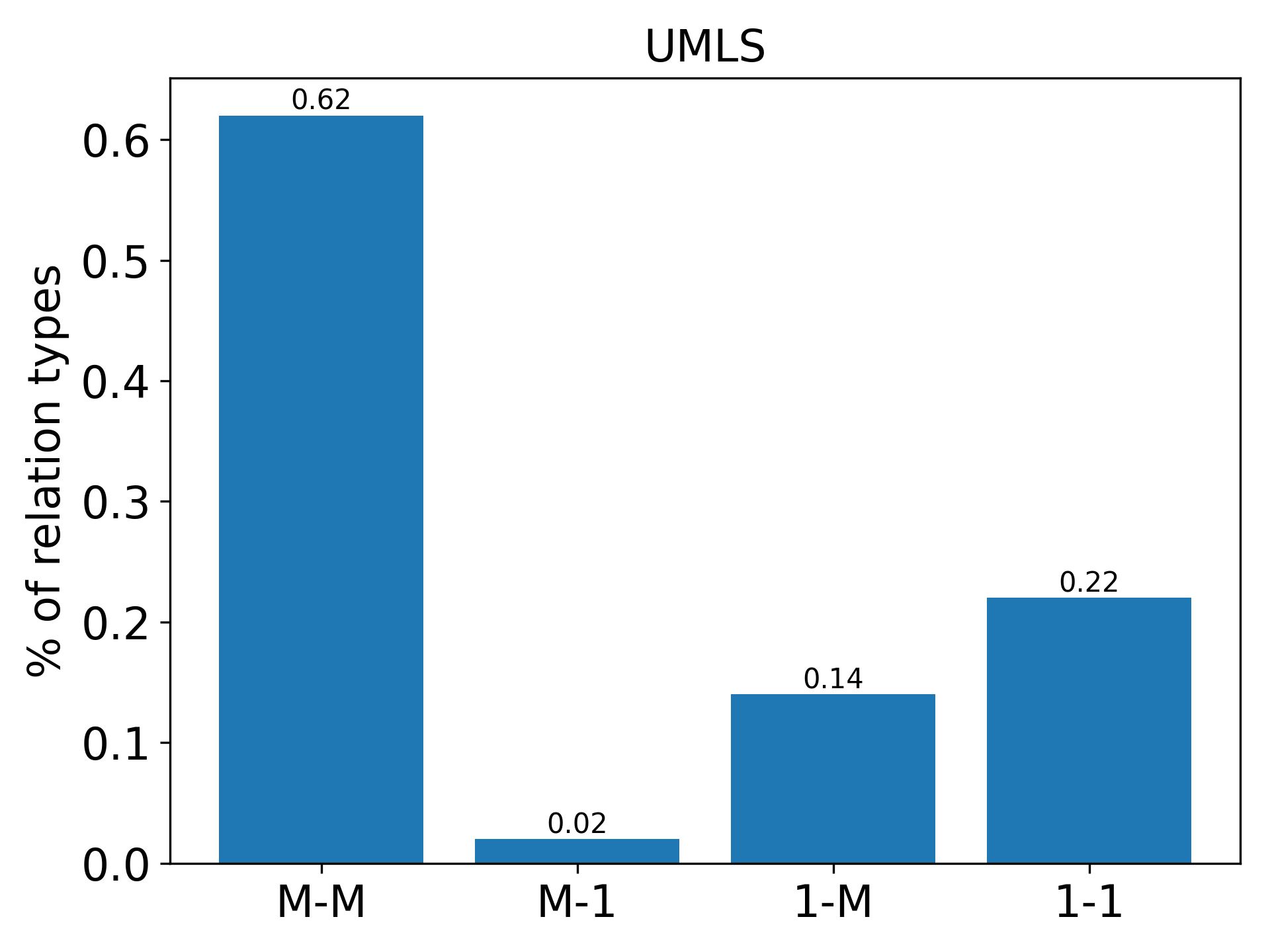 }
    \end{minipage}
    \begin{minipage}{0.24\textwidth}
         \includegraphics[width=\textwidth]{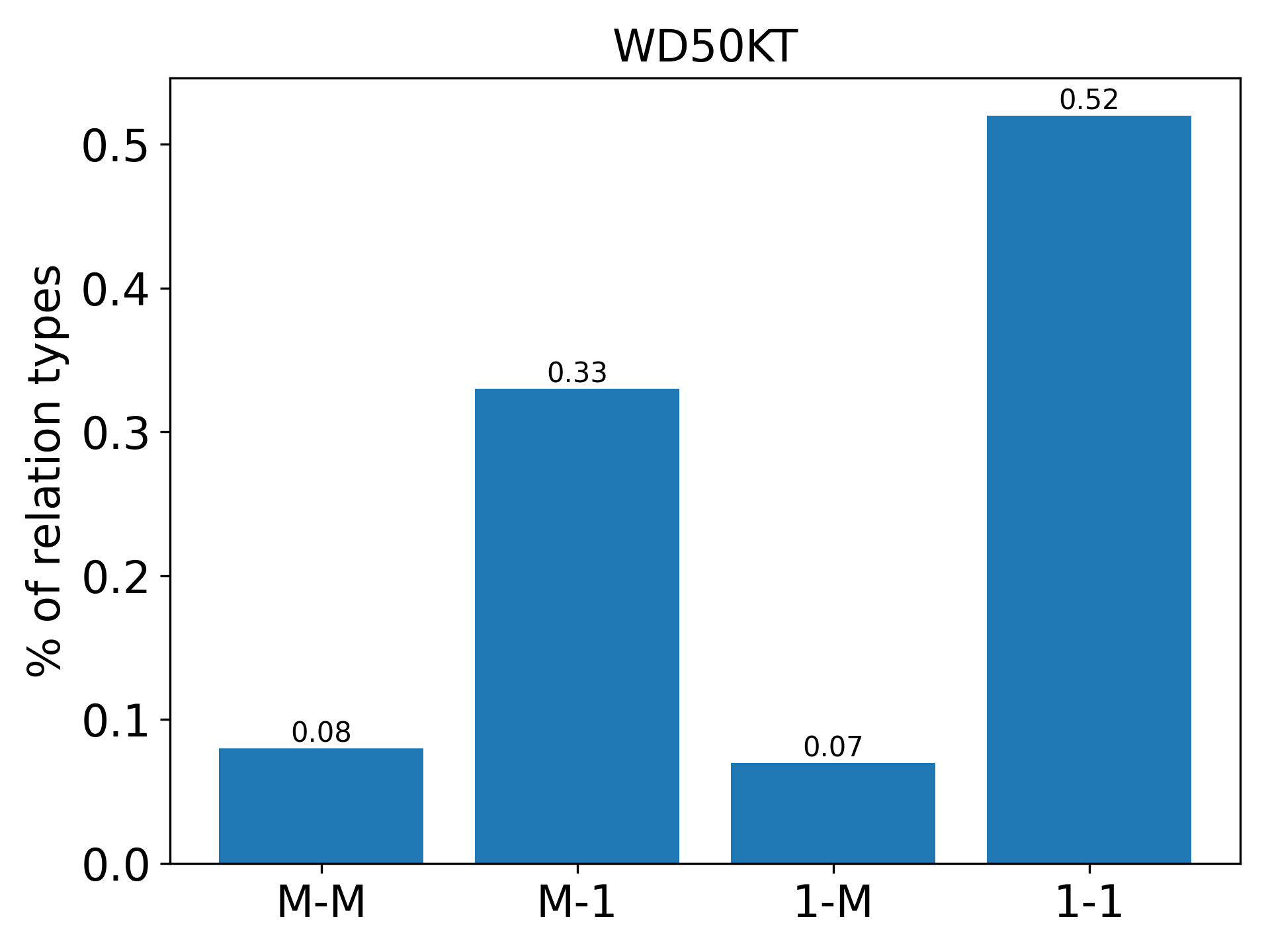}
    \end{minipage}
    
     \begin{minipage}{0.24\textwidth}
         \includegraphics[width=\textwidth]{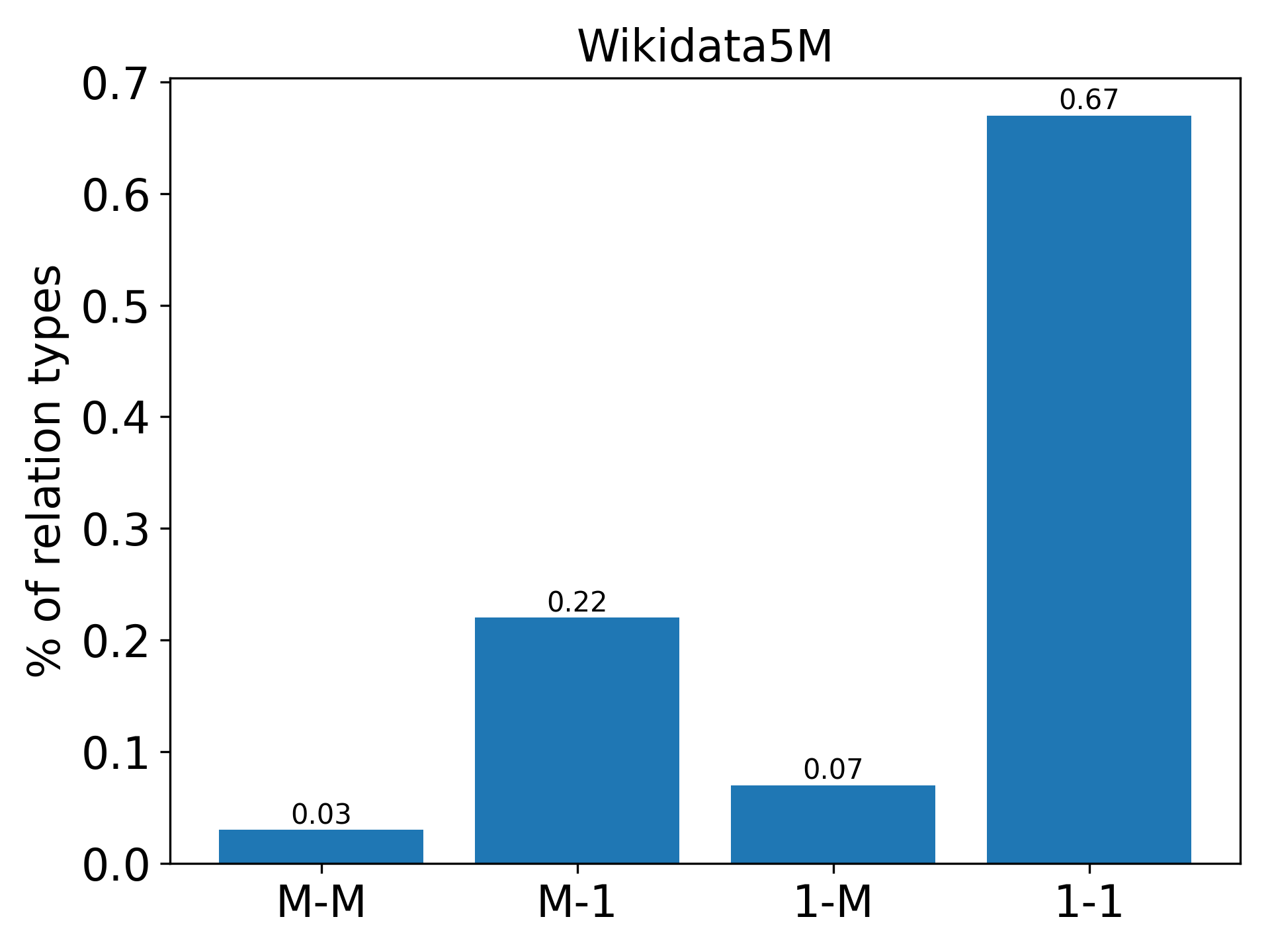 }
     \end{minipage}
     \begin{minipage}{0.24\textwidth}
         \includegraphics[width=\textwidth]{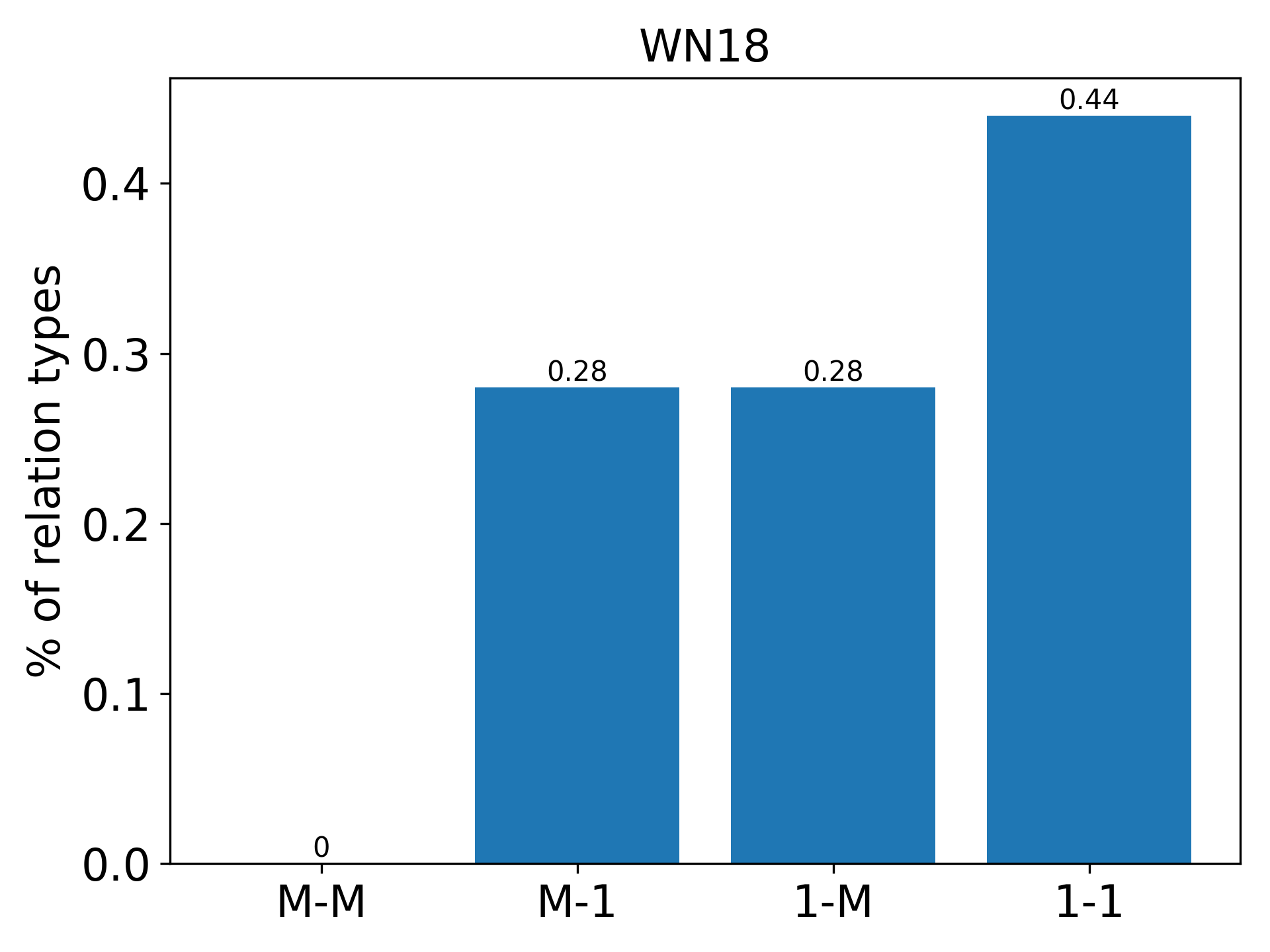 }
    \end{minipage}
    \begin{minipage}{0.24\textwidth}
         \includegraphics[width=\textwidth]{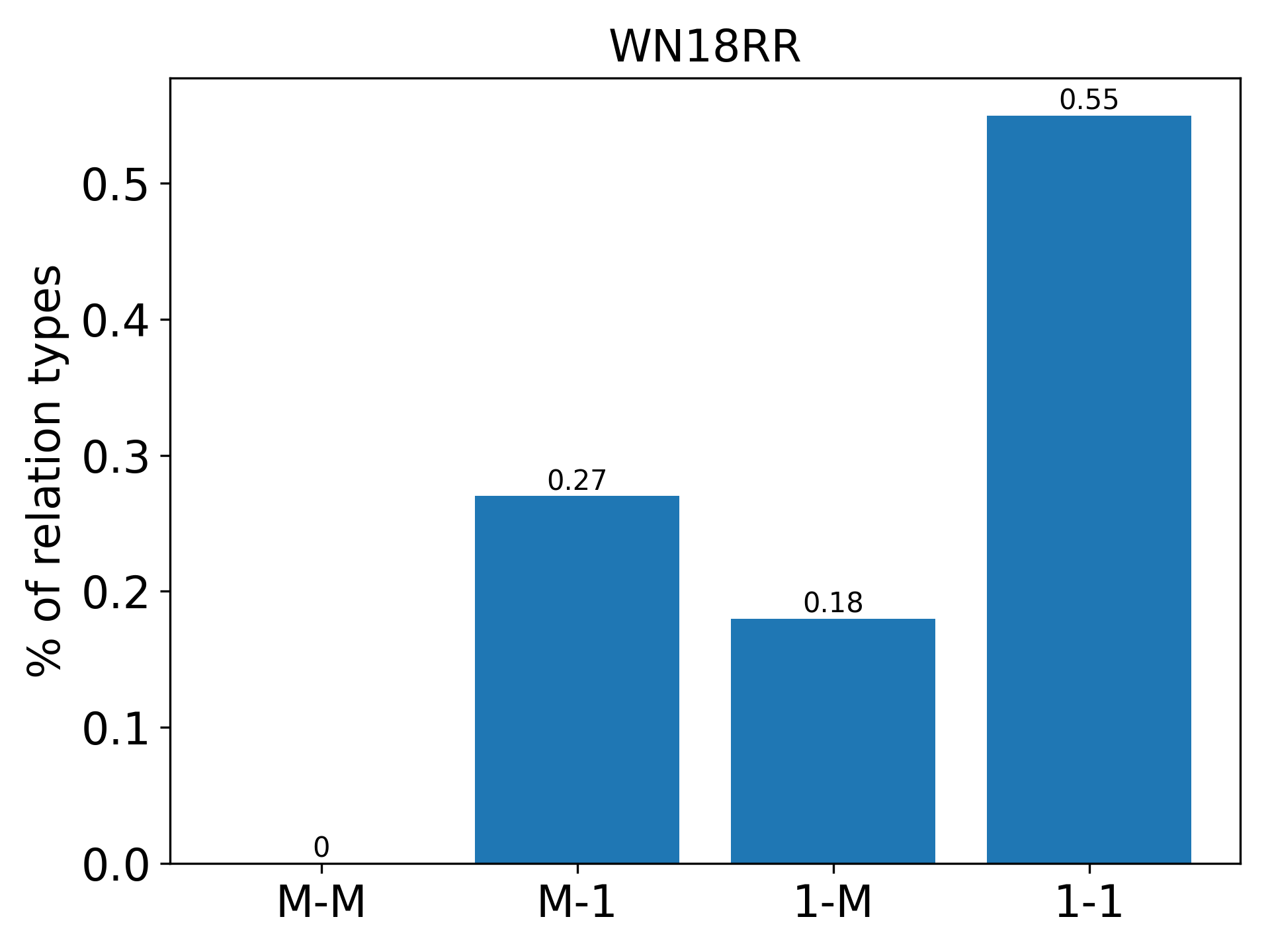}
    \end{minipage}
    \begin{minipage}{0.24\textwidth}
         \includegraphics[width=\textwidth]{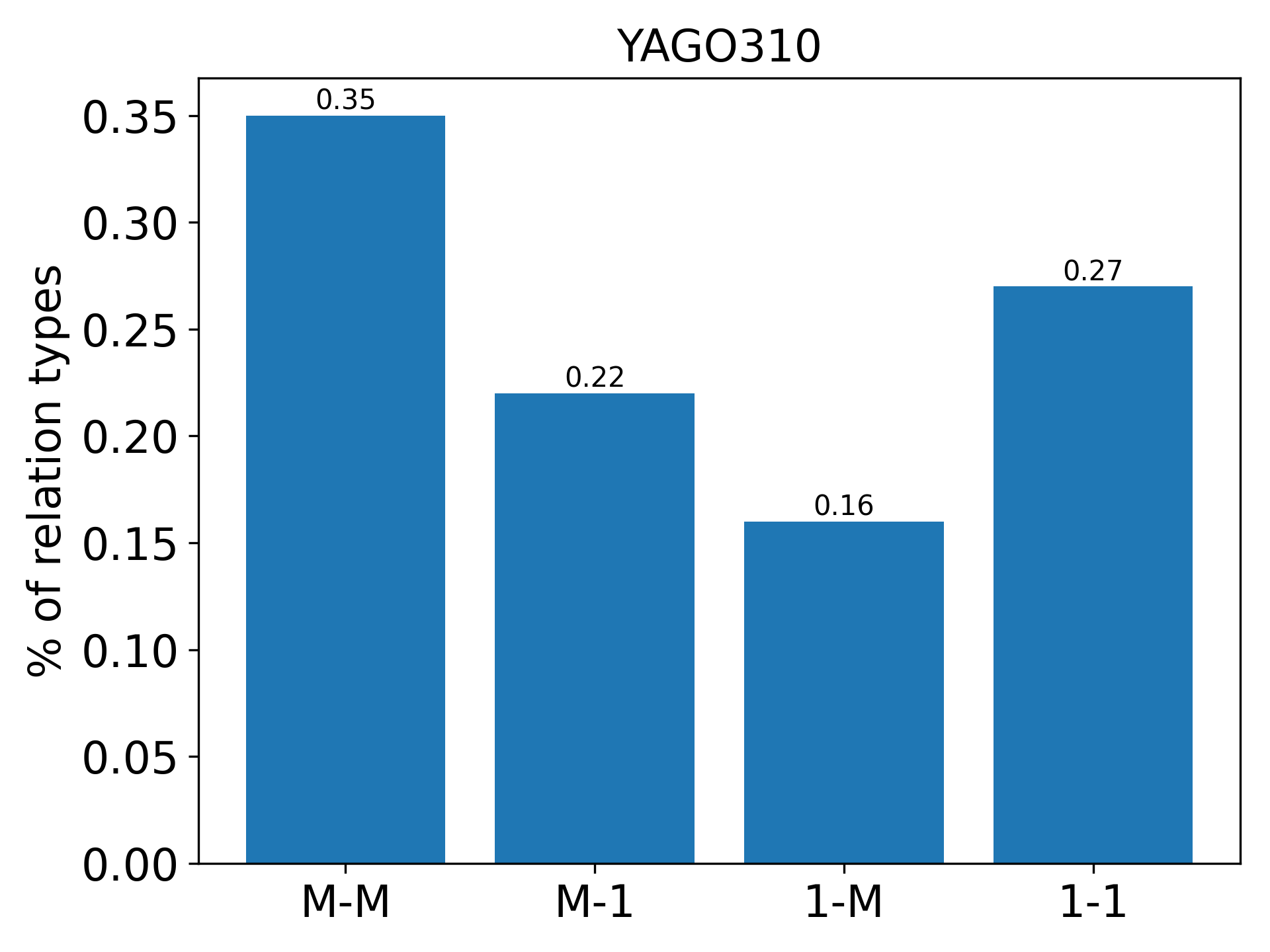}
    \end{minipage}
        \caption{Distribution of relation cardinalities in different KGs. Each bar is marked with the \# number of relations of the specified type.
        Due to space limits, the CoDExMedium dataset is shown in the Appx.}
        \label{fig: relcardinality}
\end{figure}

\textbf{Relation cardinality}: Relation cardinality describes the numerical relationship between the possible head and tail entities of a relation
(i.e., how many entities a relation can have as a tail or head). The possible types are: 
(i) one-to-one (1-1), (ii) one-to-many (1-M), (iii) many-to-many (M-M), (iv) many-to-one (M-1) \citep{kgxboard}.
For illustration, the relation \texttt{surrounds} (from the UMLS dataset) is 1-1, 
while the relation \texttt{GeneActivationGene} (from the OpenBioLink dataset) is M-M because various genes can activate multiple other genes. 
Fig.  \ref{fig: relcardinality} plots the relation cardinality distribution for each dataset considered. 
We observe several different dataset profiles: 
\textbf{(i) 1-1 dominance}: In \m{15} of the \m{28} datasets 
the leading relation type is 1-1 including many semantic datasets. 
Overall, the number of 1-1 relations is more pronounced in the semantic datasets than in the biomedical datasets. 
\textbf{(ii) M-M dominance}: In \m{11} of the \m{28} datasets, the leading relation type is M-M. 
Biological datasets (e.g. BioKG, Hetionet, OpenBioLink, PharmKG8k) are dominated by M-M relations
with the exception of PharmKG and Globi which exhibit a distinct profile. 
All societal datasets also fall in this category.
\textbf{(iii)} \textbf{mixed cardinalities}: We observe that some of the most frequently used semantic datasets (FB15k, FB15k-237, Yago310) 
have a significant number of all \m{4} cardinalities unlike the rest of the analyzed datasets.
\textbf{(iv) mixed profile with a 1-1 or M-M skew}: 
All biomedical datasets tend to be skewed towards M-M relations, except PharmKG and Globi. 
On the other hand, all semantic datasets tent to be skewed towards 1-1 relations, with the notable exception of Yago310. 
CoDex* datasets are unique since they have predominantly M-1 relations.

\begin{figure}[t]
     \centering
     \begin{minipage}{0.24\textwidth}
         \includegraphics[width=\textwidth]{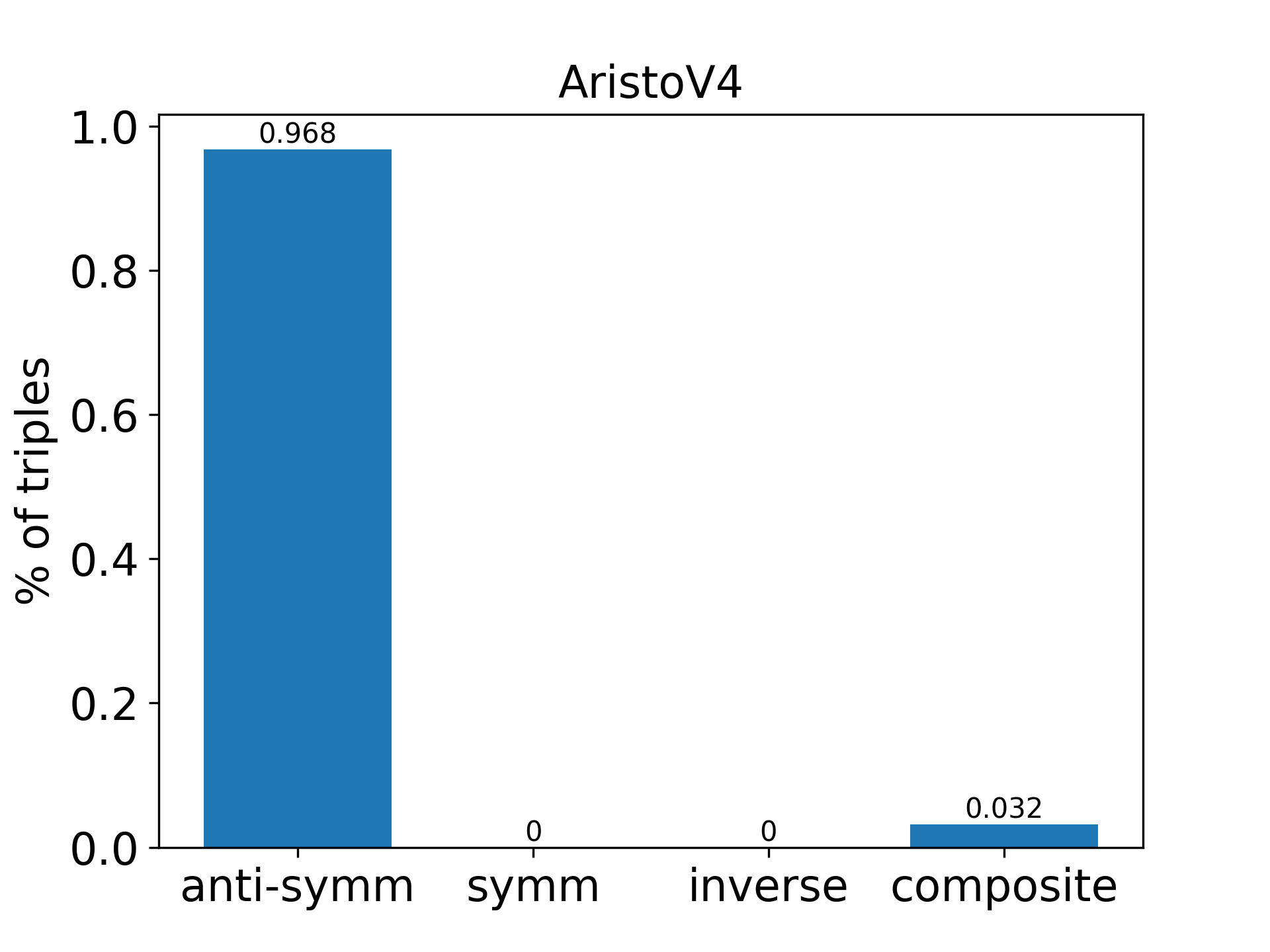}
     \end{minipage}
     \begin{minipage}{0.24\textwidth}
         \includegraphics[width=\textwidth]{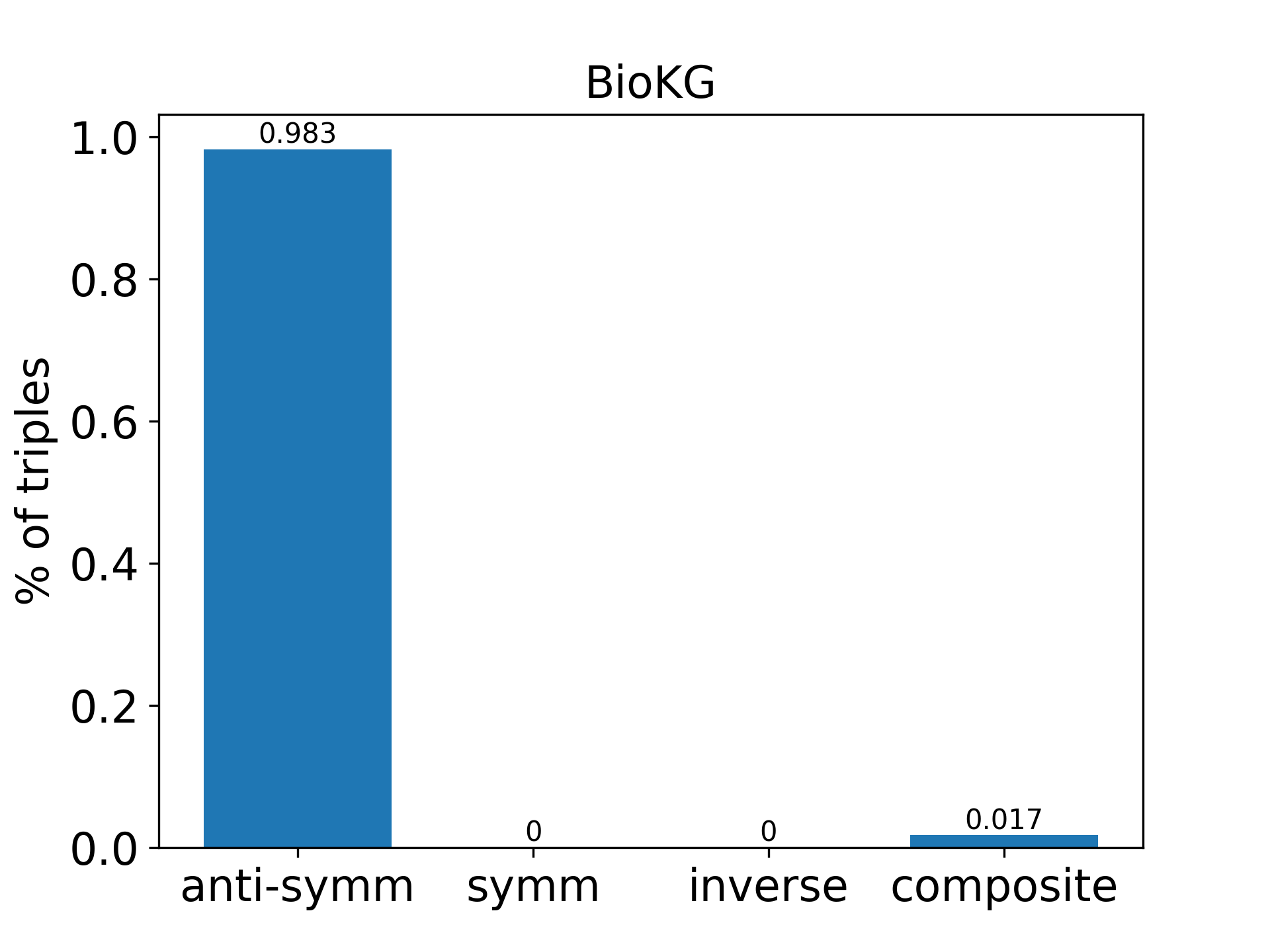}
    \end{minipage}
    \begin{minipage}{0.24\textwidth}
         \includegraphics[width=\textwidth]{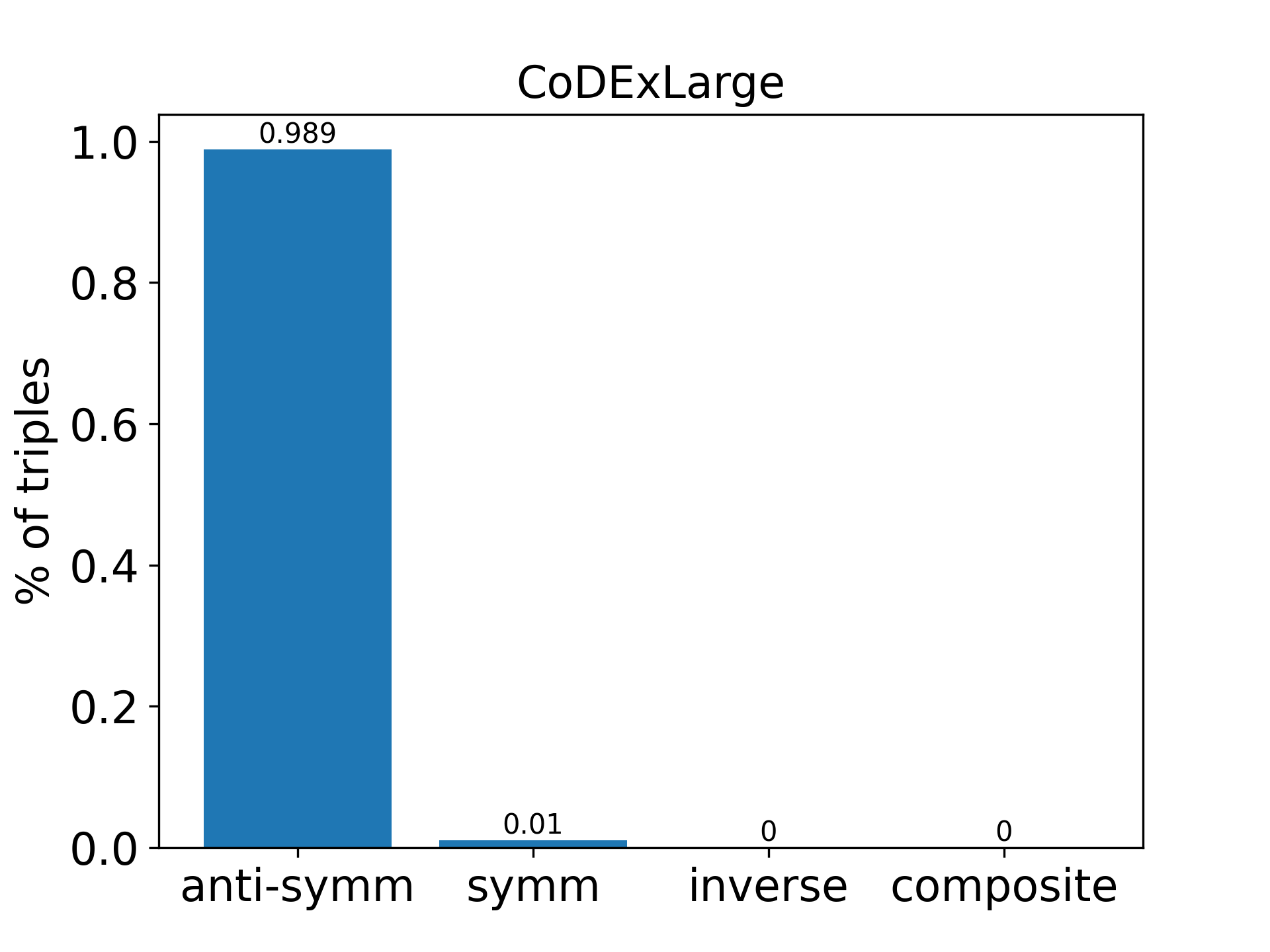}
    \end{minipage}
     \begin{minipage}{0.24\textwidth}
         \includegraphics[width=\textwidth]{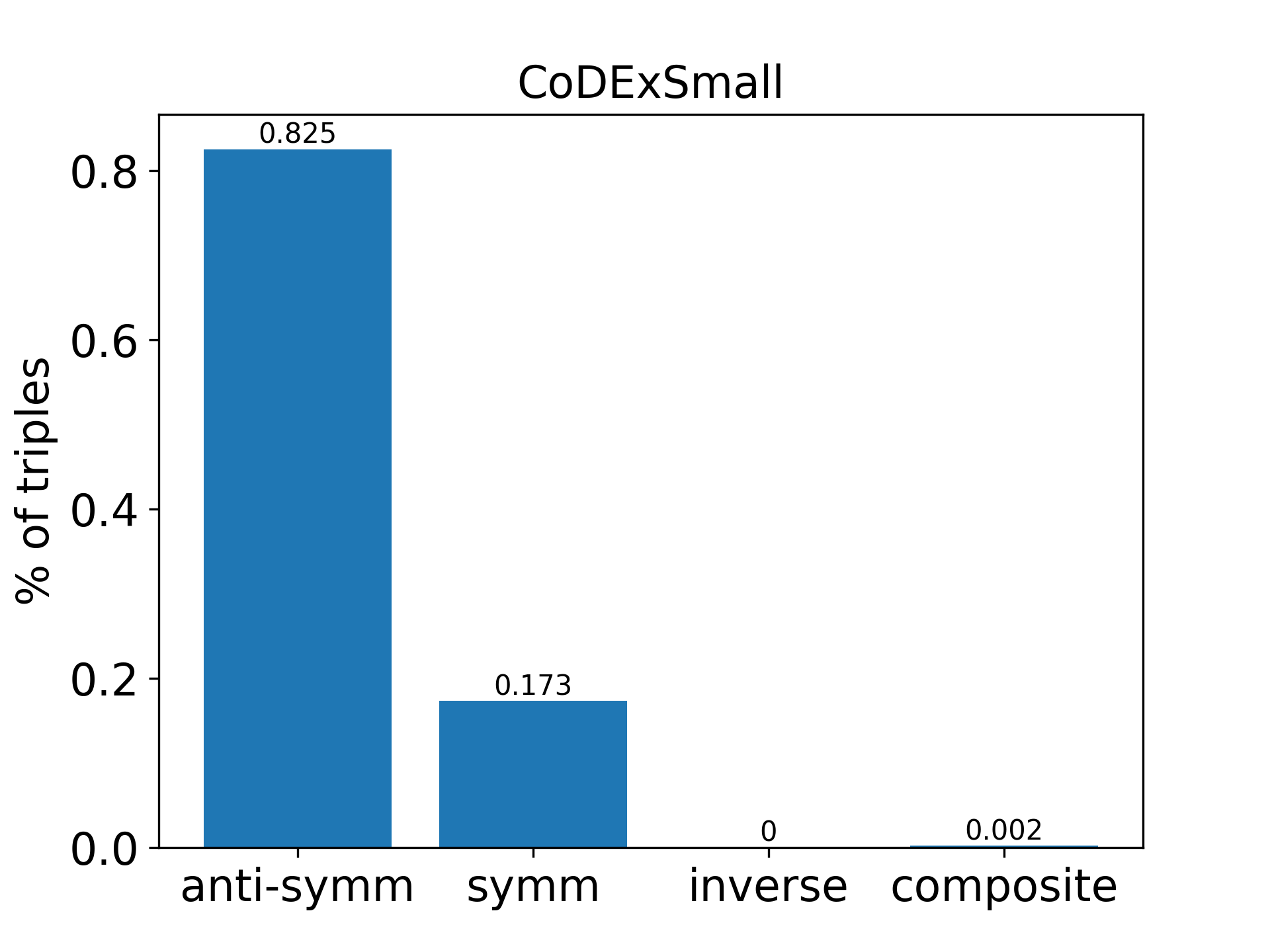 }
    \end{minipage}
    
    \begin{minipage}{0.24\textwidth}
         \includegraphics[width=\textwidth]{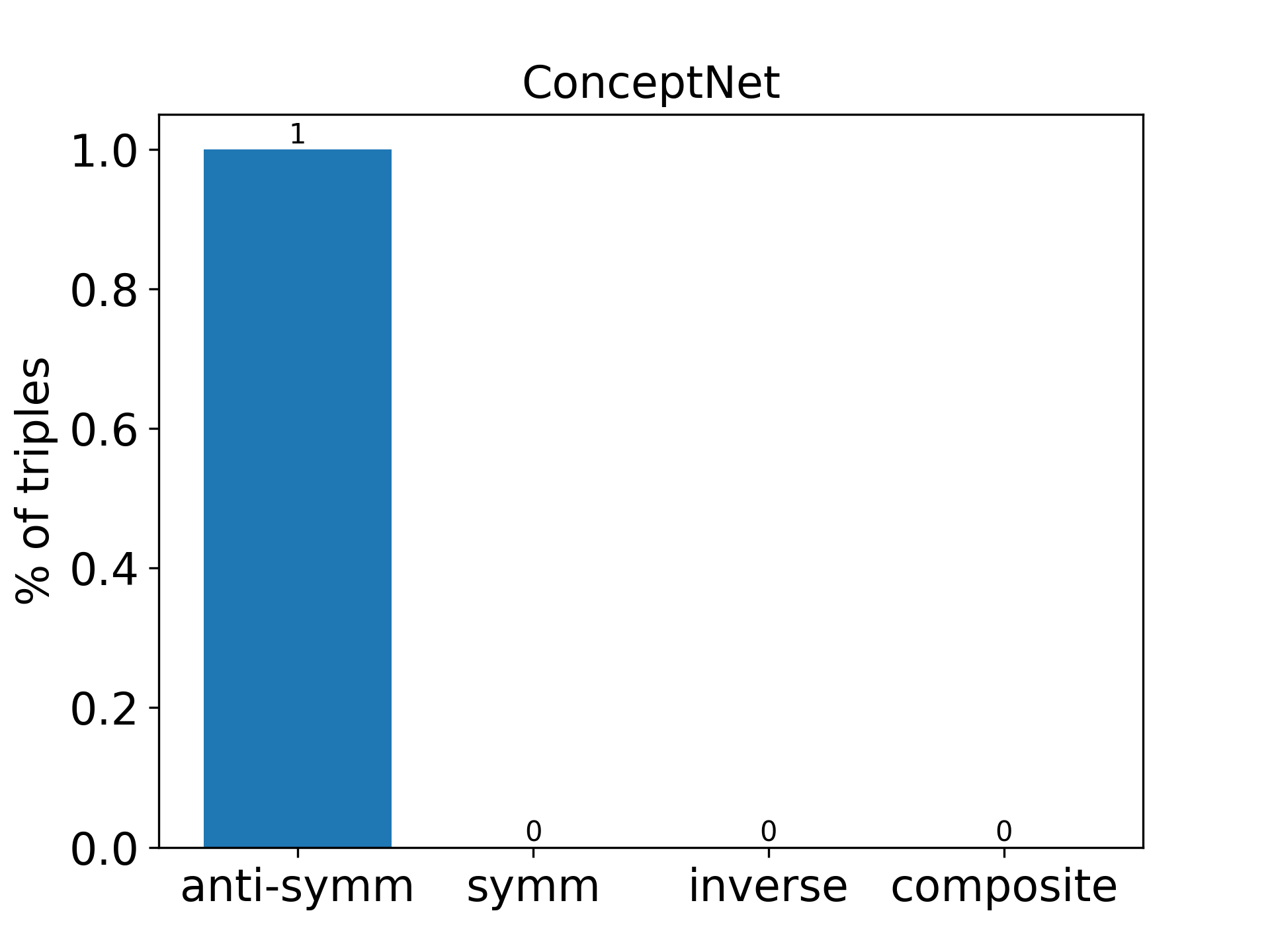 }
    \end{minipage}
     \begin{minipage}{0.24\textwidth}
         \includegraphics[width=\textwidth]{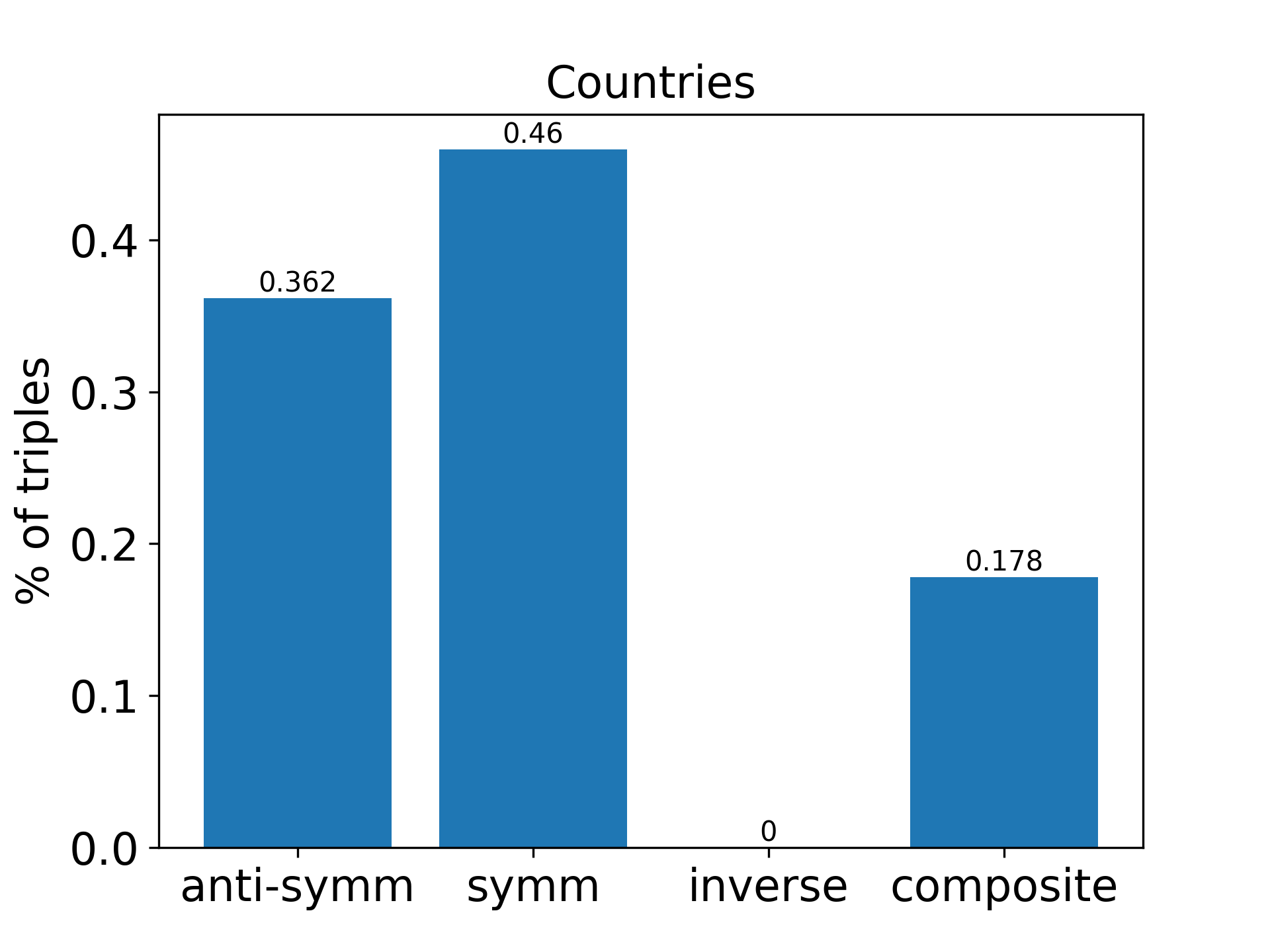 }
     \end{minipage}
     \begin{minipage}{0.24\textwidth}
         \includegraphics[width=\textwidth]{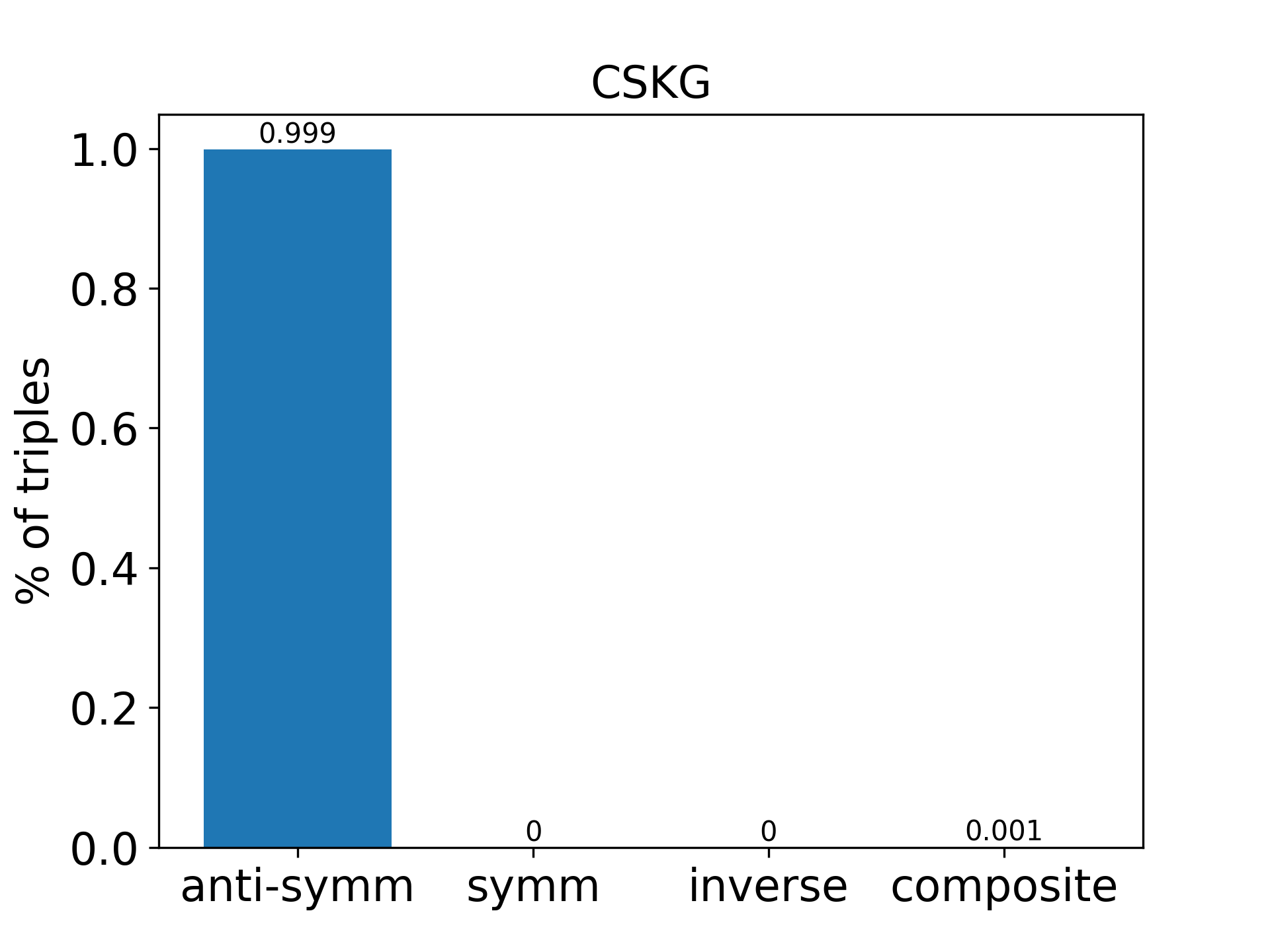 }
    \end{minipage}
    \begin{minipage}{0.24\textwidth}
         \includegraphics[width=\textwidth]{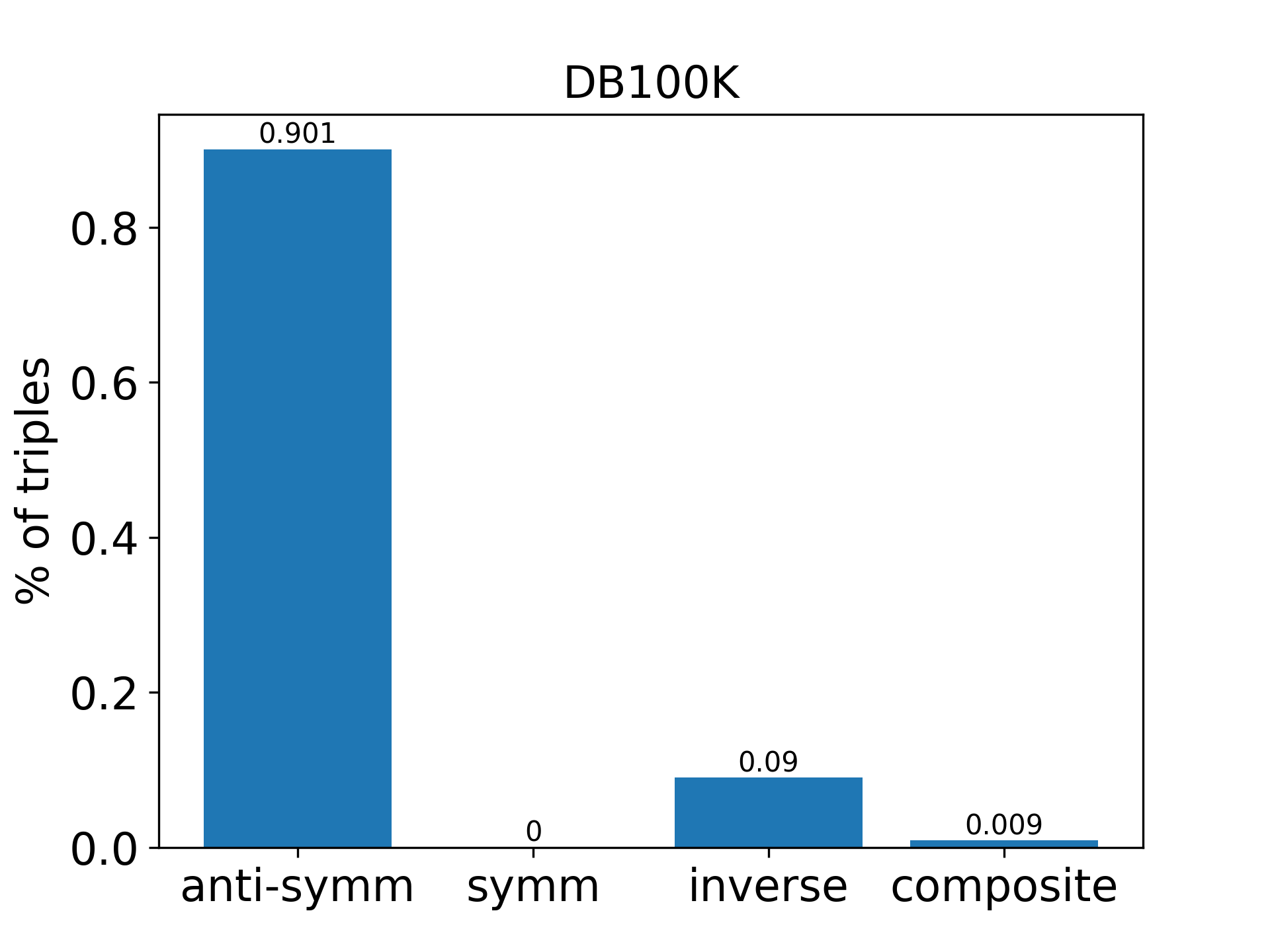 }
    \end{minipage}
    
     \begin{minipage}{0.24\textwidth}
         \includegraphics[width=\textwidth]{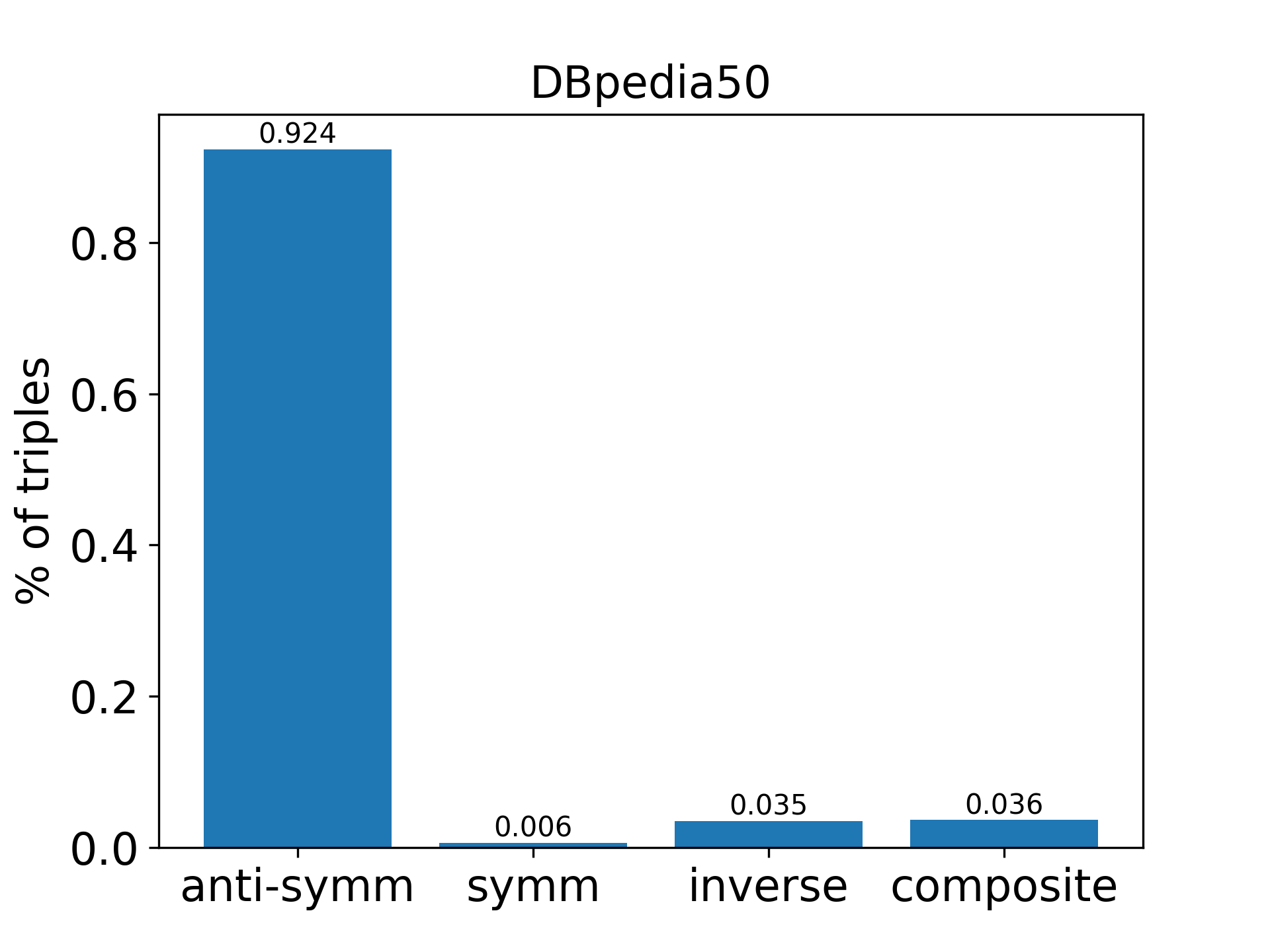 }
     \end{minipage}
     \begin{minipage}{0.24\textwidth}
         \includegraphics[width=\textwidth]{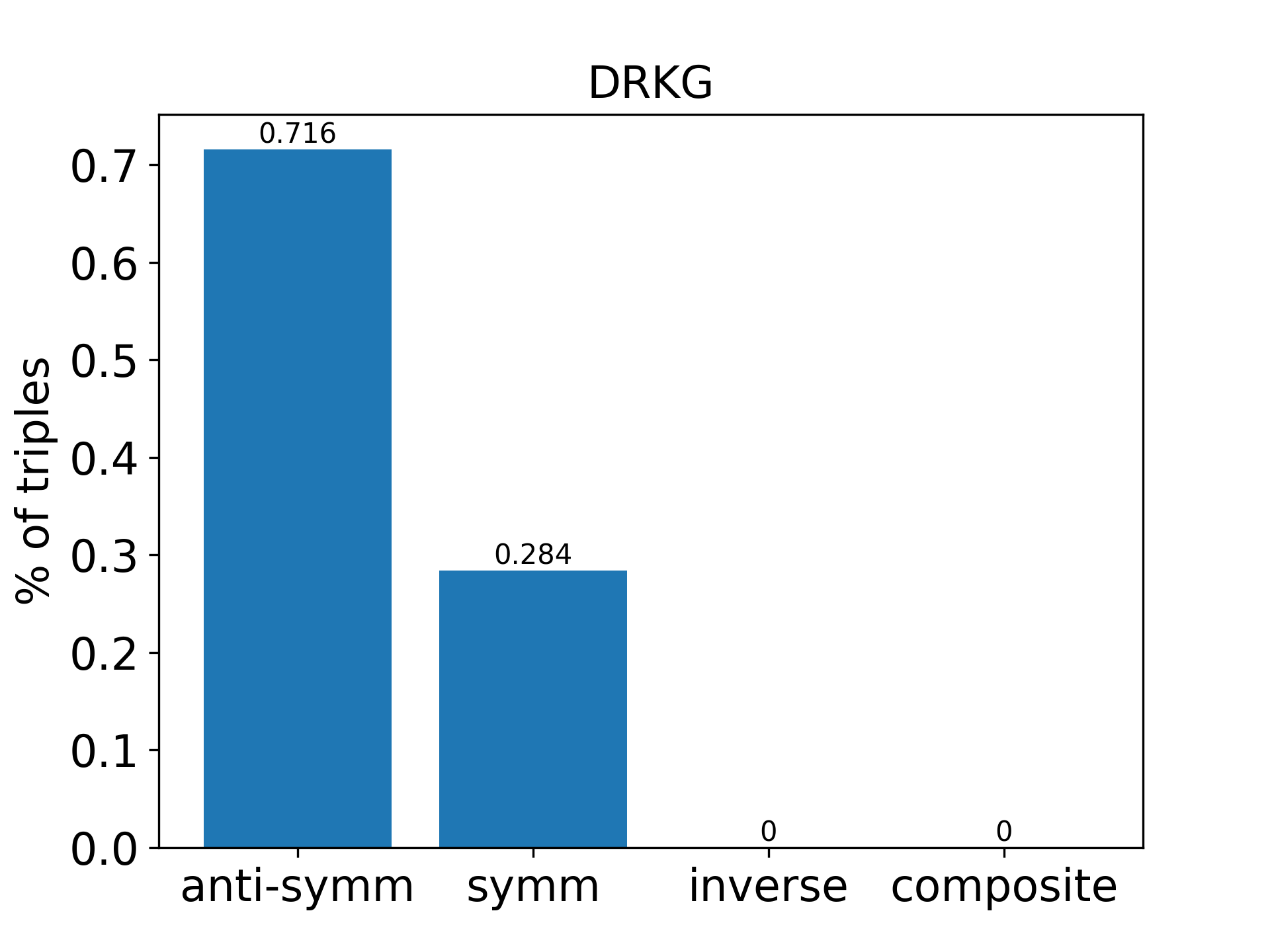 }
    \end{minipage}
    \begin{minipage}{0.24\textwidth}
         \includegraphics[width=\textwidth]{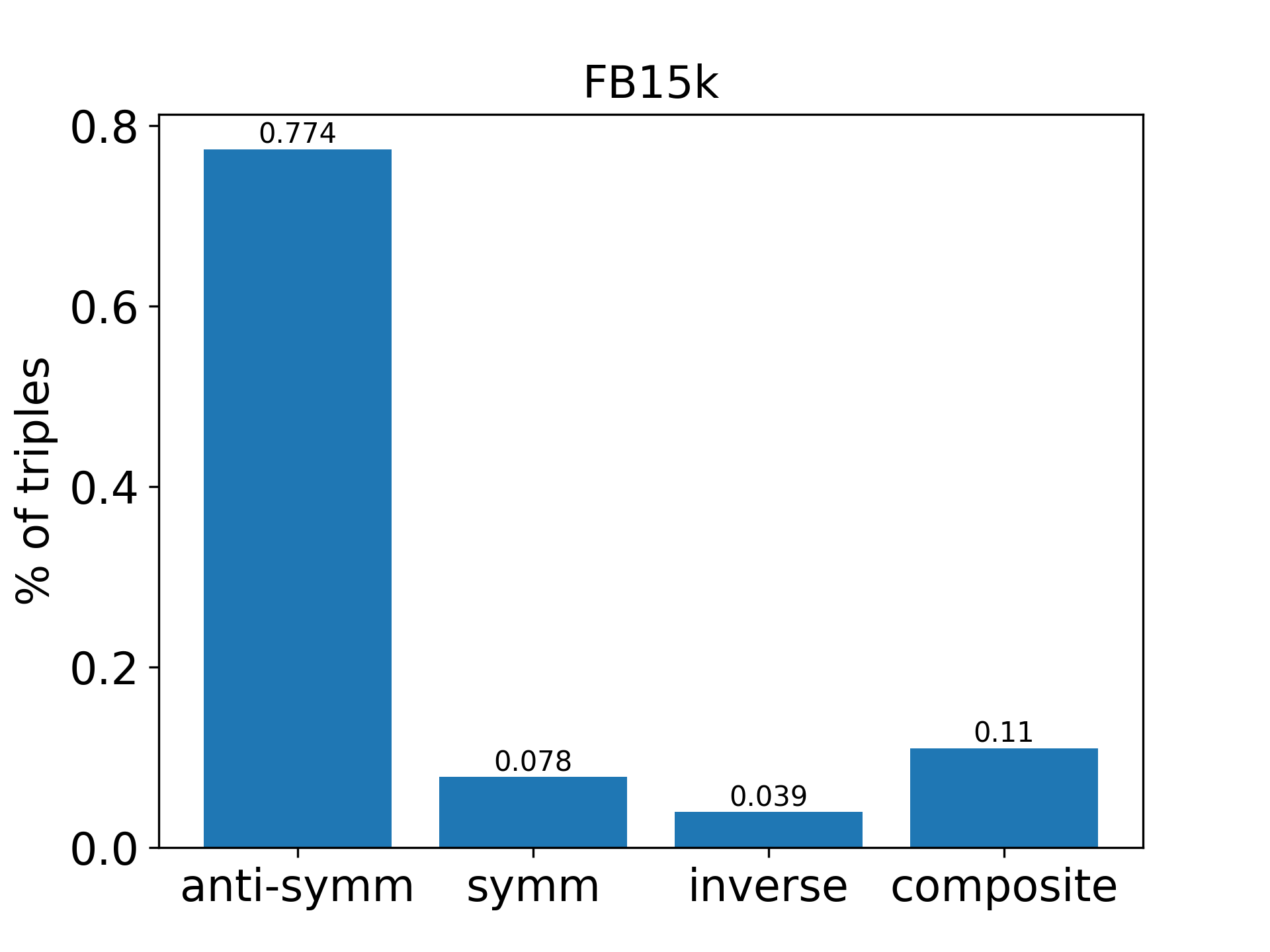 }
    \end{minipage}
     \begin{minipage}{0.24\textwidth}
         \includegraphics[width=\textwidth]{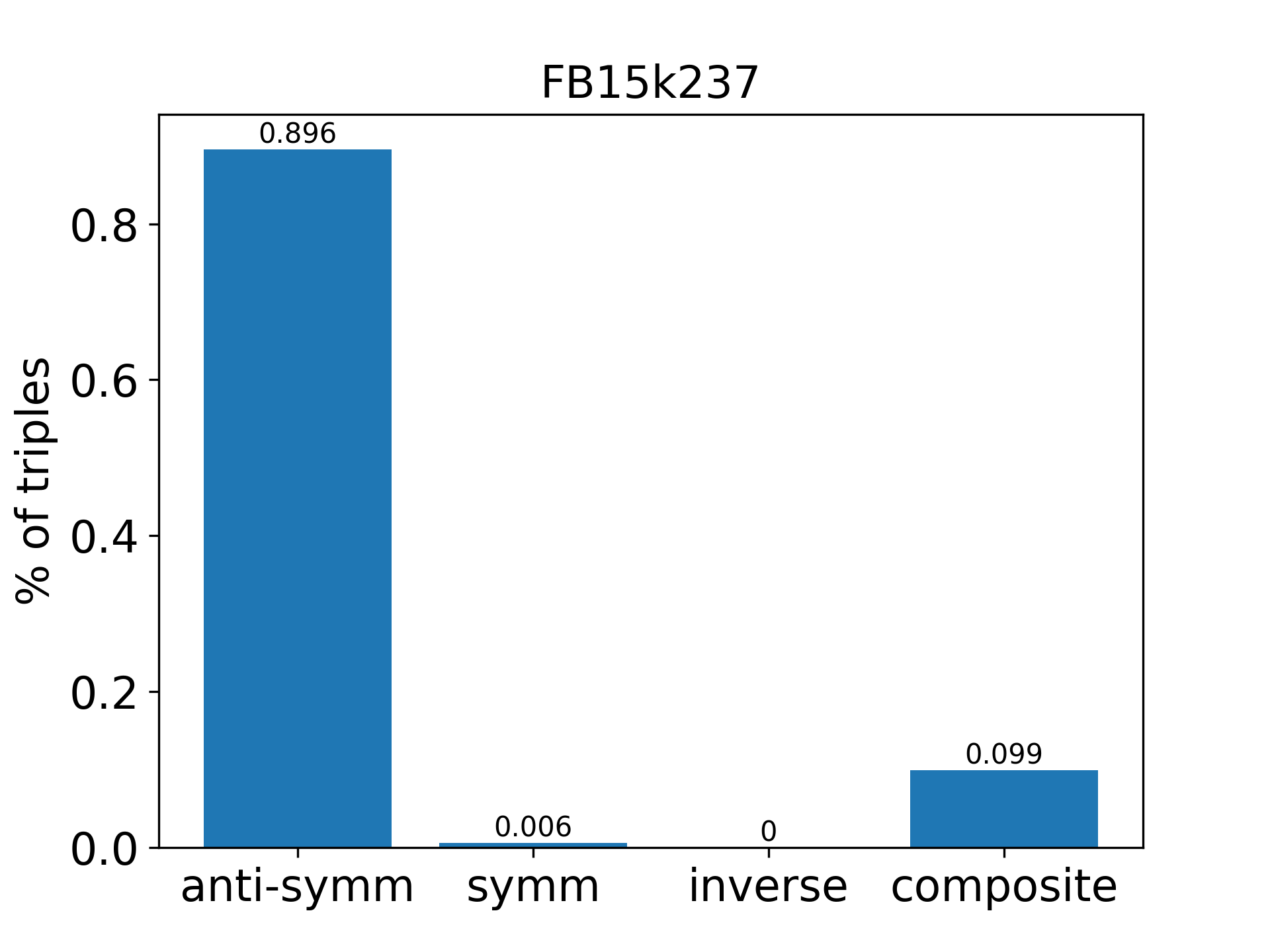 }
     \end{minipage}
     
     \begin{minipage}{0.24\textwidth}
         \includegraphics[width=\textwidth]{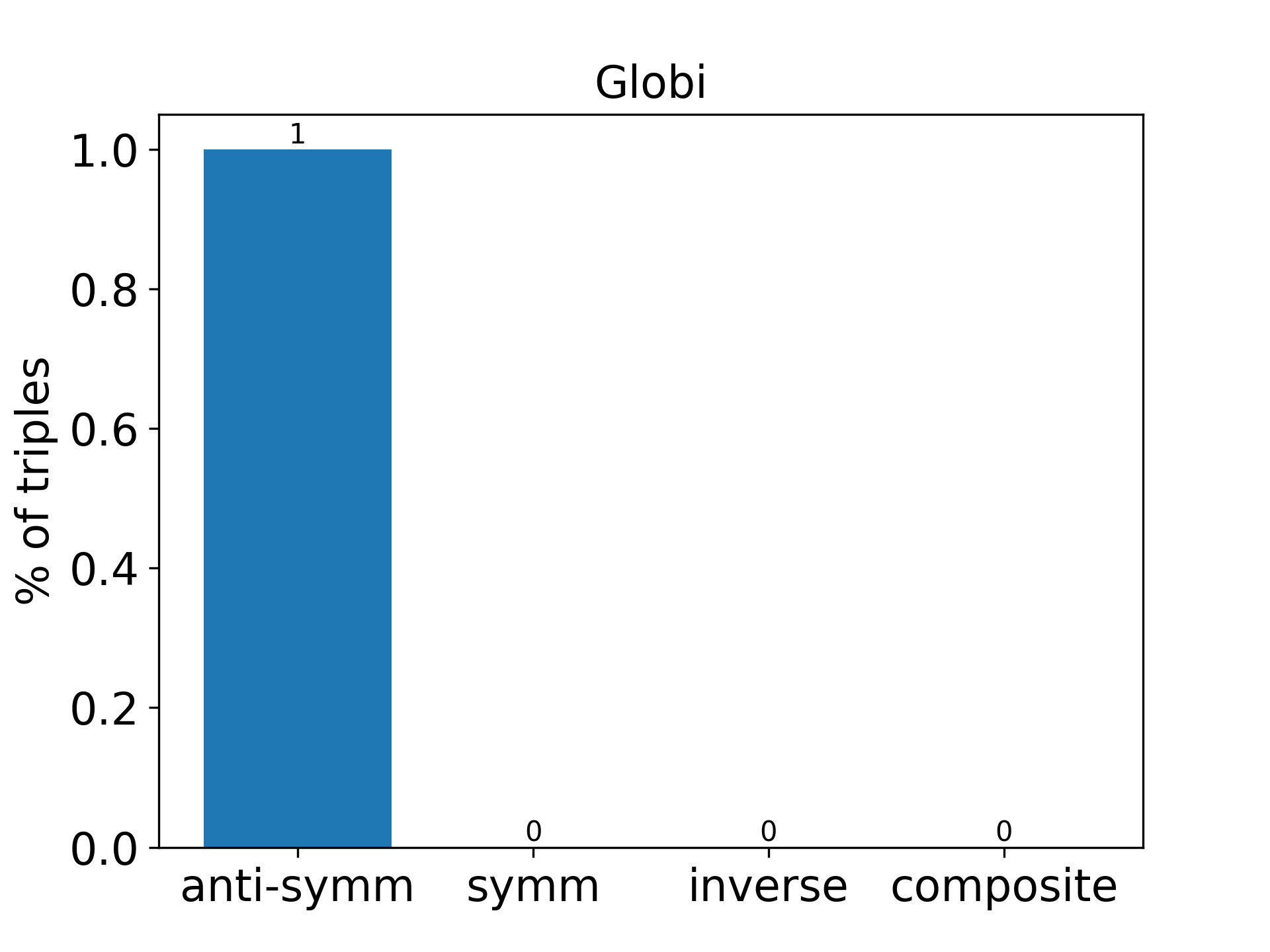 }
    \end{minipage}
    \begin{minipage}{0.24\textwidth}
         \includegraphics[width=\textwidth]{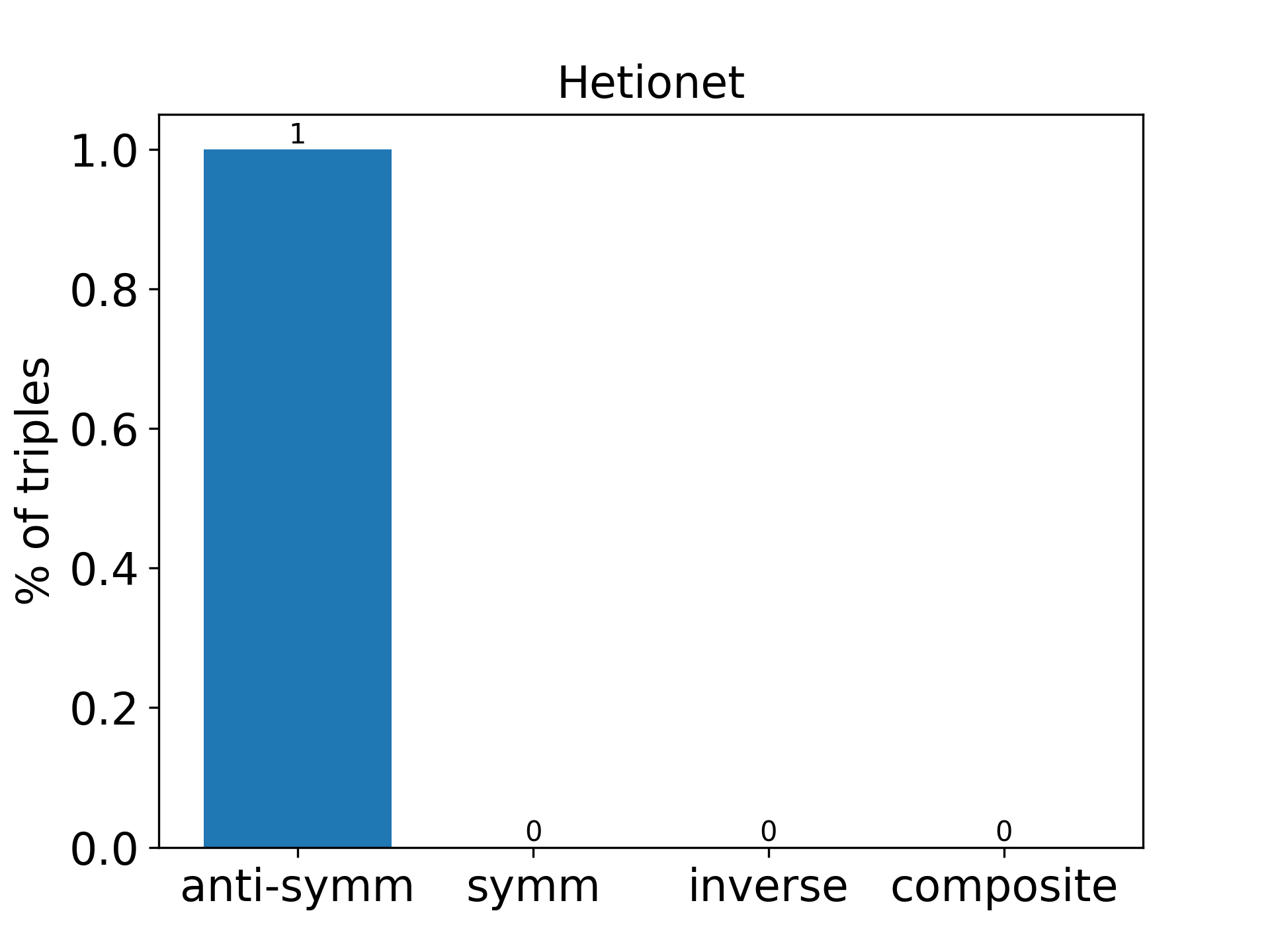 }
    \end{minipage}
     \begin{minipage}{0.24\textwidth}
         \includegraphics[width=\textwidth]{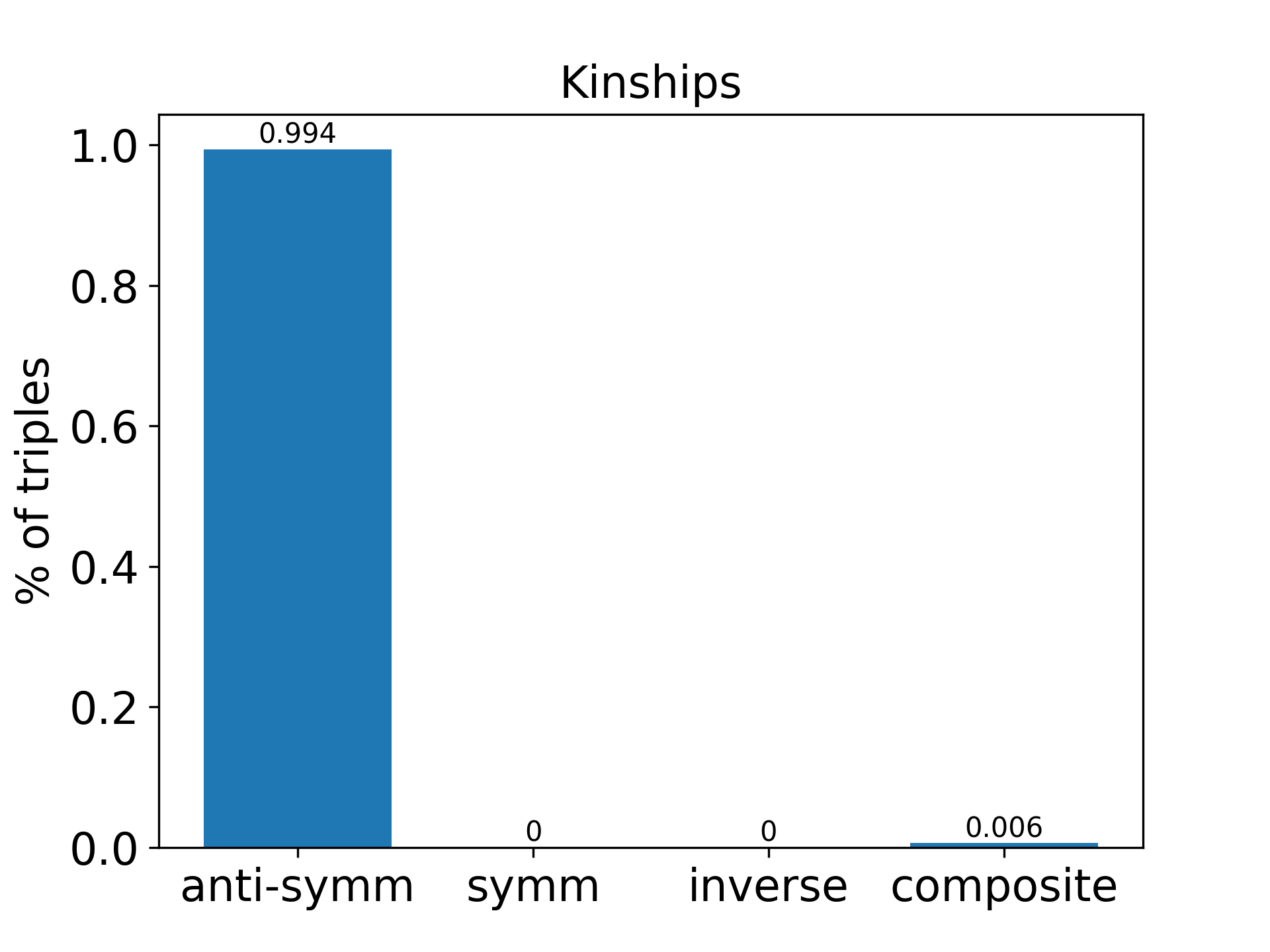 }
     \end{minipage}
     \begin{minipage}{0.24\textwidth}
         \includegraphics[width=\textwidth]{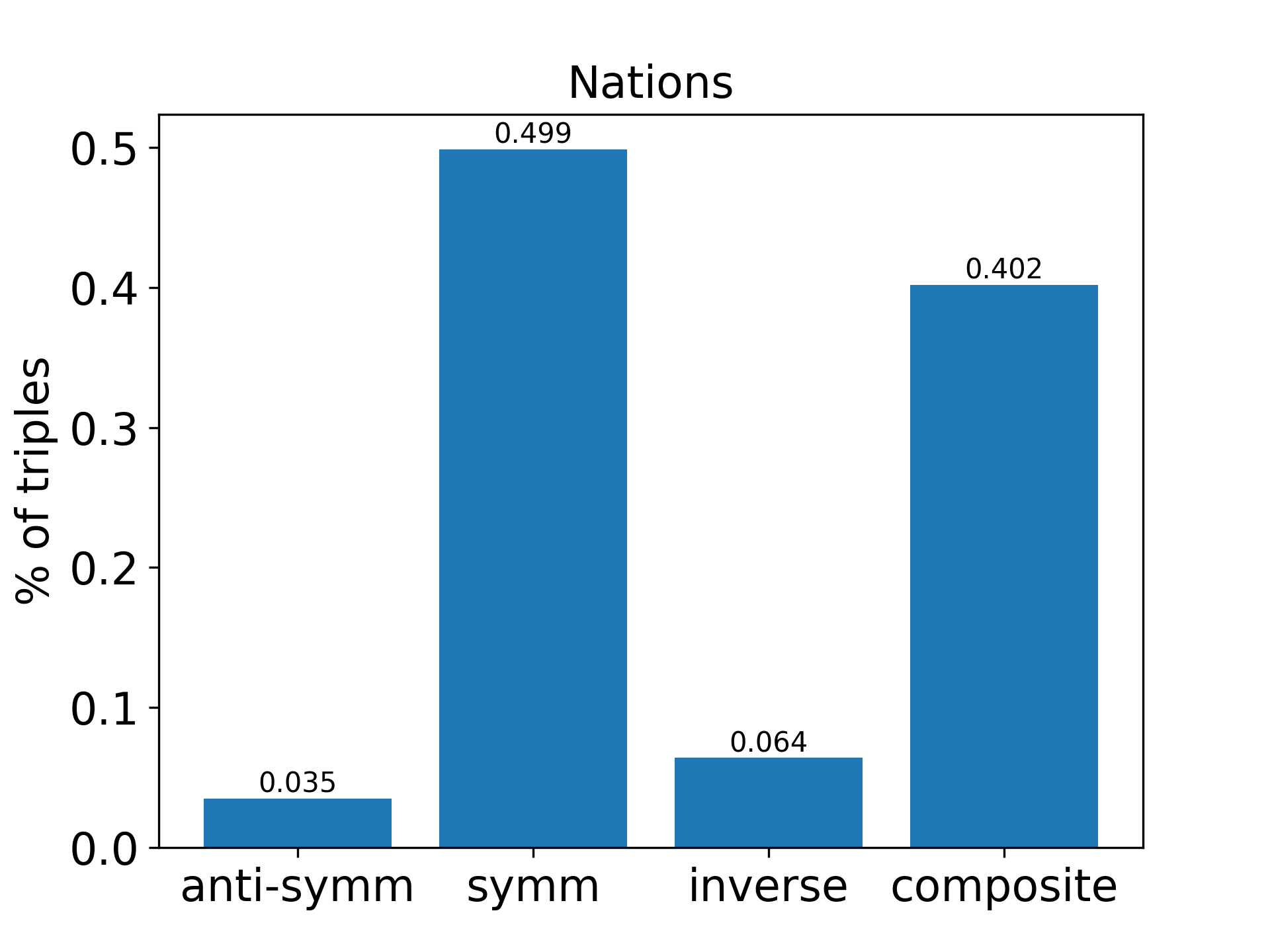}
    \end{minipage}
    
    \begin{minipage}{0.24\textwidth}
         \includegraphics[width=\textwidth]{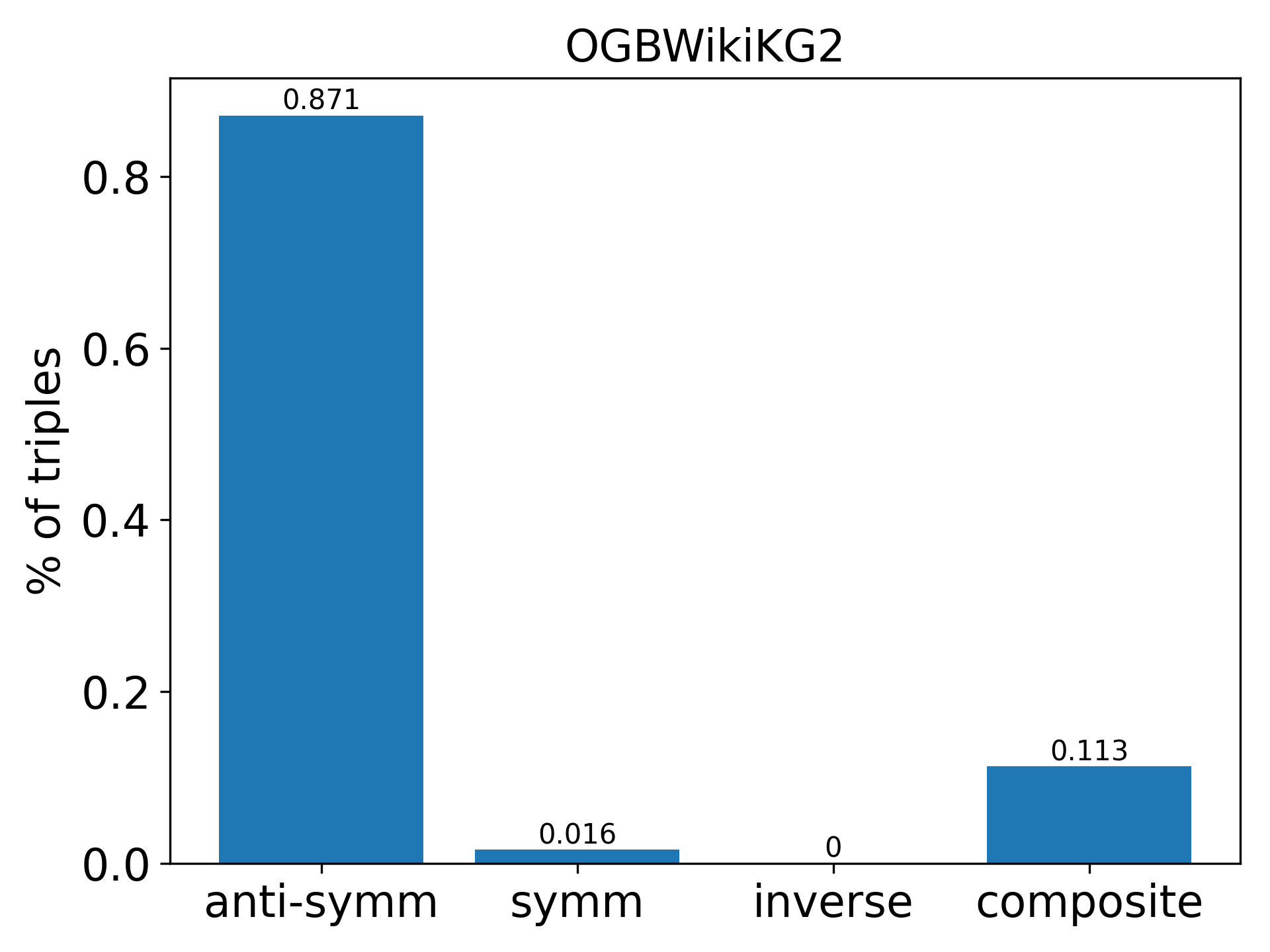}
    \end{minipage}   
    \begin{minipage}{0.24\textwidth}
         \includegraphics[width=\textwidth]{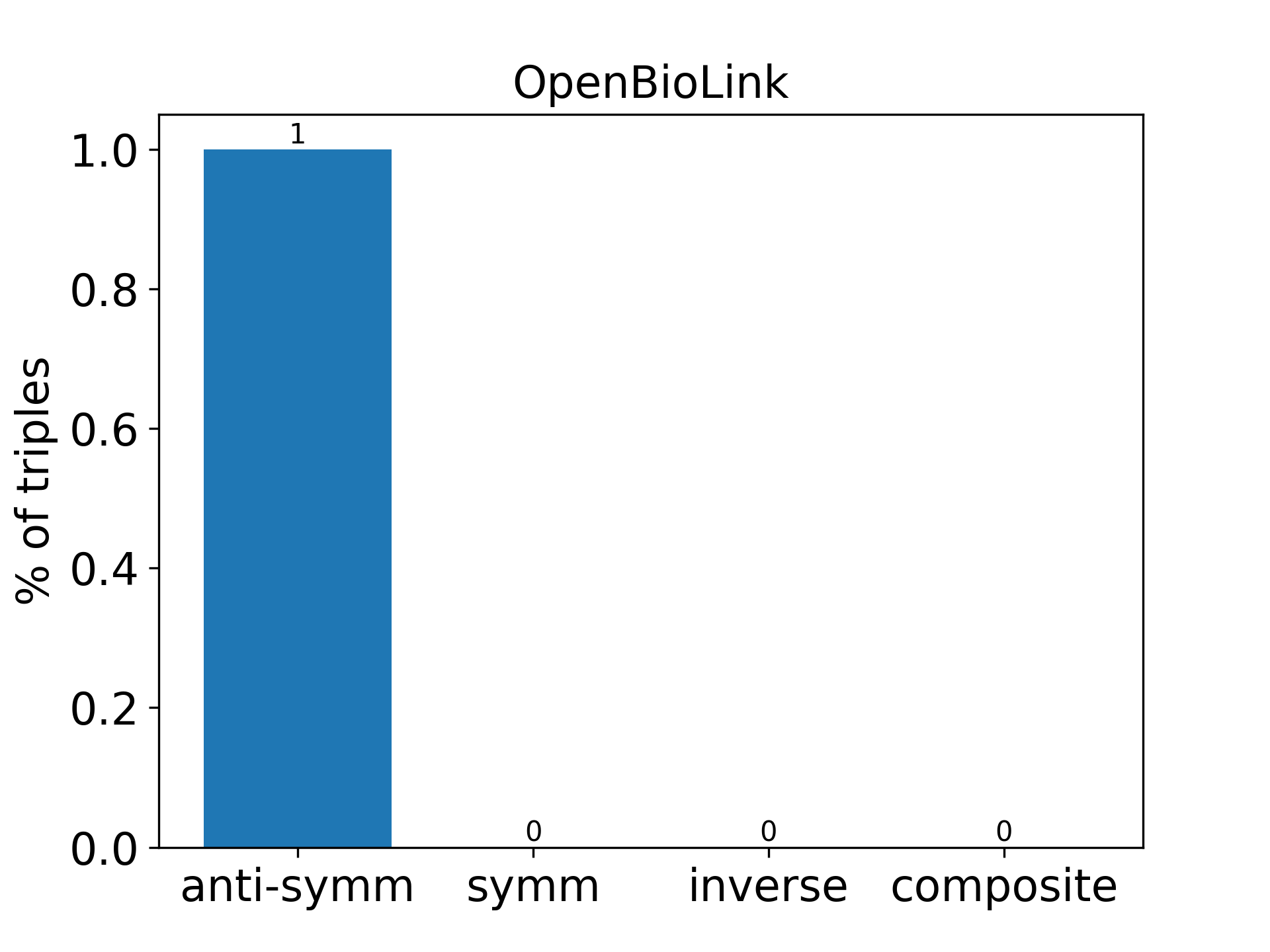 }
    \end{minipage}    
     \begin{minipage}{0.24\textwidth}
         \includegraphics[width=\textwidth]{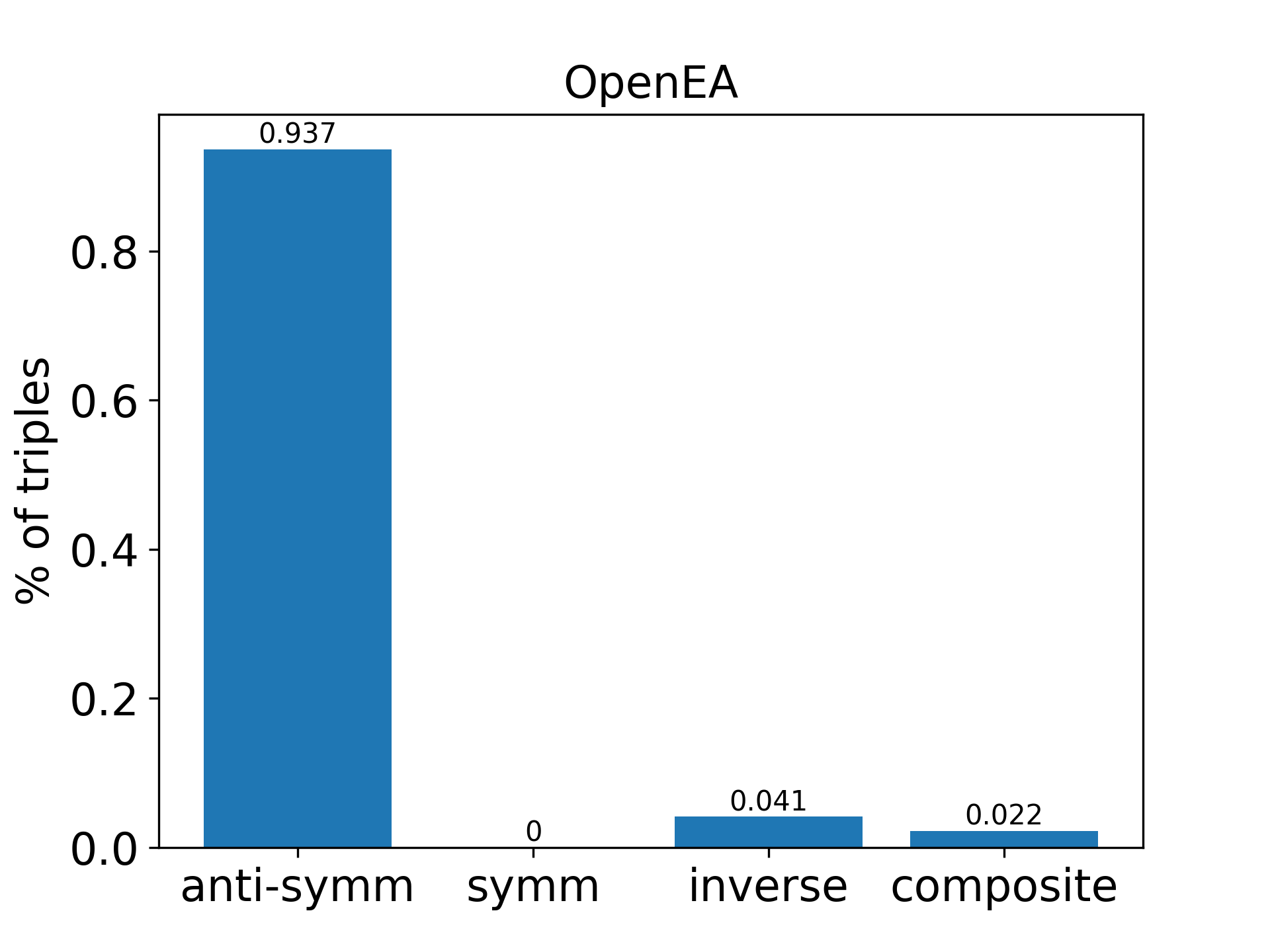 }
     \end{minipage}
     \begin{minipage}{0.24\textwidth}
         \includegraphics[width=\textwidth]{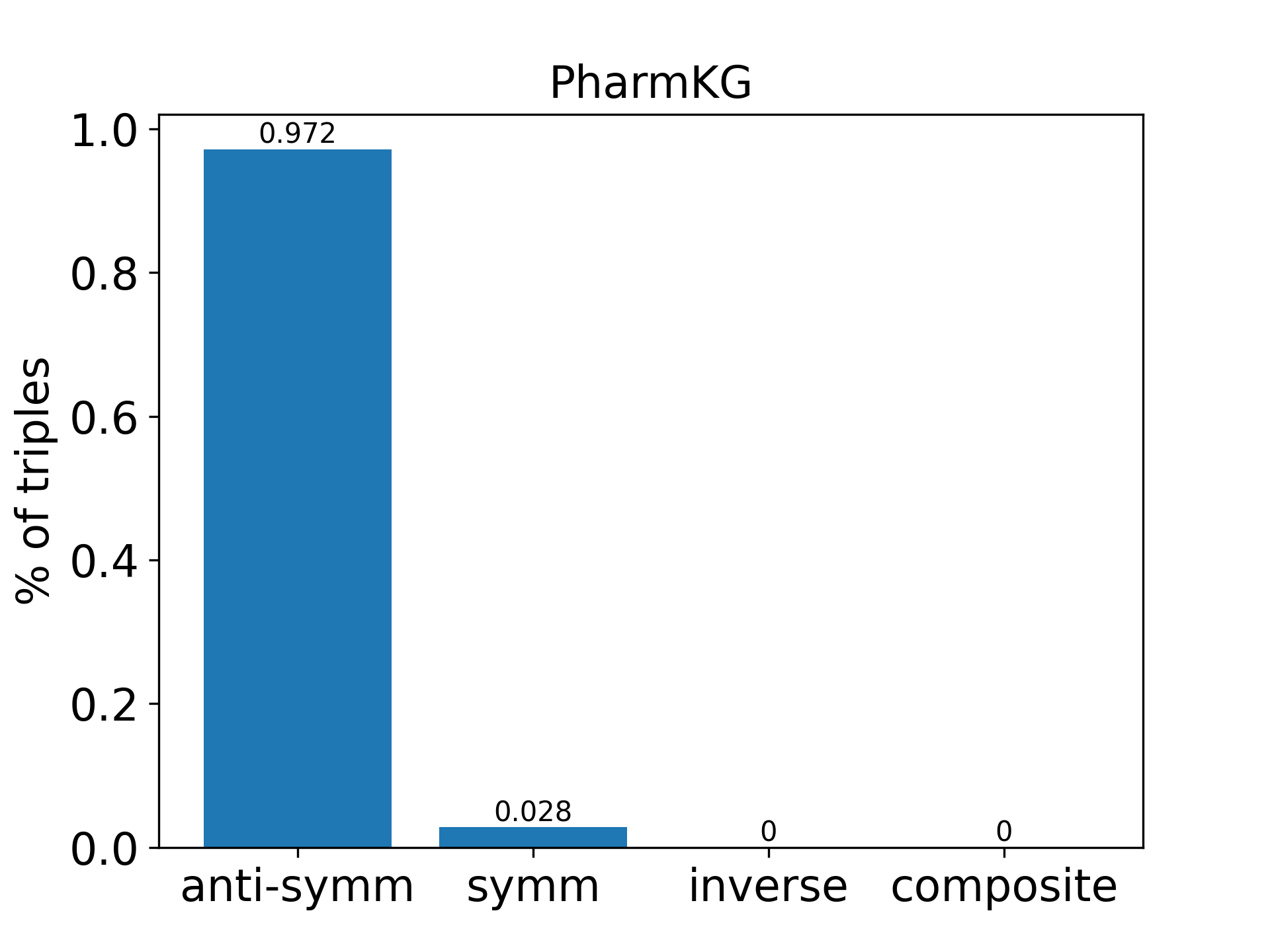 }
    \end{minipage}
    
    \begin{minipage}{0.24\textwidth}
         \includegraphics[width=\textwidth]{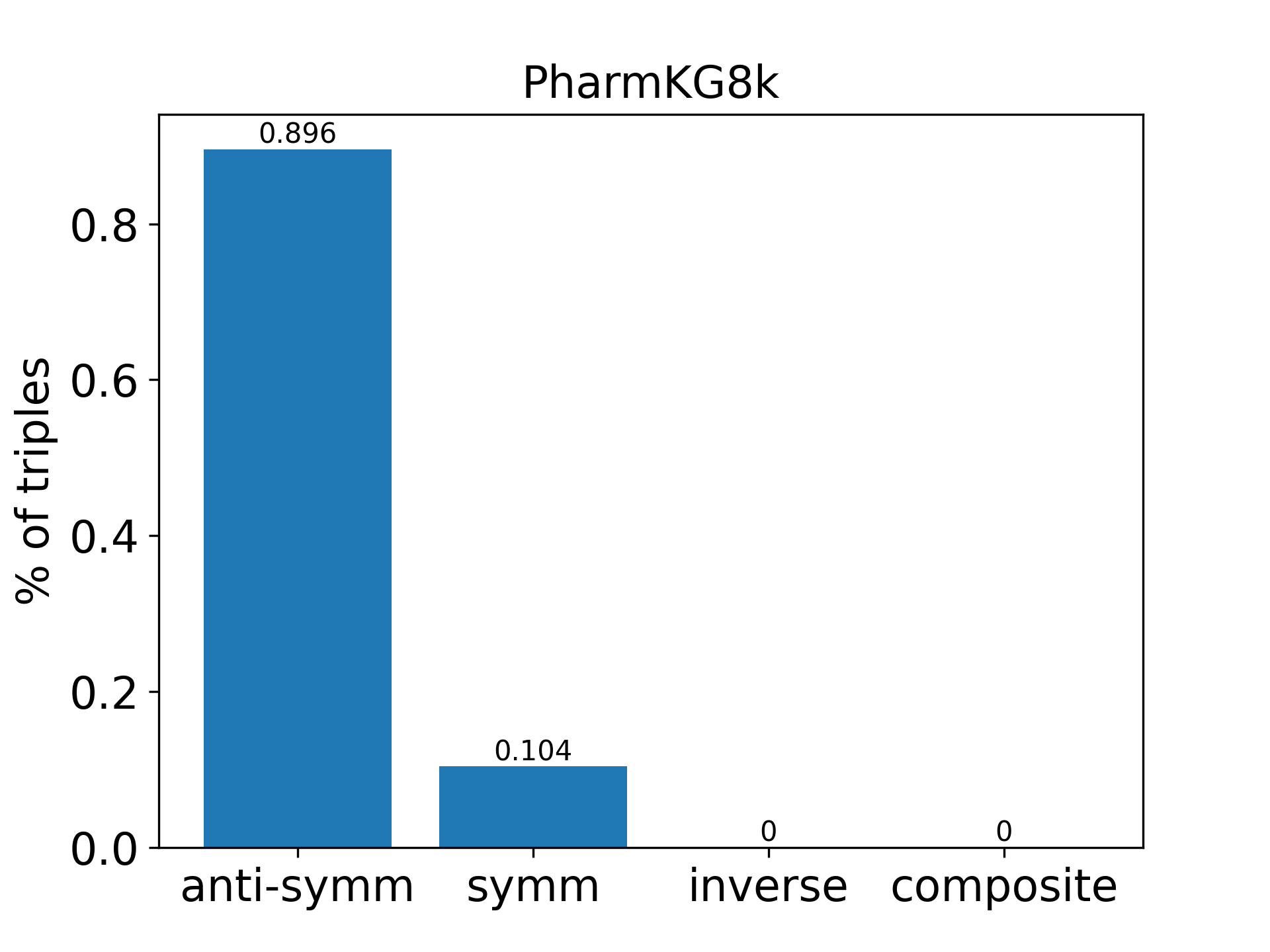}
    \end{minipage}    
     \begin{minipage}{0.24\textwidth}
         \includegraphics[width=\textwidth]{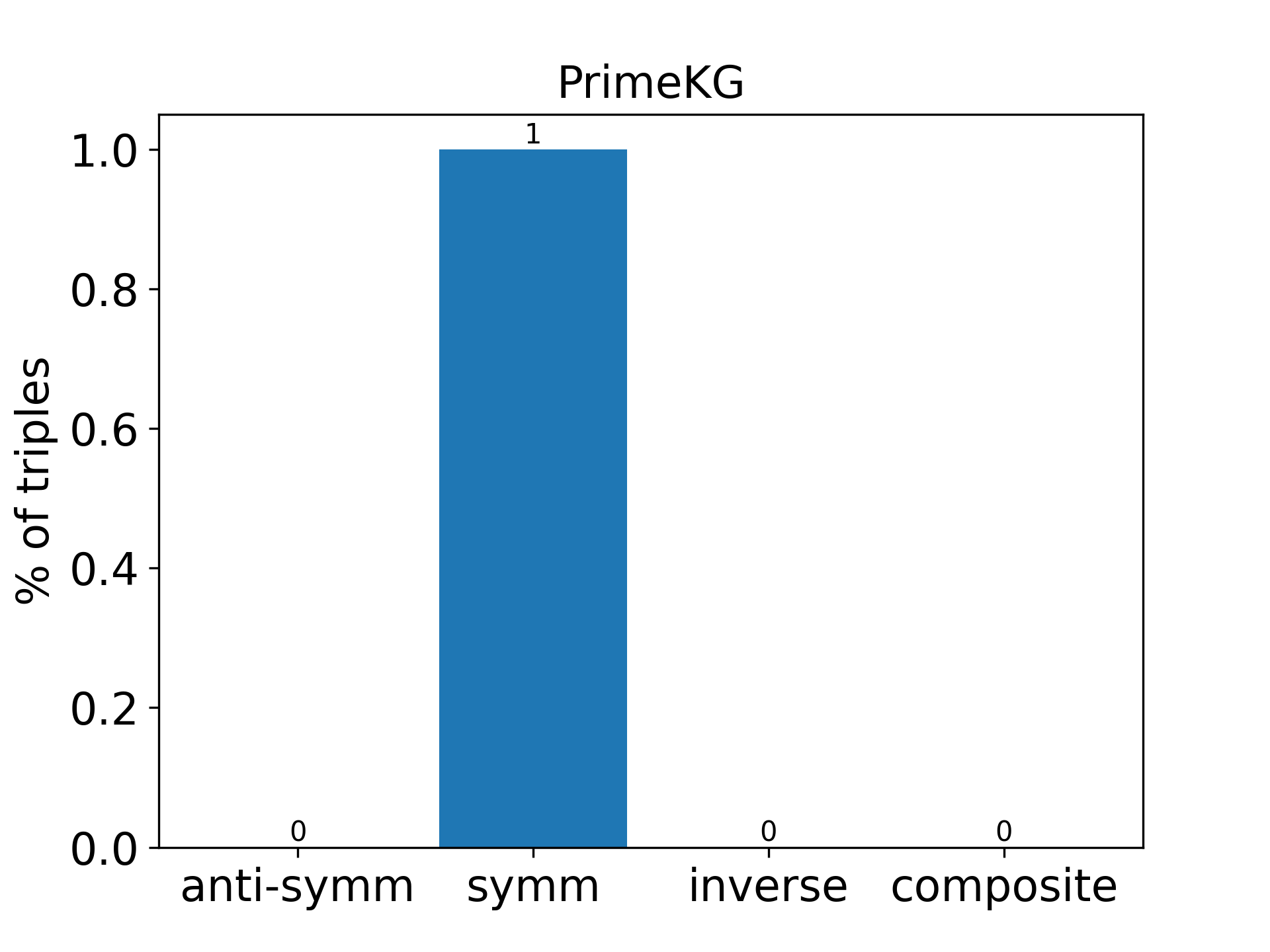 }
     \end{minipage}
     \begin{minipage}{0.24\textwidth}
         \includegraphics[width=\textwidth]{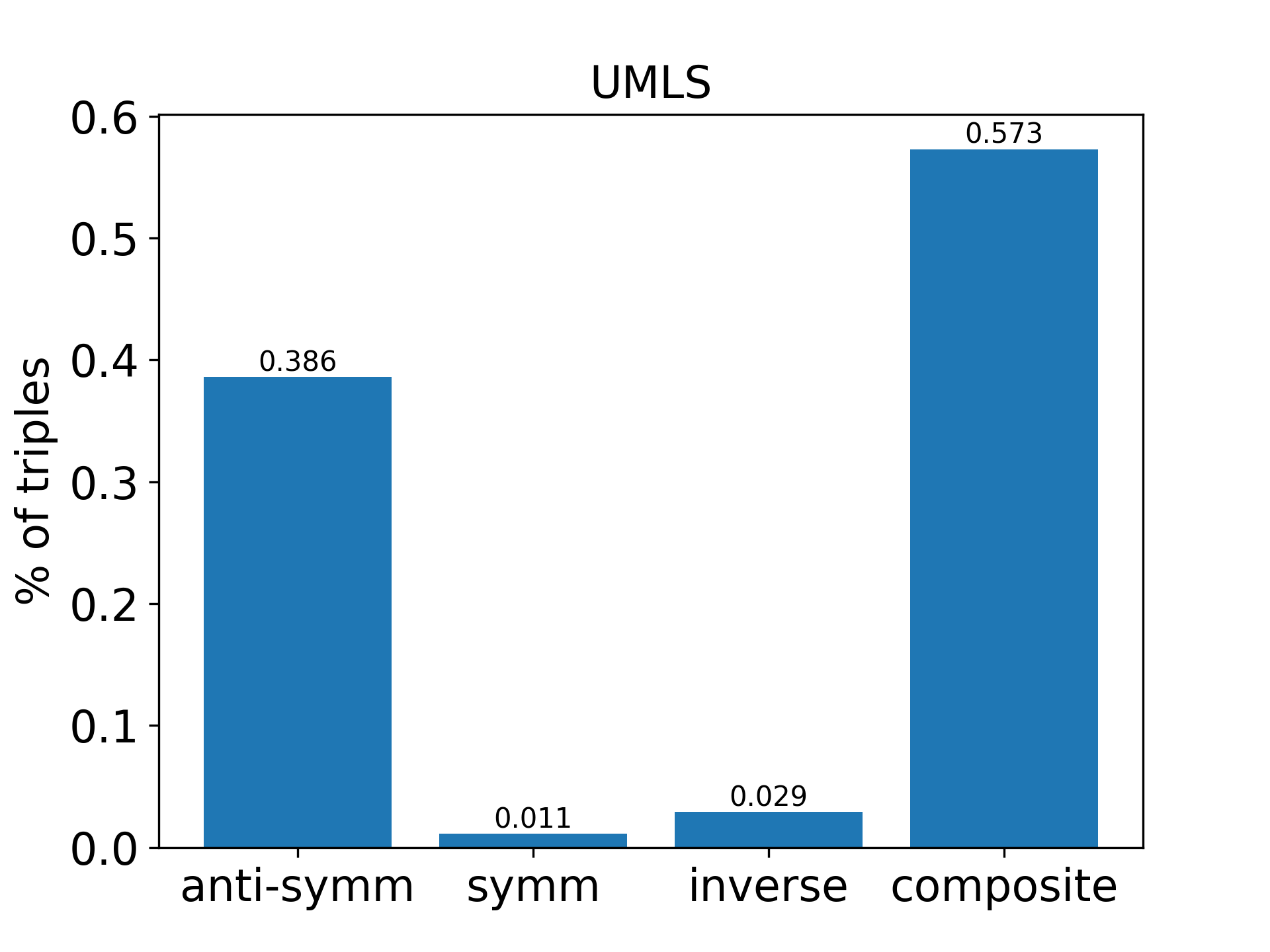 }
    \end{minipage}
    \begin{minipage}{0.24\textwidth}
         \includegraphics[width=\textwidth]{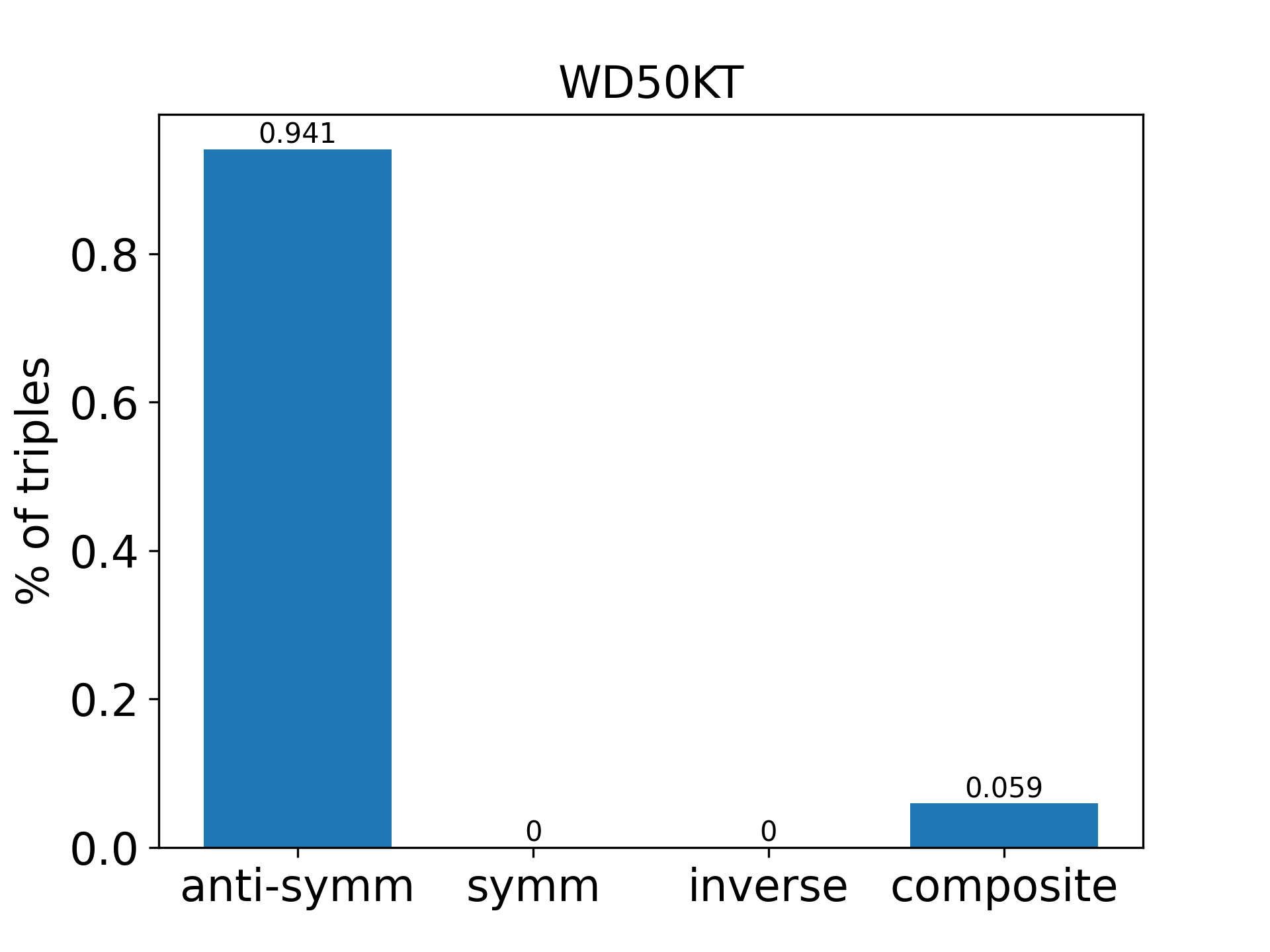}
    \end{minipage}
    
     \begin{minipage}{0.24\textwidth}
         \includegraphics[width=\textwidth]{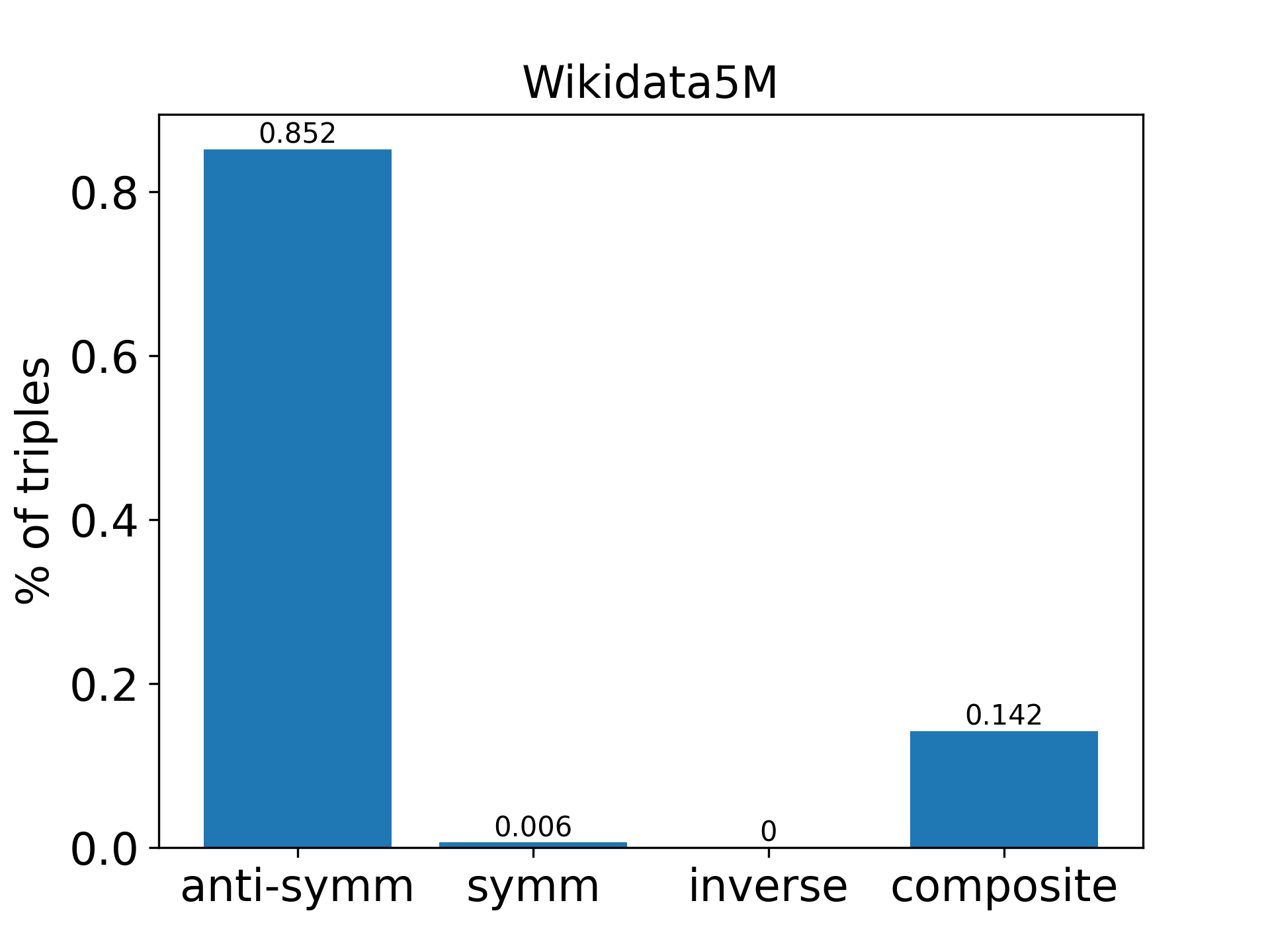 }
     \end{minipage}
     \begin{minipage}{0.24\textwidth}
         \includegraphics[width=\textwidth]{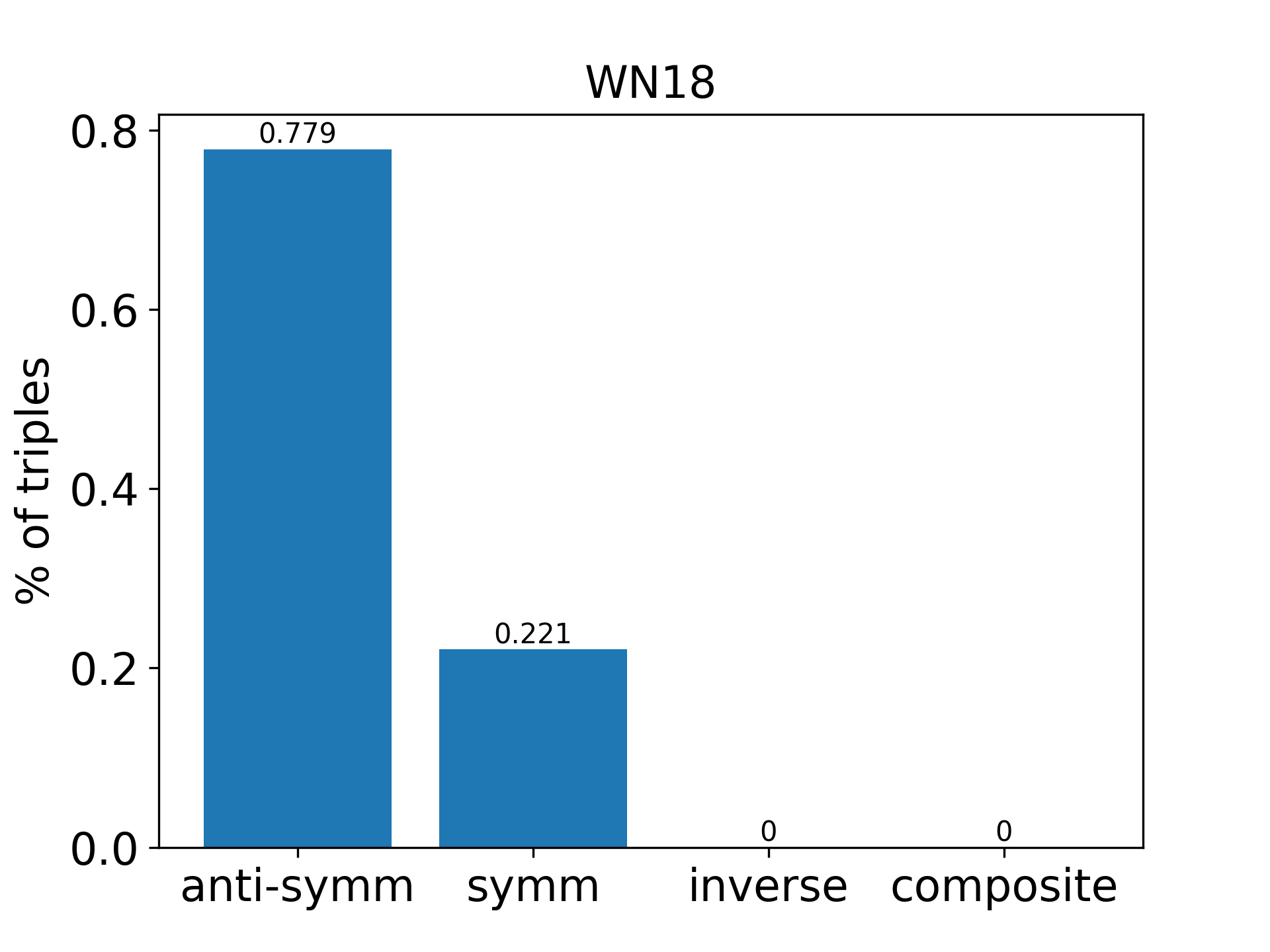}
    \end{minipage}
    \begin{minipage}{0.24\textwidth}
         \includegraphics[width=\textwidth]{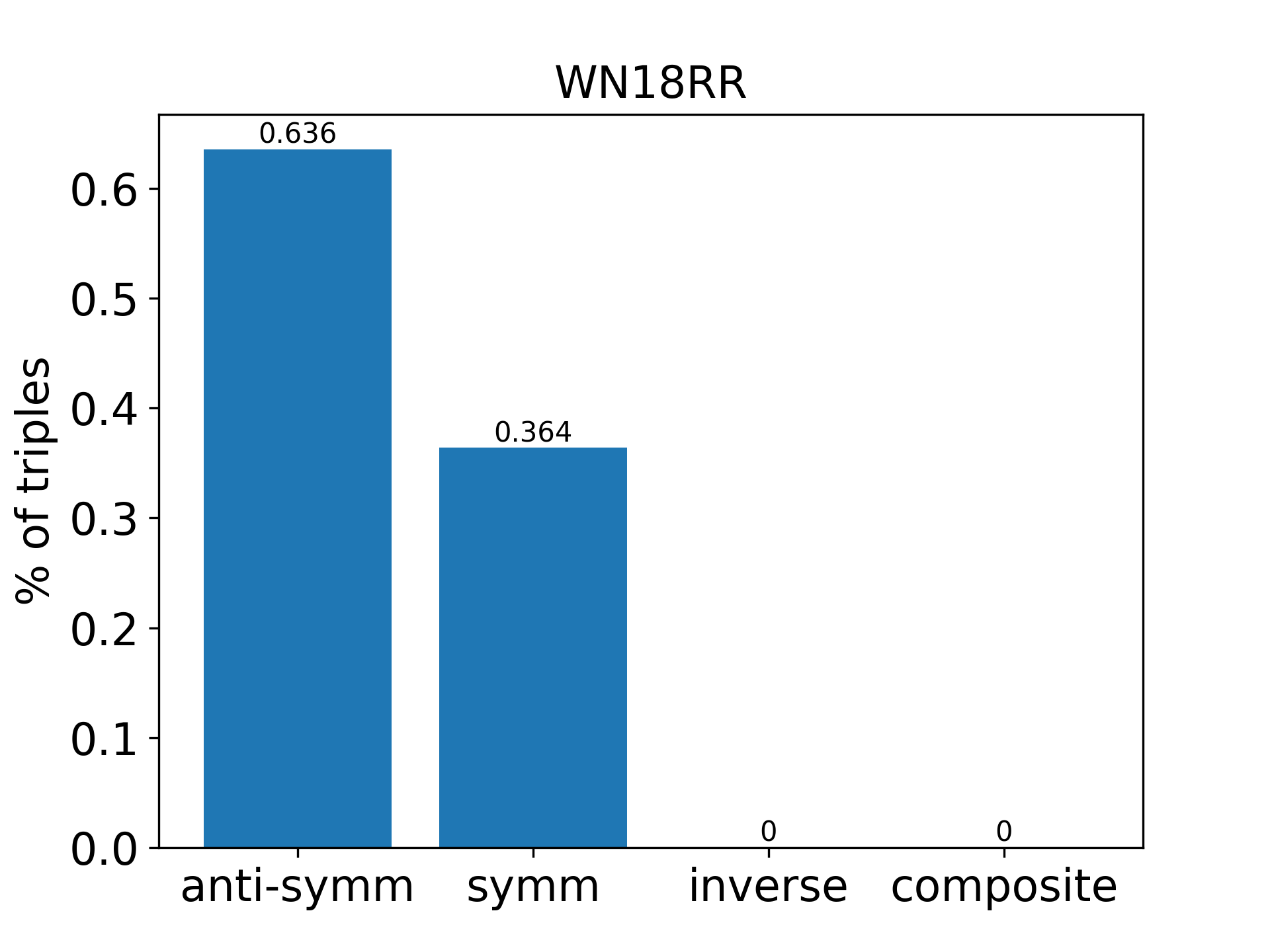}
    \end{minipage}
    \begin{minipage}{0.24\textwidth}
         \includegraphics[width=\textwidth]{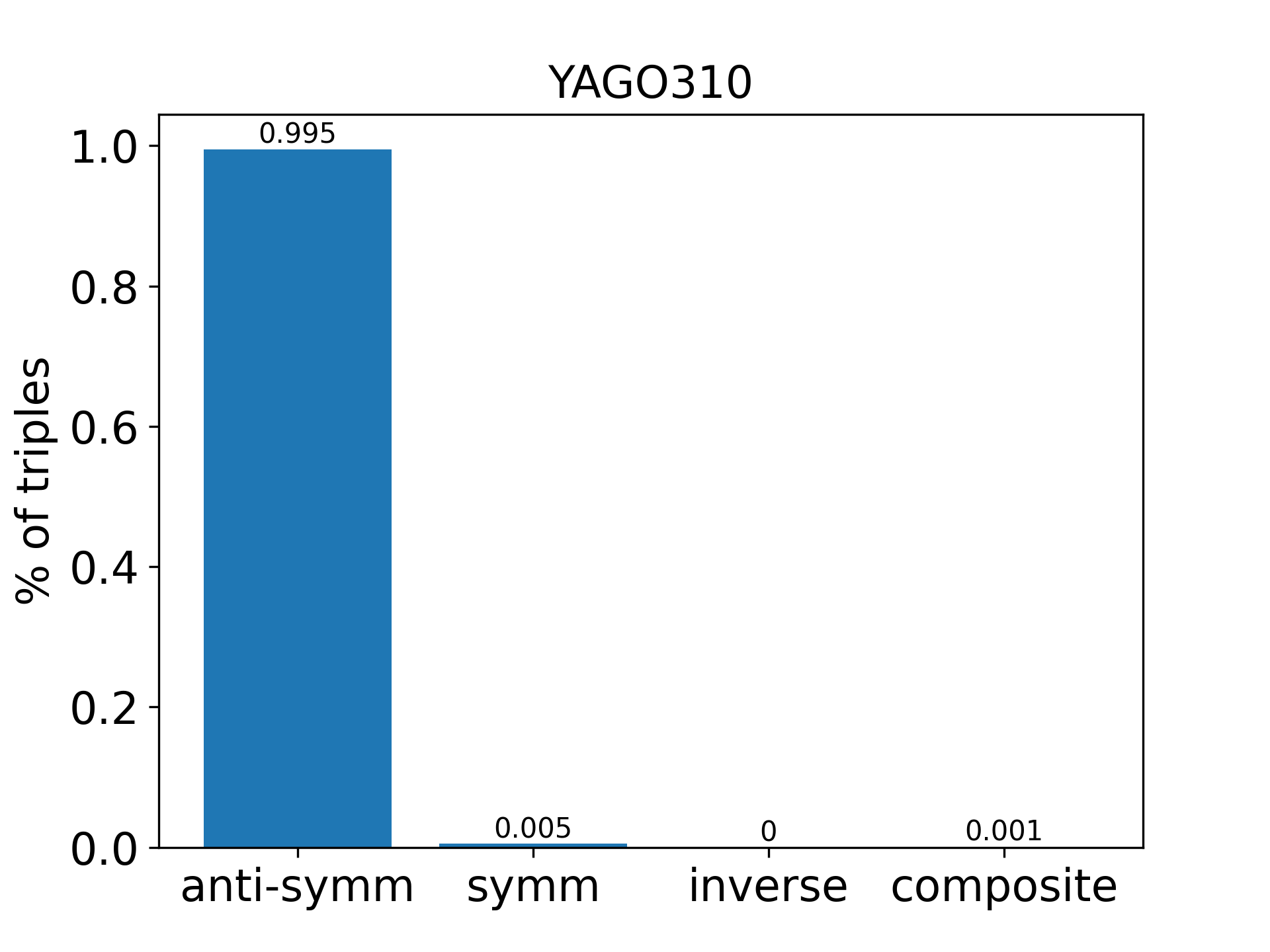}
    \end{minipage}
    
        \caption{Proportion of (anti-)symmetric, inverse and composite relations in different KGs. Each bar is marked with the \# of relations of the specified type with the \m{y}-axis showing the corresponding triples (as a \% of all triples). Due to space limits, the CoDExMedium dataset is shown in the Appx.}
        \label{fig: reltypes}
\end{figure}

\textbf{Relational patterns}. We considered four relation types previously described in the KG literature \citep {bringingiglight, toutanova}. A relation \m{r \inm R} is:
\begin{enumerate}[left=1pt, label={(\roman*)}]
    \item \textbf{asymetric} if \m{(h,r,t) \inm  T \!\! \implies\!\! (t,r,h) \! \notin\! T}
    \item \textbf{symmetric} if \m{(h,r,t) \inm T  \!\! \implies\!\!(t,r,h) \inm T}
    \item \textbf{inverse} to \m{r_{inv}\inm R} if  \m{(h,r,t) \inm T \!\! \implies\!\! (t,r_{inv},h) \inm T}.
    If there exists a \m{r^\prime \inm R}, s.t. \m{r^\prime \!\not=\!  r} and \m{r^\prime} is inverse of \m{r}, 
    then \m{r} is an inverse relation.
    \item \textbf{composite} of two relations \m{r_1,r_2 \inm R} if 
    \m{(x,r_1,y)\inm T\!\wedge\! (y,r_2,z)\inm T \!\!\implies\!\! (z, r, z)\inm T}. 
\end{enumerate}
Fig.  \ref{fig: reltypes} plots the relation pattern distribution for each dataset considered. Notably, some semantic datasets such as 
DB100K, OpenEA and DBpedia50 have a small amount of inverse relations which may affect benchmarking on these datasets in light of the "inverse relation problem" discussed above. Across all datasets we observe dominance of anti-symmetric relations, with the exception of the societal datasets and 
some of the most frequently used benchmarking datasets in NLP, such as FB15 and WN18RR, which show presence of all \m{4} relation types. 
Some semantic and societal datasets have composite relations, while none of the biomedical do.

\textbf{Metapaths}: 
KG metapaths are widely used in the biomedical literature for assessing the connectivity of KGs and deriving insights about the 
clinical or biological relevance of interactions such as gene-gene or drug effects \citep{su2020network,fu2016predicting, hetionet,zhang2020predicting}.
A metapath is defined as a sequence of relations separated by edge types (metanodes). For example, 
a metapath of length \m{\ell} is of the form 
$ e_1 \xrightarrow[]{r_1} e_2 \xrightarrow[]{r_2} \dots \xrightarrow[]{r_{\ell-1}} e_{\ell} $
where each of \m{e_1,e_2,\text{ and} e_{\ell}} belongs to a specific metanode. 
For example, in the Hetionet dataset\citep{hetionet}  the metanodes \texttt{Compound}, \texttt{Gene} and \texttt{Disease} form the metapath 
$ \texttt{Compound} \xrightarrow[]{\texttt{binds}} \texttt{Gene} \xrightarrow[]{\texttt{associates}} \texttt{Disease}$ 
of length \m{2}. 
The number of metapaths of a given length  provides a way for quantifying the level of  relational composition 
without having explicit composite relations encoded in the KG.
Interestingly, \cite{cohen2023crawling} test the reasoning abilities of PLMs by a prompting strategy that forces them to survey 
entity neighborhoods; 
although the authors do not put their work in the context of KG metapaths. 

In practice, metapaths of length of greater than \m{4} are considered too long to make a significant contribution in link prediction task \citep{hetionet, fu2016predicting}. 
Fig.  \ref{fig: multiples} (right) compares the metapath length distribution over paths of length \m{2,3} and \m{4} for all KGs (see Appx. for additional details).
From the figure we see that the biomedical KGs (with the exception of Globi) contain a significantly higher number of metapaths than the other KG types. 
However, the semantic dataset FB15k and the societal KGs Kinship and Nations exhibit a profile similar to the one of the biomedical datasets. 

\textbf{Findings}.
Tab. \ref{tbl: comparisontable} summarizes our observations above in a color coded visual to complement the analysis. 
Below we highlight the key takeaways from our empirical analysis and based on them, make several recommendations: 
\vspace{-6pt}
\begin{itemize}[left= 1pt]
    \item Given the dataset diversity observed in this study, we conclude that not all KGs are created equal.
    This has implications for model evaluation in ML, NLP and other domains in which KGs are used. 
    Thus, we recommend that researchers consider a broader set of datasets beyond the ones derived from the semantic web 
    for their KG model development and evaluation. 
    Fine-grained model evaluation -- for example, as a function of relation type, cardinality, KG density or degree distribution -- has the potential to further drive the development of new KG-based models or inform model selection given specific KG properties.
    \item Inverse relations are present in some datasets, including some released after the "inverse relation" leakage problem was reported by \cite{toutanova}. Given the implications of this problem in downstream applications,  
    we recommend KG libraries and benchmarks consider adding tools for handling/removing inverse relations in order to bring
    visibility to the leakage problems.  

    \item As mentioned in Sec. \ref{sec: kgapplication}, many recent knowledge-enhanced models have been proposed. 
    Based on our findings, we caution that the success of KG-enhanced PLMs may vary depending on the underlying KG structure. 
    While additional analysis is needed in this direction, given the varied structure of real KGs , development of domain-specific PLMs may benefit differently from different types of KGs.

      \item The overall negligible amount of composite relations in many datasets (including biological, semantic, and societal) is one
    interesting observation that merits further analysis. Composite relations lead to triangles in KGs and intuitively imply
    diminished reasoning pathways in KGs, which may have an effect on LMs-as-KGs benchmarks and evaluations.
    Additionally, the distinct presence of composite relations in FB15k237 (one of the most frequently used datasets in NLP research) may lead to flaws in KG-based NLP model evaluation, unless performance on a variety of other datasets is also considered.
    
    
    \item \cite{openbiolink} hypothesize that the size of biomedical KGs tends to be large, which calls into double whether model results from smaller datasets are informative. We argue that beyond size, practitioners should consider the role KG properties and structural patterns  
    during the design, hypotheses testing and model development.

    \item We hypothesize that analyzing the properties and structure of existing KGs can also benefit the future design of more robust KG datasets which incorporate diversity along different dimensions, such as the ones explored in this paper.
    
\end{itemize}




\vspace{-6pt}
\section{Conclusions and Future Work}\label{conclusion}
In summary, we analyzed \m{29} real KG datasets from several different domains and evaluated their properties and structural 
characteristics along various dimensions. Based on our findings we make several recommendations for KG-based model development and evaluation. 
Our study has implications for the broader use of KGs across domains: given the proliferation of models (in KG link prediction, EA, LM-as-KG evaluation) across domains (NLP, natural sciences, medicine and other disciples), it is worth investigating whether (and how) structural patterns, as well as their inter-domain variability across KGs, may correlate or influence KG model performance. Given the scope and scale of such an investigation, we leave it for a follow up study and encourage others -- within the NLP, ML and the ML for Sciences communities -- to further explore this topic.

\newpage
\bibliography{kg-bib}
\newpage

\appendix
\section{Appendix}\label{sec: appx}
\subsection{Datasets used} \label{sec: appx-data}
We used all the datasets in the PyKEEN library as described in the paper with the 
exception of several datasets (e.g., WK3l15k, WK3l120k, CN3l, and CKG) whose underlying files are no longer available for download at the URLs the library points to. 

\subsection{Degree plots}\label{sec: app-degrees}
The Nations and Kinship datasets were not included in Fig. \ref{fig: multiples} due to the high number of high degree nodes in them which leads to plot scaling issues of the remaining \m{26} datasets. The Nation's \m{14} entities have degrees in the range \m{146-514}; the Kinship's entities have degrees in the \m{192-206} range. 
For similar reasons we exclude the highest-degree entity (\texttt{men}) of the ConceptNet dataset in the plot in Fig. \ref{fig: multiples}.

\subsection{Relation types, cardinalities, and metapaths} \label{sec: appx-card}
Relationship type determination, i.e. whether a relation is (anti)-symmetric, inverse, composite, is based on association rule mining. The relation classifications are based on checking whether the corresponding rules hold with sufficient support and confidence -- we calculated the support using a confidence of \m{95\%}. We used the reference implementation in available in PyKEEN \cite{ali2021pykeen}. Note that a relation can be of several different types. 
Relation cardinality is computed similarly to the relation type. 

Metapath lengths are approximated by sampling (uniformly at random) an entity \m{e} from each KG and counting all the paths of length \m{2, 3} and \m{4} originating from \m{e}. Each KG was sampled \m{3} times, so the metapath numbers reported in Fig. \ref{fig: multiples} (right) are averaged over
\m{3} independent entity samples, for each KG. 

\subsection{Additional plots for Fig. \ref{fig: reltypes}, and Fig. \ref{fig: relcardinality}} \label{sec: appx-moreplots}
\begin{figure}[h]
     \centering
     \begin{minipage}{0.32\textwidth}
         \includegraphics[width=\textwidth]{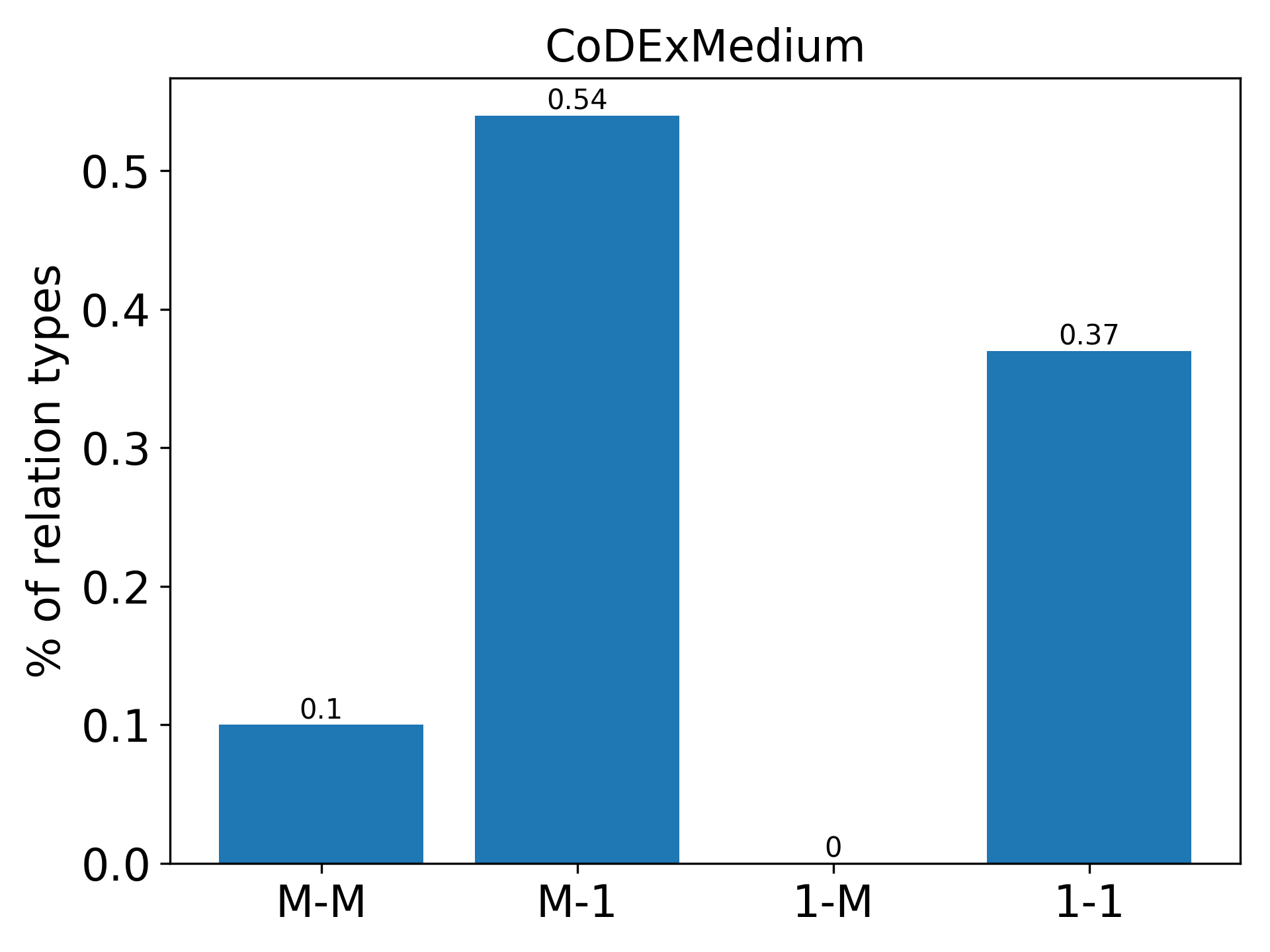}
     \end{minipage}
     \begin{minipage}{0.32\textwidth}
         \includegraphics[width=\textwidth]{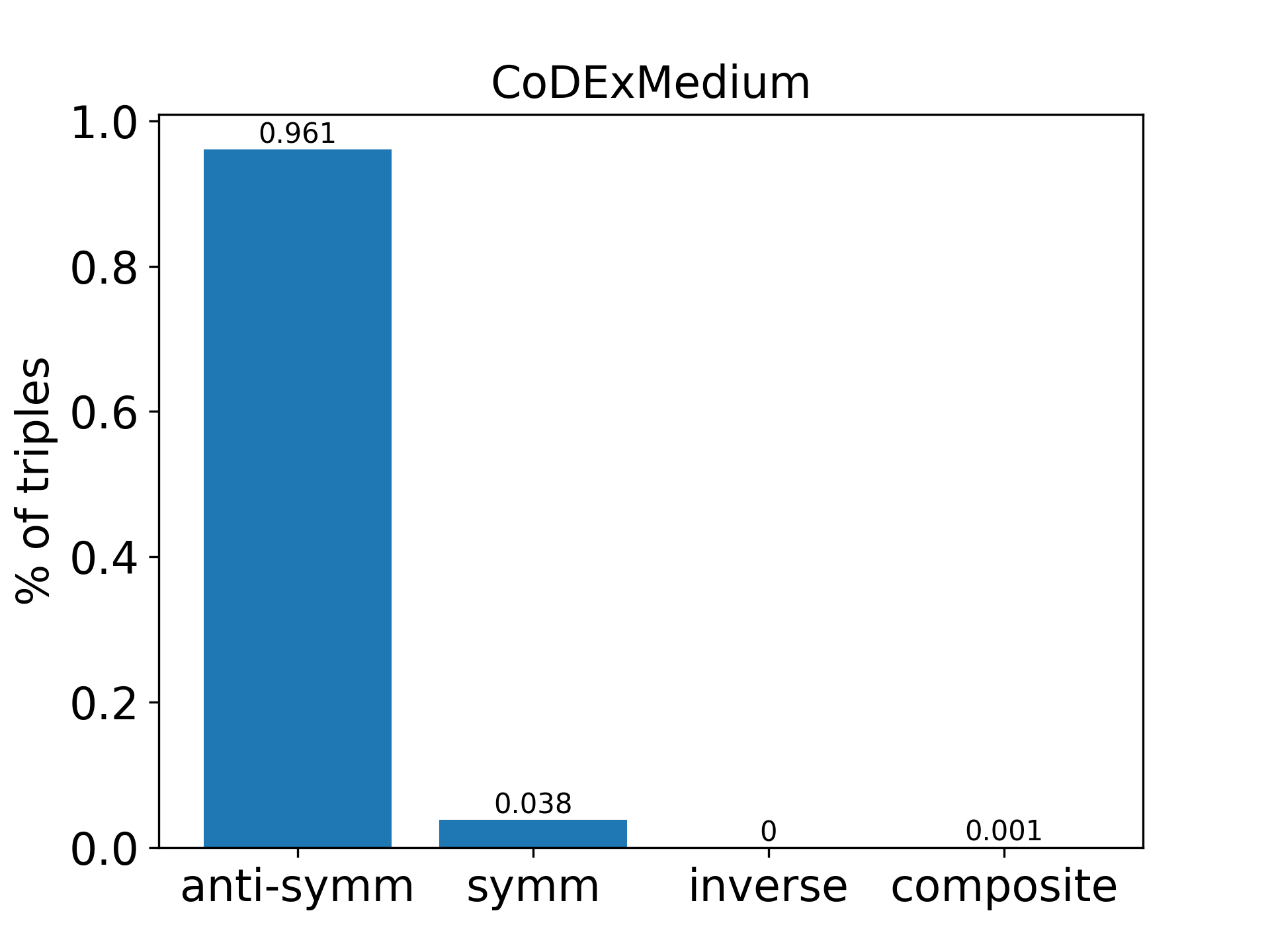}
    \end{minipage}
        \caption{CoDExMedium dataset: supplemental panels for Fig. \ref{fig: reltypes}, and Fig. \ref{fig: relcardinality}.}
        \label{fig: codex}
\end{figure}

%


\end{document}